%% file: main.tex
\documentclass[11pt]{article}

\usepackage[final]{acl}
\usepackage{array}
\usepackage{listings}
\usepackage{booktabs}
\usepackage{times}
\usepackage{latexsym}
\usepackage{xspace}
\usepackage{graphicx} 
\usepackage{natbib}
\usepackage{multirow, multicol}
\usepackage{tcolorbox}
\usepackage{graphicx}
\usepackage{subcaption}
\usepackage{amsmath}
\usepackage{float}
\usepackage{subcaption}  

\tcbset{
  colback=blue!4,      
  colframe=blue!35!black,  
  fonttitle=\bfseries, 
  boxrule=0.6pt,
  arc=2pt,
  left=6pt, right=6pt, top=6pt, bottom=6pt,
}

\newcommand{\xv}{\mathbf{x}}
\newcommand{\yv}{\mathbf{y}}

\newcommand{\CC}{\mathcal{C}}
\newcommand{\HC}{\mathcal{H}}

\usepackage{pifont}
\newcommand{\cmark}{\ding{51}}
\newcommand{\xmark}{\ding{55}}

\newcommand{\llama}{Llama-3.1-8b-Instruct\xspace}
\newcommand{\gemma}{Gemma-3-12b-Instruct\xspace}
\newcommand{\qwen}{Qwen-2.5-14b-Instruct\xspace}
\newcommand{\llamaseventy}{Llama-3.3-70b-Instruct\xspace}
\newcommand{\gptfour}{GPT-4.1\xspace}
\usepackage{todonotes}

\newcommand{\gpt}{GPT-5\xspace}
\title{Why Don't You Know? \\ Evaluating the Impact of Uncertainty Sources\\ on Uncertainty Quantification in LLMs}
\author{Maiya Goloburda$^{1,2}$ \quad Roman Vashurin$^{1}$ \quad Fedor Chernogorskii$^{1}$ \quad  Nurkhan Laiyk$^{1}$ \\ \quad \textbf{Daniil Orel}$^{1}$ \quad \textbf{Preslav Nakov}$^{1}$ \quad \textbf{Maxim Panov}$^{1}$
\\
 $^1$Mohamed bin Zayed University of Artificial Intelligence \\
$^2$School of AI, QAIRU \\
 {\small \texttt{\{Maiya.Goloburda, Preslav.Nakov, Maxim.Panov\}@mbzuai.ac.ae }} }

\begin{document}

\maketitle


\begin{abstract}
As Large Language Models (LLMs) are increasingly deployed in real-world applications, reliable uncertainty quantification (UQ) becomes critical for safe and effective use. Most existing UQ approaches for LLMs aim to produce a single confidence score -- for example, estimating the probability that a model's answer is correct.  However, uncertainty in natural language tasks arises from multiple distinct sources, including model knowledge gaps, output variability, and input ambiguity, which have different implications for system behavior and user interaction. In this work, we study how the source of uncertainty impacts the behavior and effectiveness of existing UQ methods. To enable controlled analysis, we construct a human-validated dataset that introduces distinct uncertainty sources in a controlled way, enabling the first systematic evaluation of how uncertainty source affects UQ method behavior. Our experiments reveal that method families fail in predictable, source-specific ways, and that no single method reliably handles all sources. These findings highlight the need for uncertainty-aware methods that explicitly account for the source of uncertainty in LLMs. Our dataset and code are available at \url{https://github.com/stat-ml/why_dont_you_know}.

\end{abstract}

\input{sections/introduction}

\input{sections/related_works}

\input{sections/dataset}
\input{sections/experiments}

\input{sections/discussion}

\input{sections/conclusion}

\input{sections/limitations}

\bibliography{custom}

\newpage
\raggedbottom

\appendix

\input{sections/appendix/app_dataset}

\input{sections/appendix/app_gpt_judge}

\input{sections/appendix/app_models}

\input{sections/appendix/app_methods}


\input{sections/appendix/app_generation_details}
\input{sections/appendix/app_detailed_results}

\input{sections/appendix/app_density_plots}
\input{sections/appendix/app_large_models}



\end{document}

%% file: sections/introduction.tex

\section{Introduction}


Large Language Models (LLMs) are increasingly deployed across a variety of domains, including high-stakes ones, like medicine, law, and education. Despite their impressive capabilities, they remain prone to producing factually incorrect generations~\citep{manakul-etal-2023-selfcheckgpt,farquhar2024detecting}, making it as important to know when not to trust a model's output as the output itself. This has driven growing interest in uncertainty quantification (UQ) for LLMs: methods that measure a model's confidence in its generations. 

Several families of methods have been proposed, including token probability-based, output consistency-based, hybrid (combining token probabilities and semantic consistency) and verbalized approaches~\citep{vashurin-etal-2025-benchmarking,xia-etal-2025-survey}. Despite their diversity, all share a fundamental limitation: reducing uncertainty to a single scalar confidence score.

  While such scores are useful as a proxy for the quality of the generation, they provide limited insight into the underlying source of uncertainty. However, in natural language generation, uncertainty can arise from fundamentally different sources
  -- limited model knowledge, legitimate output variability, and ambiguous inputs~\citep{Baan2023UncertaintyIN,Prabhudesai2024-tw,hou2024decomposing}. This raises an important question: \textit{how do existing UQ methods actually behave under different uncertainty sources?}
  
  

Meaningful comparison of UQ methods across uncertainty sources requires holding the domain fixed across variants of each question and keeping the format and length of model outputs comparable
, so that differences in UQ behavior can be attributed to the introduced condition rather than to confounders. We achieve this through \textit{matched question triplets}: for each question, we construct a single-answer variant where model knowledge is the sole source of uncertainty, a multi-answer variant that adds output variability on top, and an unanswerable variant requiring clarification that isolates input ambiguity. This lets the same UQ method be compared across conditions on matched content, using the single-answer variant as a baseline. Existing datasets do not offer this: they either introduce ambiguity without separating its sources -- as in AmbigQA~\citep{min-etal-2020-ambigqa} and ASQA~\citep{stelmakh-etal-2022-asqa} -- or target a single highly specific uncertainty source, such as EverGreenQA~\citep{pletenev-etal-2025-will}, which focuses on temporal uncertainty.

Our contributions are as follows:
  
  \begin{itemize}
  \item A human-validated dataset of matched question triplets -- single-answer, multi-answer, and clarification-requiring variants of the same question -- holding domain, format, and length constant to enable controlled comparison of UQ methods under model knowledge, output variability, and input ambiguity.
  
    
    \item We evaluate 16 uncertainty methods across 5 models under controlled uncertainty conditions, finding that performance degrades consistently once uncertainty sources beyond model knowledge limitations are introduced.

\item An analysis showing that output variability inflates and input ambiguity deflates uncertainty estimates across all method families, and explaining why in terms of each family's core assumptions.

  \end{itemize}

%% file: sections/related_works.tex
\section{Related Work}

\textbf{Sources of Uncertainty in NLG.} In machine learning, uncertainty is commonly decomposed into epistemic and aleatoric uncertainty~\citep{kendall2017uncertainties,Hllermeier2021}. Epistemic uncertainty stems from gaps in model knowledge and can, in principle, be reduced. Aleatoric uncertainty is considered inherent to the task and is irreducible. However, in natural language generation (NLG), uncertainty can arise from a broader variety of sources~\citep{Prabhudesai2024-tw, Baan2023UncertaintyIN, hu2023uncertaintynaturallanguageprocessing}. Existing taxonomies largely agree on three high-level origins of uncertainty: input, model, and output. These do not always map cleanly onto the epistemic/aleatoric dichotomy: output variability reflects genuine semantic diversity rather than irreducible noise, and input ambiguity is reducible through clarification~\citep{kirchhof2025position, Baan2023UncertaintyIN}. In our work, we treat these three sources as operationally distinct, as each corresponds to different model behavior and calls for a different system response.

\textbf{Uncertainty Quantification for LLMs.}
  Recent work on Uncertainty Quantification for LLMs has proposed a diverse set of methods that attempt to estimate the confidence of the model. 
  These methods can be grouped into five main categories: token-probability-based, consistency-based, hybrid, verbalized,  and internal state-based~\citep{vashurin-etal-2025-benchmarking, xia-etal-2025-survey}. 
  
  Token-probability-based methods derive uncertainty from the model's predictive distribution; consistency-based methods measure agreement across multiple sampled responses; hybrid methods combine both signals; verbalized methods prompt the model to explicitly state its confidence; and internal-state-based methods use hidden representations to estimate uncertainty. In this work, we focus on the first four families as they are the most widely adopted in practice and require no access to model internals beyond token probabilities and sampled outputs, making them applicable across a broad range of deployments. While reflecting different perspectives on how uncertainty in LLM outputs can be measured, these methods are designed to produce a single scalar uncertainty score. Consequently, existing UQ approaches do not explicitly characterize which underlying sources of uncertainty contribute to a given score, limiting their ability to distinguish between qualitatively different forms of uncertainty. Recent work has begun to expose this limitation: \citet{tomov2026illusioncertaintyuncertaintyquantification} show that UQ methods can degrade to near-random performance under aleatoric uncertainty and provide a theoretical account of this failure. We complement and expand upon this by adopting a three-source view of uncertainty, evaluating methods under model knowledge limitations, output variability, and input ambiguity in a controlled, comparable setting.


\begin{table}[t]
\centering
\footnotesize
\setlength{\tabcolsep}{4pt}
\begin{tabular}{lccc}
\toprule
\textbf{Dataset} & \textbf{3 Sources} & \textbf{Labeled} & \textbf{Comparable} \\
& \textbf{Covered?} & \textbf{Sources?} & \textbf{Variants?} \\
\midrule
AmbigQA & \cmark & \xmark & \xmark \\
MAQA & \xmark & \cmark & \xmark \\
EverGreenQA & \xmark & \cmark & \xmark \\
AbstentionBench & \xmark & \cmark & \xmark \\
CLAMBER & \xmark & \cmark & \xmark \\
\midrule
\textbf{Ours} & \cmark & \cmark & \cmark \\
\bottomrule
\end{tabular}
\caption{Comparison with existing uncertainty-focused datasets.}
\label{tab:dataset-comparison}
\end{table}


\textbf{Uncertainty-Focused Datasets.}
Several existing works address ambiguity and uncertainty in NLG, particularly in question answering, but most target a single source of uncertainty.  
  AmbiNQ~\citep{min-etal-2020-ambigqa} consists of questions that can be interpreted in several ways due to underspecification, missing references, etc.
  MAQA~\citep{yang-etal-2025-maqa} deals with questions that have  multiple possible answers. 
Other work targeted more specialized sources of uncertainty. 

For example, EverGreenQA focuses on questions with time-sensitive answers~\citep{pletenev-etal-2025-will}. Other benchmarks examine how models handle questions they cannot answer as written. CLAMBER tests whether models identify and clarify ambiguous queries~\citep{zhang-etal-2024-clamber}, while SQuADRUn and AbstentionBench test abstention on unanswerable questions~\citep{rajpurkar-etal-2018-know,kirichenko2026abstentionbench}. Because these benchmarks differ in both question content and evaluation goals, we use variants of the same question to compare UQ methods across uncertainty sources. We summarize dataset differences in Table~\ref{tab:dataset-comparison}.

%% file: sections/dataset.tex
  \begin{figure*}[ht!]
    \centering
    \includegraphics[width=\linewidth]{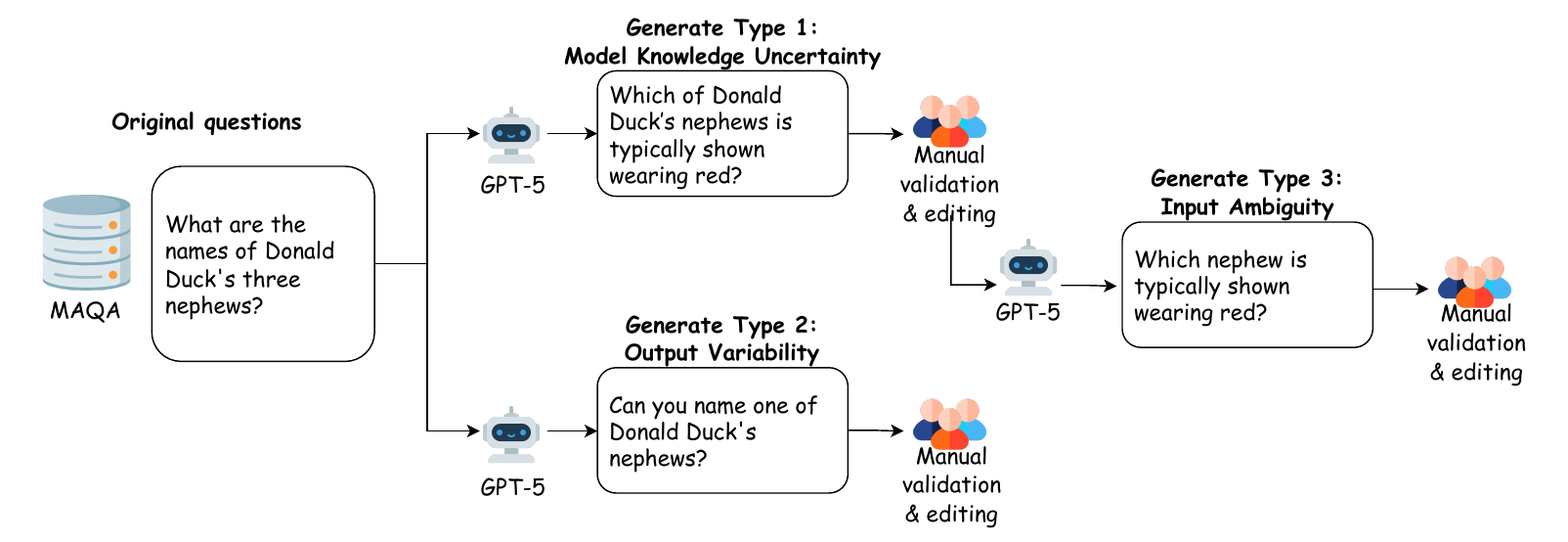}
    \caption{Question generation pipeline used to construct question triplets.}
    \label{fig:dataset_creation}
  \end{figure*}

\section{Dataset Creation}

  In this section, we describe the proposed dataset design, data sources, and construction procedure.

\subsection{Choice of Uncertainty Sources}
  
Prior work has proposed a wide range of uncertainty sources in natural language generation, including more specific subcategories like linguistic variability, and uncertainty induced by stochastic decoding. However, many of these proposed uncertainty sources are either difficult to isolate empirically, model-dependent, or tightly correlated. Thus, we intentionally focus on a minimal but operational subset of uncertainty sources that (i) frequently arise in real-world question answering, (ii) can be cleanly separated through controlled dataset construction, and most importantly (iii) correspond to distinct downstream actions a system might take. This motivates our focus on the following three sources of uncertainty: model knowledge uncertainty, output variability, and input ambiguity.

\textbf{Model Knowledge Uncertainty.} 
  When the question is well-specified and presupposes a single correct response, we can conclude that uncertainty stems from limited model knowledge. Linguistic variability, such as semantically equivalent responses that differ only in wording, can still confound the true model uncertainty. However, many modern UQ methods~\cite{kuhn2023semantic,duan-etal-2024-shifting,fadeeva2024factchecking} are well-equipped to handle this interference via semantic clustering of model outputs. Operationally, this type of uncertainty should lead either to abstention or to the use of external knowledge sources that augment the model's parametric knowledge.
  
\textbf{Output Variability.}
  A well-specified question can have several semantically distinct plausible answers. For instance, the question \textit{``Give me an example of an animal that lives in the desert.''} can be correctly answered by naming any of the desert-dwelling animals. In general setting, it's not obvious why this particular case should result in high uncertainty -- any response is as good as the other and can be given with confidence. Within probabilistic model of modern LLMs, however, this will likely result in the spread of the probability mass among candidate responses. For example, \citet{yang-etal-2025-maqa} observe empirically that peak answer probability decreases with the number of valid answers while their aggregate remains stable, indicating a spread of probability mass across valid candidates. Drawing outputs from this distribution will result in semantically inconsistent responses. Some of the most prominent families of the UQ methods will treat this signal as high uncertainty.

  \textbf{Input Ambiguity.}
  An ambiguous input naturally induces uncertainty in the output. Such ambiguity can be caused by a multitude of reasons. Unresolved references to entities or actors, temporal ambiguities, missing qualifiers, coreference and many other qualities of the question, when present, can make it underspecified and preclude answering it with confidence. Moreover, underspecification complicates further analysis of uncertainty. Indeed, it's much harder to assess if the model knows the answer to a question if it's unclear what the question is actually about. From a decision-making perspective, uncertainty stemming from input ambiguity is best resolved by asking the user to clarify the question. 




\subsection{Dataset Construction}

  To construct the dataset, we start from a subset of the MAQA dataset~\citep{yang-etal-2025-maqa} derived from Natural Questions, containing questions with a list of possible answers. 
  We start from multi-answer questions because it is generally easier to convert a multi-answer question into a single-answer than the other way around. A full dataset construction pipeline can be seen on Figure~\ref{fig:dataset_creation}.

\textbf{Type 1: Model Knowledge Uncertainty.} 
 For this type, we start from multi-answer questions in MAQA, randomly select 1 valid answer per question as the target, and prompt \gpt to rewrite the question so that only this target remains correct. The goal is a well-posed, unambiguous question whose unique answer is the chosen target.

All generated question--answer pairs are then validated by human annotators, who verify (1) that each question is correct and unambiguous and (2) that the corresponding answer is factually accurate. To check correctness, annotators perform an independent Google search and cross-check the retrieved evidence. If an answer is incorrect or inconsistent with the question, they manually rewrite the question to match the target answer; questions that are difficult to convert into a single-answer format are rejected. This step ensures both the linguistic precision and the factual reliability of the single-answer data.


\textbf{Type 2: Output Variability.}
  We then construct questions of Type 2. The questions in the MAQA dataset are associated with a list of correct answers and are typically phrased to imply that all valid answers should be provided. However, our investigation requires that a question be answered correctly by naming one valid answer rather than enumerating all possible answers. Thus, we reformulate each question to require only a single correct response, where any one of the valid answers is accepted as correct. For example, the original question ``What were the names of the two main factions in the English Civil War?'' was reformulated into ``Name one of the two main factions in the English Civil War.'' This reformulation is performed using \gpt model, followed by human validation. Note that model knowledge uncertainty remains present in this setting: a model may not know any of the valid answers. We can cleanly separate the two sources of uncertainty post-hoc -- a generation matching any valid answer reflects output variability as the operative source, while an incorrect answer indicates model knowledge limitations.



\textbf{Type 3: Input Ambiguity.}
We derive Type 3 questions from validated Type 1 questions to keep input ambiguity separate from output variability. This ensures each base question has a unique answer, so any uncertainty in Type 3 stems solely from underspecification. We prompt \gpt to remove key context - typically replacing specific referents with vague expressions - introducing referential ambiguity while avoiding overlap with other uncertainty sources. As a result, questions must be underspecified enough that no interpretation is more defensible than its alternatives, yet retain enough domain content to avoid collapsing into generic, context-free forms. \gpt struggled to achieve this unassisted, which we attribute to the difficulty of representing genuine ambiguity rather than a construction flaw. Annotators verify that no interpretation is meaningfully more likely than another and that sufficient context was removed, manually editing cases that remain weakly ambiguous or resolvable by convention. We reject questions that collapse into generic forms (e.g., ``Who said this?''). By design, Type 3 questions are unanswerable without clarification.

\input{sections/tables/eval_settings}

Finally, we compile question triplets, discarding any triplet with a question rejected at any stage. This yields 539 questions per uncertainty type (1,617 total), with uncertainty sources introduced in a controlled way and consistent structure and domain across all types. The balanced design ensures fair comparison across uncertainty sources. While \gpt assists in question generation, every question underwent mandatory human accept/edit/reject annotation, with substantial edits, especially for Type 3 questions, limiting the influence of \gpt's own preferences on the data. See prompts and annotator guidelines in Appendix~\ref{sec:dataset_construction}.

%% file: sections/tables/eval_settings.tex
\begin{table*}[h!]
\centering
\small
\scalebox{0.9}{
\begin{tabular}{p{1.8cm} p{2cm} p{5cm} p{5.6cm}}
\toprule
\textbf{Setting} & \textbf{Question Types} & \textbf{Description} & \textbf{Post-hoc Disentanglement} \\
\midrule
\textbf{MK} & Type 1 & Well-formed questions with a unique correct answer. Model knowledge is the sole source of uncertainty. & Correct: no uncertainty present (model knew the answer). Incorrect: reflects model knowledge limitations alone. \\
\midrule
\textbf{MK + OV} & Type 2 & Questions admitting multiple valid answers. Output variability is introduced alongside model knowledge uncertainty. & Correct: reflects output variability (the model knew \emph{an} answer, but which one varies). Incorrect: reflects model knowledge limitations. \\
\midrule
\textbf{MK + IA} & Type 1 + Type 3 & Mix of well-formed and underspecified questions. A well-calibrated method must remain confident on correctly answered Type 1 and assign high uncertainty to Type 3. & For Type 1: correct/incorrect as in MK. For Type 3: any committed answer (rather than a request for clarification) reflects input ambiguity, regardless of whether the committed answer happens to be correct. \\
\midrule
\textbf{MK + OV + IA} & Type 1 + Type 2 + Type 3 & All question types combined, reflecting realistic deployment conditions where all uncertainty sources co-occur. & Combines the conditioning above across question types, reflecting realistic settings where all three sources co-occur and cannot be disentangled by the system in advance. \\
\bottomrule
\end{tabular}}
\caption{Evaluation settings. Model knowledge (MK) uncertainty is present in all settings as an inherent property of any deployed system; remaining settings layer output variability (OV) and input ambiguity (IA) on top. The final column describes how post-hoc correctness conditioning disentangles the constituent sources within each setting.}
\label{tab:eval_settings}
\end{table*}

%% file: sections/experiments.tex

\section{Experiments}

\subsection{Experimental Setup}
\label{sec:eval_setup}
  We run all experiments through the \texttt{lm-polygraph} library~\citep{fadeeva-etal-2023-lm,vashurin-etal-2025-benchmarking}, which offers standardized implementations of UQ methods, ensuring consistent evaluation across methods.

\textbf{Models.} We evaluate 4 open-weight models -- \llama, \gemma, \qwen, and \llamaseventy and a proprietary \gptfour~\citep{llama,gemma,qwen, gpt41}. Detailed in Appendix~\ref{sec:models}. 


\textbf{Methods.}
We evaluate a broad range of methods from four families. Token probability-based methods include Sequence Probability (SP), Perplexity (PPL), Mean Token Entropy (MTE)~\citep{fomicheva-etal-2020-unsupervised}, 
  and Monte Carlo (Normalized) Sequence Entropy (MCSE/MCNSE)~\citep{kuhn2023semantic,malinin2021uncertainty}. Consistency-based methods include Number of Semantic Sets (NumSemSets), Degree Matrix (DegMat), Sum of Eigenvalues of the Graph Laplacian (EigValLaplacian), Consistency and Eccentricity~\citep{lin2024generating}. State-of-the-art hybrid methods include Semantic Entropy~\citep{kuhn2023semantic}, 
  Semantic Density~\citep{qiu2024semantic}, CoCoA SP, CoCoA PPL, and CoCoA MTE~\citep{vashurin2025cocoa}. The verbalized family is represented by 2S-Verbalized Uncertainty~\citep{tian-etal-2023-just}. All methods are described in detail in Appendix~\ref{sec:uq_methods}.

\input{sections/tables/prr_main}

\textbf{Evaluation Metrics.}
We adopt Prediction Rejection Ratio (PRR) as our primary evaluation metric for selective generation, following prior work~\citep{vashurin-etal-2025-benchmarking}. PRR measures how well an uncertainty score ranks predictions by quality, normalized against random and oracle rejection (Figure~\ref{fig:prr}); full details are in Appendix~\ref{app:prr}. We cap rejection at 75\%, as rejecting all generations is impractical in real-world applications. We additionally report AUROC, treating correctness as a binary label and the uncertainty score as the ranking signal (Appendix~\ref{sec:detailed_experimental_results}).

\textbf{Quality Metric.} We employ an LLM-as-a-judge (\gpt) to assess generation quality. For Type 1 and Type 2 questions, the judge labels each response as correct or incorrect with respect to reference answer. For Type 3 questions, it determines whether the model requested clarification (correct) or committed to a specific answer (incorrect). Each generation is assigned a binary score of 1 for correct and 0 for incorrect. Full prompt templates and human validation are in Appendix~\ref{sec:judge}.

\textbf{Evaluation Settings.} We assess uncertainty quantification methods across 4 settings, summarized in Table~\ref{tab:eval_settings}. The settings add uncertainty sources progressively, letting us trace how each affects UQ performance. The \textbf{MK} setting evaluates Type 1 questions in isolation, establishing a baseline where model knowledge uncertainty is the sole active uncertainty source. The \textbf{MK + OV} setting evaluates Type 2 questions, where output variability is introduced in addition to the model knowledge uncertainty.

Because model knowledge uncertainty remains present in MK + OV, we additionally disentangle the two post-hoc by conditioning on answer correctness: correct generations isolate output variability, since the model produced a valid answer but any of several were acceptable, while incorrect generations isolate model knowledge limitations (Table~\ref{tab:eval_settings}, Figure~\ref{fig:ue_distributions} in Section~\ref{sec:discussion}). The \textbf{MK + IA} setting jointly evaluates methods on Type 1 and Type 3 questions: since Type 3 questions are unanswerable by construction, they are mixed with Type 1 questions to create a setting where a well-calibrated method must remain confident on correct answers, assign high uncertainty to incorrect ones, and also assign high uncertainty to unanswerable inputs. Finally, the \textbf{MK + OV + IA} setting combines all three question types, reflecting the realistic deployment scenario where all sources of uncertainty co-occur and no single source dominates.

\subsection{Results}



\begin{figure*}[t!]
    \centering
    \begin{subfigure}{0.32\textwidth}
        \centering
        \includegraphics[width=\linewidth]{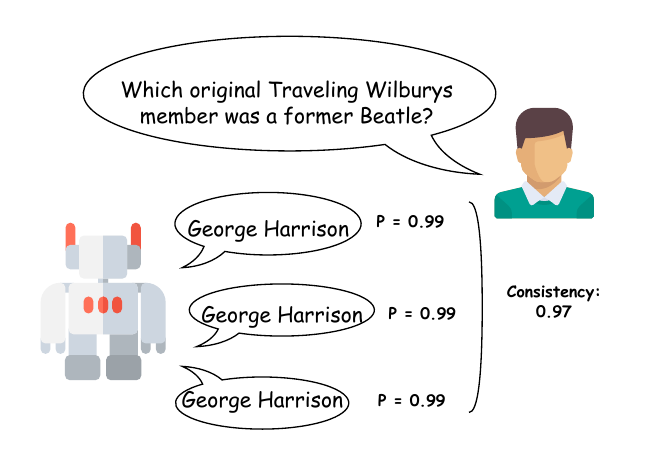}
    \end{subfigure}
    \hfill
    \begin{subfigure}{0.32\textwidth}
        \centering
        \includegraphics[width=\linewidth]{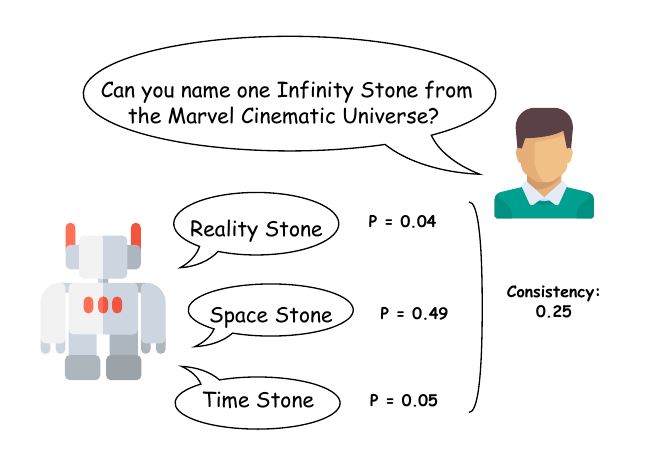}
    \end{subfigure}
    \hfill
    \begin{subfigure}{0.32\textwidth}
        \centering
        \includegraphics[width=\linewidth]{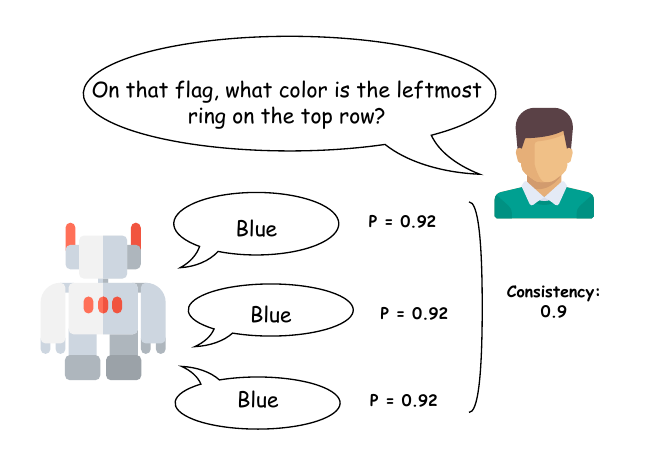}
    \end{subfigure}

    \caption{Examples of UQ failure modes across question types. Type 1: consistent, high-probability answers are correct. Type 2: inconsistency reflects valid output diversity, not poor quality. Type 3: the model commits to one interpretation of an ambiguous input, producing falsely confident, consistent answers.}
    \label{fig:examples}
\end{figure*}

  \textbf{Selective Generation Performance.}
Table~\ref{tab:prr_avg_methods_rows} reports PRR scores averaged across all five models. Detailed experimental results, as well as corresponding AUROC scores are reported in Appendix~\ref{sec:detailed_experimental_results}. As can be seen from these results, all methods experience a drop in performance once model knowledge limitation is no longer the only source of uncertainty. 

In the \textbf{MK}-only setting, Semantic Density achieves the highest PRR, followed by Verbalized2S, suggesting that hybrid and verbalized confidence signals are most informative when the input is well-formed and a unique correct answer exists. Verbalized2S improves with model size (Appendix~\ref{sec:detailed_experimental_results}), consistent with \citet{kadavath2022languagemodelsmostlyknow,tao2025revisiting}.
When output variability is introduced, all methods experience a drop in PRR (MK+OV), indicating that existing approaches struggle to distinguish between a model's knowledge limitations and the multiplicity of valid answers; Section~\ref{sec:discussion} separates these two effects via correctness conditioning and traces the failure to each method family's core assumptions. Hybrid methods remain the best performing on average, though the performance gap between them and token-probability- and consistency-based methods narrows compared to the MK-only setting. Verbalized2S suffers the sharpest performance drop, however, the performance is model-dependent: it remains competitive for \gptfour, while collapsing for \llamaseventy. It must be noted, however, that while the drop is consistent on average, it is not uniform across all method--model pairs. Namely, for \llama and \qwen, 4 and 7 of the 16 methods, respectively, score slightly higher under MK+OV than under MK; we discuss a possible explanation in (Appendix~\ref{sec:detailed_experimental_results}).

Under input ambiguity (MK+IA), all methods again degrade, but for a different reason: a model committing to one interpretation may produce overconfident, self-consistent answers, masking the underlying uncertainty.

Hybrid methods again lead, but the gap to pure token-probability methods narrows further. Verbalized confidence remains heavily model-dependent -- as detailed in Appendix~\ref{sec:detailed_experimental_results}. For \llamaseventy and \qwen Verbalized confidence is top performering method under this setting, suggesting that some models are better able to recognize and express uncertainty stemming from question underspecification.


In the combined setting (MK+OV+IA), MCSE emerges as the strongest method, with Semantic Entropy close behind. When all sources of uncertainty are present simultaneously, hybrid methods appear to offer diminishing returns for uncertainty quantification performance.

AUROC follows a similar pattern to PRR, with UQ performance degrading in the same manner 
across settings, indicating that this degradation is not an artifact of the particular 
evaluation metric (Appendix~\ref{sec:detailed_experimental_results}).

  \textbf{Model Behavior.} As shown in Table~\ref{tab:gen_quality}, all models achieve comparable accuracy on single-answer and 
multi-answer questions, and none reject answering in either 
setting unprompted. For ambiguous inputs, rather than requesting clarification, 
models overwhelmingly commit to an arbitrary interpretation and respond with 
apparent confidence. Generation length distributions are 
comparable across uncertainty types, confirming that differences in UQ behavior 
are not artifacts of length variation (Appendix~\ref{sec:gen_details}).

%% file: sections/tables/prr_main.tex
\begin{table*}[h!]
\centering
\scalebox{0.93}{
  \begin{tabular}{lcccc}
  \toprule
  \textbf{Metric} & \textbf{MK} & \textbf{MK+OV} & \textbf{MK+IA} & \textbf{MK+OV+IA} \\
  \midrule
  SP & 0.580 & 0.417 {\small {\color{red}$\downarrow$}0.163} & 0.438 {\small {\color{red}$\downarrow$}0.142} & 0.415 {\small {\color{red}$\downarrow$}0.165} \\
  PPL & 0.556 & 0.372 {\small {\color{red}$\downarrow$}0.184} & 0.415 {\small {\color{red}$\downarrow$}0.141} & 0.372 {\small {\color{red}$\downarrow$}0.184} \\
  MTE & 0.593 & 0.405 {\small {\color{red}$\downarrow$}0.188} & 0.445 {\small {\color{red}$\downarrow$}0.148} & 0.401 {\small {\color{red}$\downarrow$}0.192} \\
  MCSE & 0.587 & \underline{0.425 {\small {\color{red}$\downarrow$}0.162}} & 0.439 {\small {\color{red}$\downarrow$}0.149} & \textbf{0.426 {\small {\color{red}$\downarrow$}0.161}} \\
  MCNSE & 0.561 & 0.371 {\small {\color{red}$\downarrow$}0.190} & 0.400 {\small {\color{red}$\downarrow$}0.161} & 0.368 {\small {\color{red}$\downarrow$}0.193} \\
  \midrule
  Consistency & 0.511 & 0.386 {\small {\color{red}$\downarrow$}0.125} & 0.358 {\small {\color{red}$\downarrow$}0.153} & 0.320 {\small {\color{red}$\downarrow$}0.191} \\
  NumSemSets & 0.448 & 0.302 {\small {\color{red}$\downarrow$}0.146} & 0.336 {\small {\color{red}$\downarrow$}0.112} & 0.308 {\small {\color{red}$\downarrow$}0.140} \\
  EigValLaplacian & 0.580 & 0.420 {\small {\color{red}$\downarrow$}0.159} & 0.427 {\small {\color{red}$\downarrow$}0.153} & 0.392 {\small {\color{red}$\downarrow$}0.187} \\
  DegMat  & 0.555 & 0.394 {\small {\color{red}$\downarrow$}0.160} & 0.372 {\small {\color{red}$\downarrow$}0.183} & 0.336 {\small {\color{red}$\downarrow$}0.218} \\
  Eccentricity  & 0.553 & 0.415 {\small {\color{red}$\downarrow$}0.139} & 0.397 {\small {\color{red}$\downarrow$}0.157} & 0.380 {\small {\color{red}$\downarrow$}0.173} \\
  \midrule
  Semantic Entropy & 0.589 & \textbf{0.429 {\small {\color{red}$\downarrow$}0.161}} & 0.444 {\small {\color{red}$\downarrow$}0.145} & \underline{0.425 {\small {\color{red}$\downarrow$}0.164}} \\
  Semantic Density & \textbf{0.644} & 0.392 {\small {\color{red}$\downarrow$}0.252} & 0.427 {\small {\color{red}$\downarrow$}0.217} & 0.379 {\small {\color{red}$\downarrow$}0.265} \\
  CocoaMSP & 0.596 & 0.422 {\small {\color{red}$\downarrow$}0.174} & \underline{0.447 {\small {\color{red}$\downarrow$}0.148}} & 0.410 {\small {\color{red}$\downarrow$}0.186} \\
  CocoaPPL & 0.580 & 0.393 {\small {\color{red}$\downarrow$}0.187} & 0.431 {\small {\color{red}$\downarrow$}0.149} & 0.383 {\small {\color{red}$\downarrow$}0.197} \\
  CocoaMTE & 0.605 & 0.413 {\small {\color{red}$\downarrow$}0.192} & \textbf{0.453 {\small {\color{red}$\downarrow$}0.152}} & 0.400 {\small {\color{red}$\downarrow$}0.205} \\
  \midrule
  Verbalized2S & \underline{0.611} & 0.262 {\small {\color{red}$\downarrow$}0.349} & 0.422 {\small {\color{red}$\downarrow$}0.189} & 0.319 {\small {\color{red}$\downarrow$}0.292} \\
  \bottomrule
  \end{tabular}
}
\caption{PRR scores averaged across all five models. Inline arrows show drop from MK setting. \textbf{Bold} and \underline{underline} denote best and second-best per column.}
\label{tab:prr_avg_methods_rows}
\end{table*}


%% file: sections/discussion.tex
\begin{figure*}[t]
    \centering
    \begin{subfigure}[b]{0.24\textwidth}
        \includegraphics[width=\textwidth]{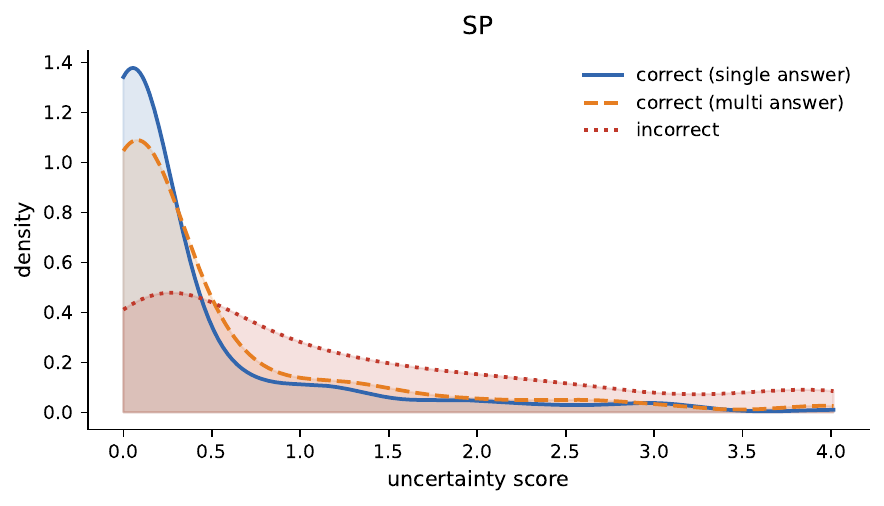}
        \caption{Sequence Probability}
    \end{subfigure}
    \hfill
    \begin{subfigure}[b]{0.24\textwidth}
        \includegraphics[width=\textwidth]{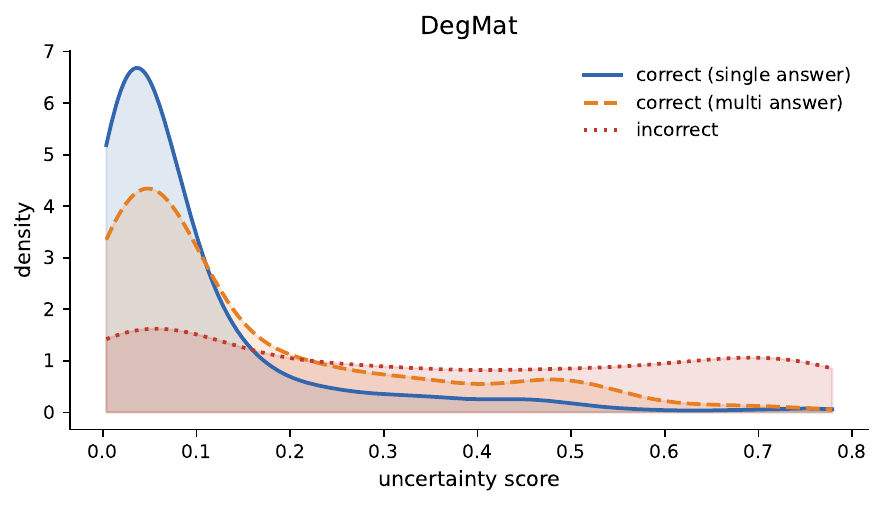}
        \caption{Degree Matrix}
    \end{subfigure}
    \hfill
    \begin{subfigure}[b]{0.24\textwidth}
        \includegraphics[width=\textwidth]{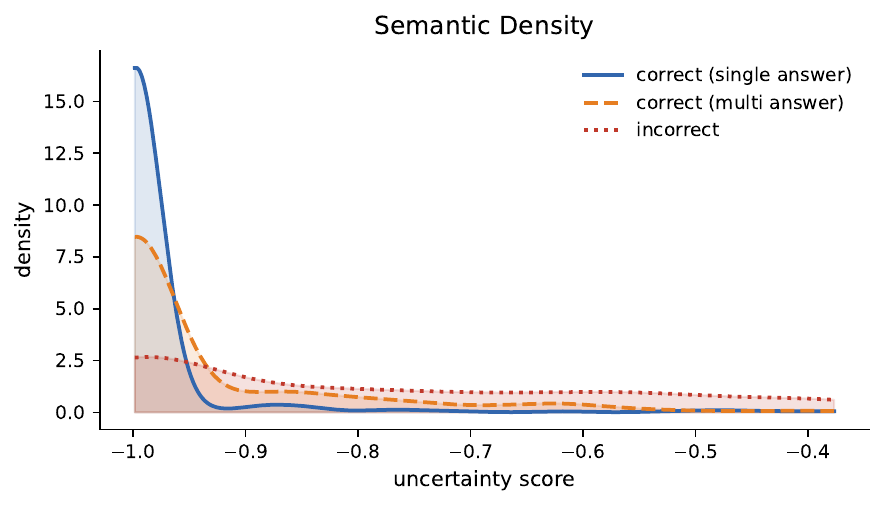}
        \caption{Semantic Density}
    \end{subfigure}
    \hfill
    \begin{subfigure}[b]{0.24\textwidth}
        \includegraphics[width=\textwidth]{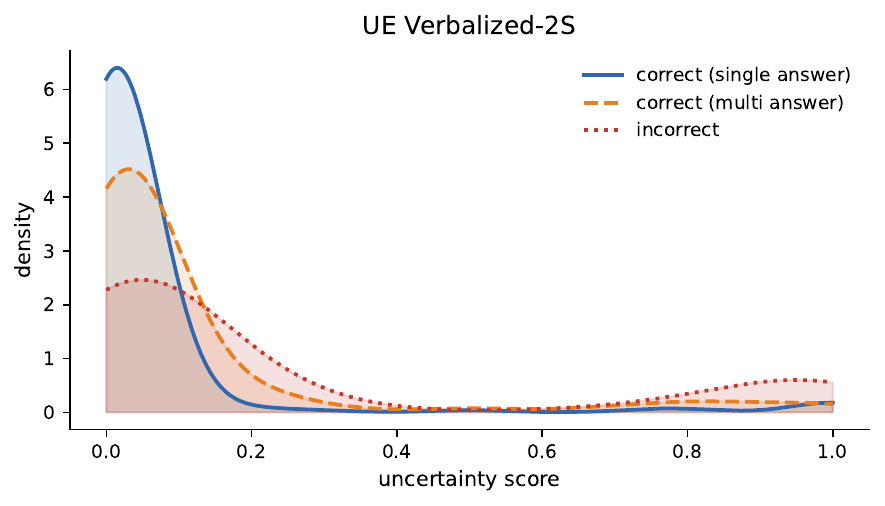}
        \caption{Verbalized-2S}
    \end{subfigure}
    \caption{Density of UQ scores for correct single- and multi-answer,
    and incorrect responses. Output variability shifts the correct distribution
    toward higher uncertainty, increasing overlap with incorrect answers. Model - \gemma.}
    \label{fig:ue_distributions}
\end{figure*}

\begin{figure*}[h!]
    \centering
    \begin{subfigure}[b]{0.24\textwidth}
        \includegraphics[width=\textwidth]{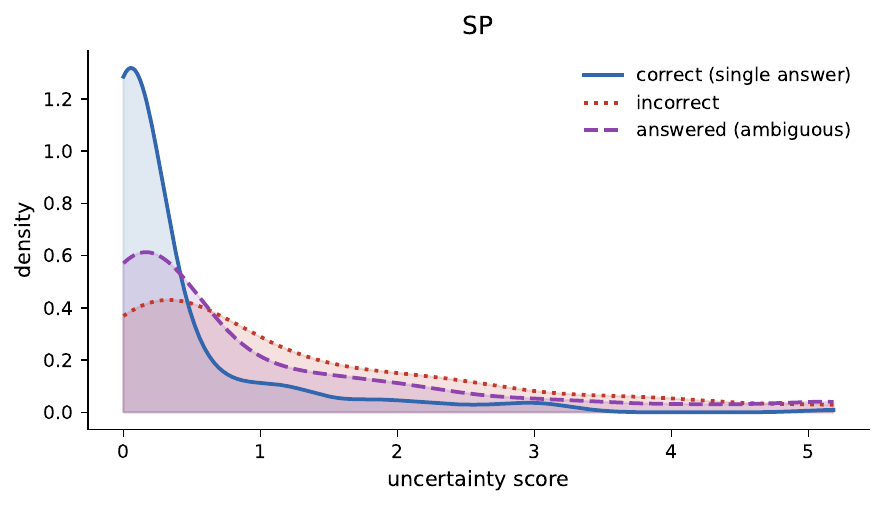}
        \caption{Sequence Probability}
    \end{subfigure}
    \hfill
    \begin{subfigure}[b]{0.24\textwidth}
        \includegraphics[width=\textwidth]{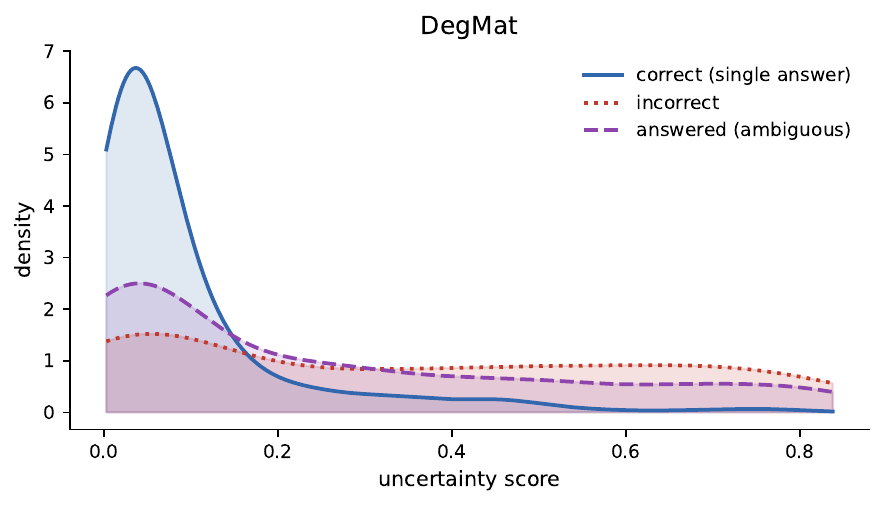}
        \caption{Degree Matrix}
    \end{subfigure}
    \hfill
    \begin{subfigure}[b]{0.24\textwidth}
        \includegraphics[width=\textwidth]{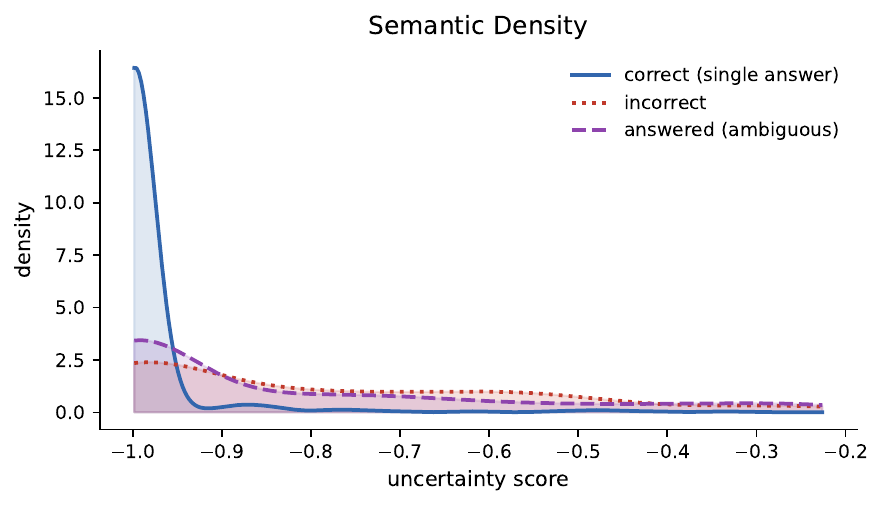}
        \caption{Semantic Density}
    \end{subfigure}
    \hfill
    \begin{subfigure}[b]{0.24\textwidth}
        \includegraphics[width=\textwidth]{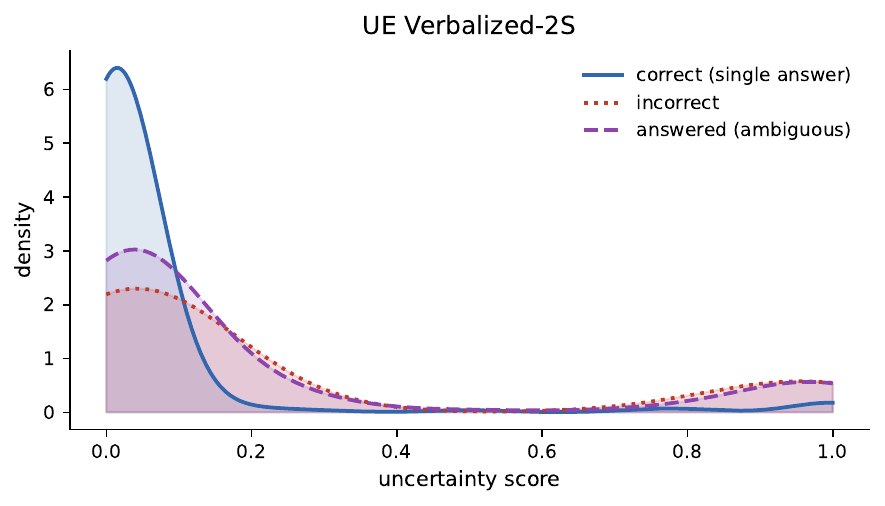}
        \caption{Verbalized-2S}
    \end{subfigure}
\caption{Density of UQ scores for correct, incorrect, and ambiguous-answered questions. 
Models commit to an arbitrary interpretation with apparent confidence: uncertainty for 
ambiguous inputs is lower than for incorrect answers, indicating overconfidence. Model - \gemma.}
    \label{fig:ue_distributions_ambiguity}
\end{figure*}

\section{Discussion}
\label{sec:discussion}
Our results reveal a clear pattern: existing UQ methods underperform when sources of uncertainty apart from model knowledge limitations are introduced. However, output variability and input ambiguity expose qualitatively different failure modes in UQ methods (Figure~\ref{fig:examples}). 

\textbf{Existing methods are most suited for well-specified, single-answer settings.} When model knowledge is the only source of uncertainty, UQ methods perform at their best. In this setting, both token-probability-based and consistency-based methods align well with generation quality. A model that knows the answer will produce a confident, and self-consistent response, while a model that does not know will assign low probabilities and generate a variety of answers. Because both signals point in the same direction, hybrid methods not only provide the best performance here, but also provide the most gain from uniting signals. Verbalized uncertainty is also competitive here, with performance improving as model size increases -  larger models appear better calibrated at introspecting on their own knowledge limitations, with \gptfour achieving a PRR of 0.855 against 0.427 and 0.437 of \llama and \gemma (see Appendix~\ref{sec:detailed_experimental_results}). 

\textbf{Output variability inflates uncertainty estimates.} When a question has multiple valid answers, UQ scores increase across all methods (Figure~\ref{fig:ue_distributions}, Appendix~\ref{sec:density_plots}). Token-probability-based methods are affected through how they read the model's distribution: when an input admits $N$ equally valid completions $\{y_1,\ldots,y_N\}$, probability mass is split across them, resulting in $p(y_i \mid x) = 1/N$ and thus $U_{\text{SP}}(y_i \mid x) = \log N$. Each correct completion therefore loses probability as valid alternatives multiply---a dilution the method misreads as a knowledge gap. 

Consistency-based methods fail similarly: a well-calibrated model produces diverse valid responses, so low sample agreement reflects legitimate diversity rather than genuine uncertainty (Figure~\ref{fig:examples}). Both families thus penalize the model for doing what the task requires, an effect that hybrid methods compound. Verbalized confidence degrades as well, but the extent varies by model.

\textbf{Input ambiguity deflates uncertainty estimates.}
When a question is underspecified, the opposite failure occurs: UQ scores remain lower than those for incorrectly answered questions, masking genuine ambiguity (Figure~\ref{fig:ue_distributions_ambiguity}, Appendix~\ref{sec:density_plots}). Models commit to an arbitrary interpretation and answer with apparent confidence rather than signaling that the question is unclear (Figure~\ref{fig:examples}). This is also supported by the near zero clarification rates (Table~\ref{tab:gen_quality}). Token-probability–based methods consequently assign high likelihood to generic responses compatible with several interpretations simultaneously, returning low uncertainty despite unresolved ambiguity in the input. Consistency-based methods are similarly deceived: a model that collapses onto a single interpretation can produce a highly consistent outputs, so agreement is high and uncertainty appears low. Hybrid methods again inherit failure modes from both. Verbalized methods for some models, can better detect this source of  uncertainty suggesting that sufficiently capable models can recognize underspecification even without explicit instruction to do so.

%% file: sections/conclusion.tex
\section{Conclusion and Future Work}

In this work, we built a human-curated dataset of 539 question triplets, where each entity comes with a single-answer variant, a multi-answer variant, and a variant that requires clarification. We then used it to compare 16 UQ methods across five LLMs under controlled conditions of model knowledge limitations, output variability, and input ambiguity.

Averaged across five models, all methods achieved their highest PRR on single-answer questions, where limited model knowledge was the only source of uncertainty. PRR fell under both output variability and input ambiguity, but in opposite ways. With several valid answers, probability and samples spread across them, so correct answers were scored like errors. Under input ambiguity, models committed to one interpretation and produced consistent answers, giving low uncertainty even when clarification was required.

Our results show that a single scalar confidence score is not enough for reliable LLM deployment: because the source of uncertainty determines the right system response, future UQ methods should disentangle these sources and report not only how uncertain the model is but also why. Extending the benchmark to multi-hop reasoning, long-form generation, and multilingual data, and evaluating reasoning models with white-box methods, would test whether the same failures recur.

%% file: sections/limitations.tex
\section*{Limitations}

Although our work offers insight into the nature of uncertainty sources, it has several limitations.


First, our analysis is aimed at question-answering tasks. Although QA provides a controlled setting for isolating and studying different uncertainty sources, uncertainty may manifest differently in other generation tasks such as summarization, translation, or creative writing. The dataset is also limited to English factoid questions, so our findings may not transfer to other languages or to question distributions that differ from this source.

Second, our main experiments focus on standard instruction-tuned models and do not include reasoning models. Reasoning models present a fundamentally different setting for UQ, as their extended chain-of-thought generation introduces distinct challenges -- such as reasoning path divergence and multi-step error propagation -- that are not present in instruction-tuned models~\citep{10.1145/3711896.3736569}. The UQ methods we evaluate were specifically designed for standard autoregressive generation, and instruction-tuned models remain widely deployed in production environments due to their lower inference cost and latency. In Appendix~\ref{app:reasoning_models}, we provide a limited evaluation of two reasoning models (GPT-5.4 and GPT-5.6), restricted to black-box methods since these models do not expose token probabilities; it shows the same performance degradation as additional sources of uncertainty are introduced. A more comprehensive study of UQ under reasoning model settings, including white-box methods, remains an important direction for future work.

Third, our dataset prioritizes controlled, human-validated examples and is therefore limited in scale. However, with 539 questions per uncertainty type, the dataset is sufficiently large to detect moderate effect sizes, and the observed differences in UQ method performance are consistent across all five models, suggesting our findings reflect genuine behavioral patterns. 

\section*{Ethics Statement}

\paragraph{Intended Use and Scope.}
This work is an evaluation study of UQ methods for LLMs. It does not introduce new model architectures, training procedures, or deployed systems, and aims to better understand the limitations of existing UQ approaches and support safer use of LLMs.

\paragraph{Data Collection and Usage.} Our dataset is constructed from publicly available question--answer sources and synthetic transformations. No personal data, sensitive attributes, or user-specific information are included.

The dataset is designed for research purposes only and does not target any specific demographic group. While we apply human validation, we acknowledge that source data and model-generated content may reflect existing societal biases.



\paragraph{Human Annotation.}
Human annotators (graduate students with NLP/ML background) were involved in validating and editing question–answer pairs. Annotation was limited to factual verification and linguistic clarity. Annotators were not exposed to harmful or sensitive content beyond standard factual QA material, and no personal data was handled during the process.

\paragraph{Risks and Potential Misuse.}
A potential risk is that improved or misinterpreted uncertainty estimates could increase user trust in incorrect model outputs. Our findings explicitly show that existing UQ methods can be misleading under output variability and input ambiguity. We therefore caution against using UQ scores as a sole indicator of correctness, especially in high-stakes applications. Uncertainty signals should be combined with additional safeguards such as human oversight, retrieval, or verification systems, especially under ambiguous inputs.

\paragraph{Limitations and Generalization.}
Our evaluation focuses on controlled question answering scenarios with three isolated sources of uncertainty. While this enables systematic analysis, it does not capture the full complexity of real-world interactions. Results may not directly transfer to other tasks (e.g., dialogue, long-form generation) or to proprietary, black-box, or tool-augmented systems, where uncertainty sources may interact or be resolved implicitly.

\paragraph{Broader Impact.}
By identifying failure modes of current UQ methods, this work aims to reduce overconfidence and improve reliability in LLM-based systems. We emphasize that uncertainty should be treated as a structured phenomenon and used to inform cautious and context-aware system behavior, rather than as a standalone signal for automated decision-making.

%% file: sections/appendix/app_dataset.tex
\section{Dataset Construction}
\label{sec:dataset_construction}
\raggedbottom
\interlinepenalty=10000

\subsection{Prompts}
\label{sec:prompts}

In this section, we provide the prompts used to generate the dataset.

\noindent\begin{minipage}{\columnwidth}
\begin{lstlisting}[language={},caption={Model prompt for Type 1: passes the original question, its valid answers, and the target answer.}]
You are given the following data:
Original question: {question}
List of possible answers:
{answers}
Target answer: {target_answer}
Rewrite the question so that it can only be answered correctly with the target answer.
Keep the rewritten question concise, factual, and natural-sounding - as if it appeared in a quiz or exam, not a textbook.
Use a distinctive clue (creator, time, feature, purpose, achievement, or location) that clearly singles out the target answer from the others.
Avoid unnecessary technical jargon or excessive qualifiers.
Return only this JSON:
{
  "target_answer": "<target answer>",
  "rewritten_question": "<natural single-answer question>"
}
\end{lstlisting}
\end{minipage}
\medskip

\noindent\begin{minipage}{\columnwidth}
\begin{lstlisting}[language={},caption={System prompt for Type 1: rewrite a multi-answer question so that only the selected target answer is correct.}]
You are a rewriting assistant that refines multi-answer questions into clear, natural single-answer questions.
You will be given:
1. An original question that may have multiple correct answers.
2. A list of possible answers.
3. A specific target answer from that list.
Your goal:
- Rewrite the question so it can only be answered correctly with the target answer.
- Keep the question fluent, natural, and concise (preferably under 25 words).
- The result should sound like a well-phrased quiz or interview question - not an encyclopedia entry.
Guidelines:
- Use discriminative details that uniquely identify the target (e.g., creator, origin, main feature, time period, famous example, key achievement).
- Add just enough context to exclude other listed answers, but avoid long or overly technical descriptions.
- Do NOT include the target answer itself in the question.
- Do NOT fabricate or distort facts.
- Maintain the same knowledge domain as the original question.
Tone and style:
- Favor natural English phrasing over formal or taxonomic style.
- Avoid stacked adjectives, nested clauses, or parentheses.
- Write as if asking a person in conversation: short, clear, and specific.
Output strictly in JSON:
{
  "target_answer": "<target answer>",
  "rewritten_question": "<natural single-answer question>"
}
\end{lstlisting}
\end{minipage}
\medskip

\noindent\begin{minipage}{\columnwidth}
\begin{lstlisting}[language={},caption={Model prompt for Type 2: passes the original question and its valid answers.}]
You are given the following data:
Original question: {{question}}
List of valid answers: {{answers}}
Rewrite the question so that:\n
- It can be answered with ANY of the listed answers,
- It requests only one response (e.g., "Name one..."),
- It keeps the same meaning and domain,
- It sounds fluent and natural.
Return only this JSON:\n
{
  "rewritten_question": "<single-response question>"
}
\end{lstlisting}
\end{minipage}
\medskip

\noindent\begin{minipage}{\columnwidth}
\begin{lstlisting}[language={},caption={System prompt for Type 2: rephrase a multi-answer question to request a single example that any valid answer satisfies.}]
You are a rewriting assistant that reformulates broad, multi-answer questions into natural, single-response questions.
Your task:
Rephrase a question that could have many valid answers so it instead expects one reasonable example - while keeping the same meaning and topic.
Examples:
- "Name countries in Europe" -> "What is a country in Europe?"
- "List programming languages" -> "What is an example of a programming language?"
- "Which animals live in the desert?" -> "Can you name an animal that lives in the desert?"
- "Give examples of renewable energy sources" -> "What is an example of a renewable energy source?"
Guidelines:
- The rewritten question must remain correct if answered by ANY of the provided valid answers.
- Make it sound natural and varied - not every question should start with "Name one...".
- Use phrasing such as "What is...", "Can you give...", "Which is...", "What's an example of...", etc.
- Keep it concise (under 25 words).
- Do NOT add new information or change the scope.
- Avoid phrases like "exactly one" or "only one".
Output strictly in JSON:
{
  "rewritten_question": "<single-response question>"
}
\end{lstlisting}
\end{minipage}
\medskip

\noindent\begin{minipage}{\columnwidth}
\begin{lstlisting}[language={},caption={Model prompt for Type 3: passes the question and its answer.}]
You are given the following data:
Question: {{question}}
Answer: {{target_answer}}
Create one natural-sounding ambiguous WH-question and its corresponding disambiguation question.
The ambiguous question should stay within a single domain (e.g., history, sports, geography, science) 
and use one or more ambiguity types: referential, temporal, locational, quantitative, causal, or contextual.
Make the phrasing fluent and human - avoid mechanical patterns like 'based in that city'.
Keep both questions concise (<=25 words) and clearly related.
Return ONLY this JSON:
{
"rewritten_question": "<ambiguous question>",
"target_answer": "<clarification question>"
}
\end{lstlisting}
\end{minipage}
\medskip

\noindent\begin{minipage}{\columnwidth}
\begin{lstlisting}[language={},caption={System prompt for Type 3: remove one key piece of information and write the clarification question that restores it.}]
You generate realistic pairs of questions that illustrate genuine ambiguity and its resolution.
Each output pair must include:
1) An ambiguous, natural-sounding WH-question (Who/What/When/Where/Which/How) that cannot be answered correctly without extra context.
2) A disambiguation question - the minimal, specific WH-question that restores the missing info.
Quality constraints:
- Keep both questions <=25 words, fluent, and conversational.
- Preserve the same domain/topic and answer type.
- Create ambiguity by removing exactly critical piece of info (referent, time, place, number, condition, or cause).
- Anchor the question with at least one concrete noun phrase from the input (e.g., competition name, country, team, era, instrument, field).
- Do not produce low-content templates (e.g., "Who said that in that year?", "Which team did that then?").
- Avoid stacked deictics ("that... that...") and vague placeholders ("that thing," "those people") unless there is a clear, single missing slot.
- Do not introduce new facts or entities.
- The disambiguation question must directly request the missing slot (e.g., "Which year do you mean?", "Which club?"), not meta-questions like "What do you mean?".
Output strictly as JSON:
{
  "rewritten_question": "<ambiguous question>",
  "target_answer": "<clarification question>"
}
Examples:
Input:
Question: Which club became the first to win the European Cup, Cup Winners' Cup, and UEFA Cup?
Answer: Juventus
Output:
{
  "rewritten_question": "Which club was the first to win all three of those tournaments in UEFA competition?",
  "target_answer": "Which three UEFA tournaments are you referring to?"
}
Input:
Question: What year was that building built?
Answer: 1979
Output:
{
  "rewritten_question": "When was that building constructed?",
  "target_answer": "Which building are you referring to?"
}
Input:
Question: Which Indo-Aryan language, primarily spoken in the Jammu region and added to the Eighth Schedule in 2003, is one of India's officially recognized languages?
Answer: Dogri
Output:
{
  "rewritten_question": "Which Indo-Aryan language, mainly spoken in that region, is officially recognized in India?",
  "target_answer": "Which region do you mean?"
}
\end{lstlisting}
\end{minipage}
\medskip


\subsection{Annotator Guidelines}
\label{sec:annotators}

In-house ML and NLP graduate students, all fluent in English, annotated the data in Google Sheets. Each generated question was marked as accepted, accepted with edits, or rejected, following the type-specific guidelines below.



\textbf{Type 1: Model Knowledge Limitations.} For this scenario, we construct unambiguous questions that have exactly one correct answer. The data is generated by GPT-5 model and verified by human annotators. Out of 592 questions, 497 were accepted, 67 were accepted with edits, and 28 questions were rejected.  

\paragraph{Annotation Guidelines (Type 1: Model Knowledge Limitations.)}
Annotators were instructed to ensure that each question--answer pair is clear, 
unambiguous, and factually correct. For each pair, annotators first verified 
that the question is grammatical, understandable, and admits only one valid 
interpretation and answer. Answer correctness was verified via Google Search. 
If a question was found to be incorrect or ambiguous, annotators were asked to 
edit the question (not the answer) minimally so that the given answer becomes 
correct and the question is well-posed. Each question was assigned one of three 
statuses: \textit{accepted} (correct as-is), \textit{accepted with edits} 
(correct after minimal edits), or \textit{rejected} (impossible to make 
single-answer or too unclear to fix).









  


\textbf{Type 2: Output Variability.} We construct unambiguous questions with multiple valid answers, any one of which is accepted as correct. The data is generated by GPT-5 model and verified by human annotators. Out of 592 questions, 543 were accepted, 48 were accepted with edits, and 1 question was rejected.

\paragraph{Annotation Guidelines (Type 2: Output Variability.)}
Annotators were instructed to ensure that each rewritten question is 
grammatically correct, fluent, and consistent with the original list of valid 
answers. Annotators verified that the question is natural-sounding and 
maintains a conversational tone (e.g., ``Name one...'', ``Give an example 
of...''), preserves the meaning of the original multi-answer question, and 
remains valid when answered by \textit{any} of the listed possible answers. 
Each question was assigned one of two statuses: \textit{accepted} (correct 
as-is) or \textit{accepted with edits} (correct after minimal edits).









\textbf{Type 3: Input Ambiguity.}
  We construct a set of ambiguous questions for scenario 3 that require the model to ask a clarification question. The initial data was generated using GPT-5 and subsequently verified by human annotators. Out of 592 questions, 348 were accepted as-is, 218 required edits, and 26 were rejected entirely.

  Overall, we found that the model performs poorly on this task. There is a clear trade-off when introducing instructions: providing too much freedom leads the model to remove excessive contextual information, often collapsing many questions into generic forms such as ``Who said that?'' or ``Who did that?''. Conversely, restricting the degree of information removal results in questions that retain too much context or in clarification questions that are confusing or refer to entities that no longer appear in the ambiguous version. As a result, many questions had to be substantially rewritten by human annotators to preserve as much of the question triplets as possible.

\paragraph{Annotation Guidelines (Type 3: Input Ambiguity.)}
Annotators were instructed to ensure that each rewritten question pair 
correctly introduces ambiguity by removing necessary context, and that the 
provided clarification question is natural and sufficient to resolve it.
For the ambiguous question, annotators verified that a key piece of information 
is omitted such that the question cannot be answered without clarification, 
while remaining grammatical, fluent, and within the same domain as the 
original. They also checked that no single interpretation is meaningfully more 
likely than the others. Trivial ambiguity (e.g., ``Who said that?'') was not 
accepted.

For the clarification question, annotators verified that it is a specific 
WH-question that directly resolves the ambiguity (e.g., ``Which year do you 
mean?'') without introducing new information, and that the original question 
becomes fully answerable once the clarification is provided. Both questions 
were required to be natural-sounding and concise ($\leq$25 words), avoiding 
mechanical phrasing such as ``based in that city''. Each pair was assigned one 
of three statuses: \textit{accepted} (meets all criteria as-is), 
\textit{accepted with edits} (acceptable after minimal edits), or 
\textit{rejected} (original question too short or generic to introduce 
meaningful ambiguity without degenerating into empty templates). Rejected 
questions were also removed from the final triplets.

\interlinepenalty=0

%% file: sections/appendix/app_gpt_judge.tex

\section{Prediction Rejection Ratio}
\label{app:prr}

  \begin{figure}[h!]
    \centering
    \includegraphics[width=0.9\linewidth]{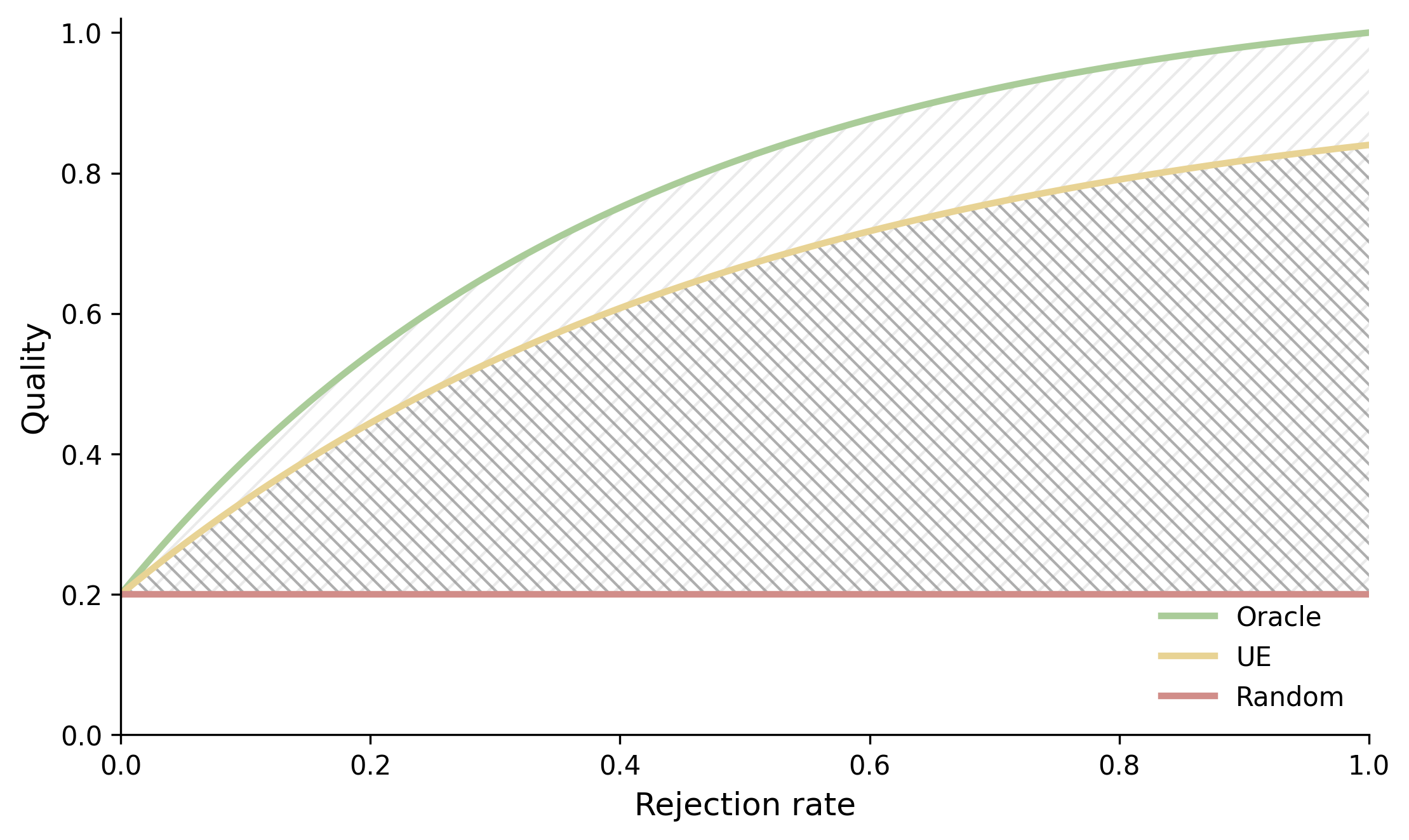}
    \caption{Prediction–Rejection Ratio (PRR) curve of non-rejected prediction quality vs. rejection rate. Oracle denotes the optimal strategy; Random and UQ correspond to random and uncertainty-based rejection.}
    \label{fig:prr}
  \end{figure}
PRR rejects predictions by decreasing uncertainty and tracks the mean quality of the rest (Figure~\ref{fig:prr}). With AUC, the area under this curve for uncertainty (unc), random (rnd), and oracle rejection,
\begin{equation}
    \text{PRR} = \frac{\text{AUC}_{\text{unc}}-\text{AUC}_{\text{rnd}}}{\text{AUC}_{\text{oracle}}-\text{AUC}_{\text{rnd}}}.
\label{eq:prr}
\end{equation}

\section{LLM-as-a-judge Validation}
\label{sec:judge}

Here we provide details on LLM-as-a-judge.



\begin{lstlisting}[language={},caption={LLM-as-a-judge prompt for Type 1.}]
You are grading the model answer to a SINGLE-ANSWER factual question.

Choose exactly one label for model answer:

- refused  
  The model explicitly refused, declined, or said it cannot answer.

- correct  
  The model clearly gives the correct answer (or an unambiguous synonym/alias of it),  
  and does NOT also present a different conflicting answer.

- incorrect  
  The model attempts an answer, but:
  - the correct answer (or any reasonable synonym) does NOT appear, OR
  - it presents multiple different candidates (e.g. "X or Y", "maybe X or Y")
    where not all candidates are the correct one.

Grading rules:
- Focus on the final answer content, ignore explanation and reasoning.
- Ignore differences in capitalization, punctuation, and minor wording changes.
- Extra descriptive context is fine as long as the main answer is correct and not contradicted.

Return ONLY the label.

Question:
{original_question}

Model answer:
{model_answer}

Correct answer:
{target_answer}
\end{lstlisting}

\newpage 

\begin{lstlisting}[language={},caption={LLM-as-a-judge prompt for Type 2.}]
You are grading the model answer to a MULTI-ANSWER factual question.

Select one of the following labels for model answer:
- refused: the model clearly refused or declined to answer.
- correct: every distinct answer item the model gives is part of the correct answer set
           (a subset is allowed; the model may give only  one of the correct items).
- incorrect: at least one distinct answer item is NOT in the correct answer set.

Grading rules:
- Treat the model's answer as a SET of distinct items it is clearly presenting as answers.
- The model may give only one item or multiple items.
- Missing items from the correct set are allowed and still count as "correct"
  as long as every given item is in the correct set.
- Ignore order and duplicates.
- Minor wording differences, capitalization, and simple aliases/synonyms are allowed
  (e.g. "Bob Dylan" vs "Robert Zimmerman").
- If the model just says it doesn't know or refuses, use "refused".

Return ONLY the label.

Question:
{original_question}

Model answer:
{model_answer}

Correct answer set:
{target_answer}
\end{lstlisting}

\begin{lstlisting}[language={},caption={LLM-as-a-judge prompt for Type 3.}]
You are grading the model's response to an AMBIGUOUS user input.

Choose exactly one label for the model response:

- clarification  
  The model asks for more information, points out the ambiguity, or refuses
  to answer. It does NOT commit to a specific factual answer.

- answered  
  The model chooses a specific interpretation of the input and attempts to answer it, commiting to a specific factual answer.

Grading rule:
- If the model both asks for clarification AND clearly gives a concrete answer,
  label it as "answered".

Return ONLY the label.

User input:
{original_question}

Model response:
{model_answer}
\end{lstlisting}

\begin{figure}[ht]
    \centering

    \begin{subfigure}[b]{0.32\textwidth}
        \centering
        \includegraphics[width=\linewidth]{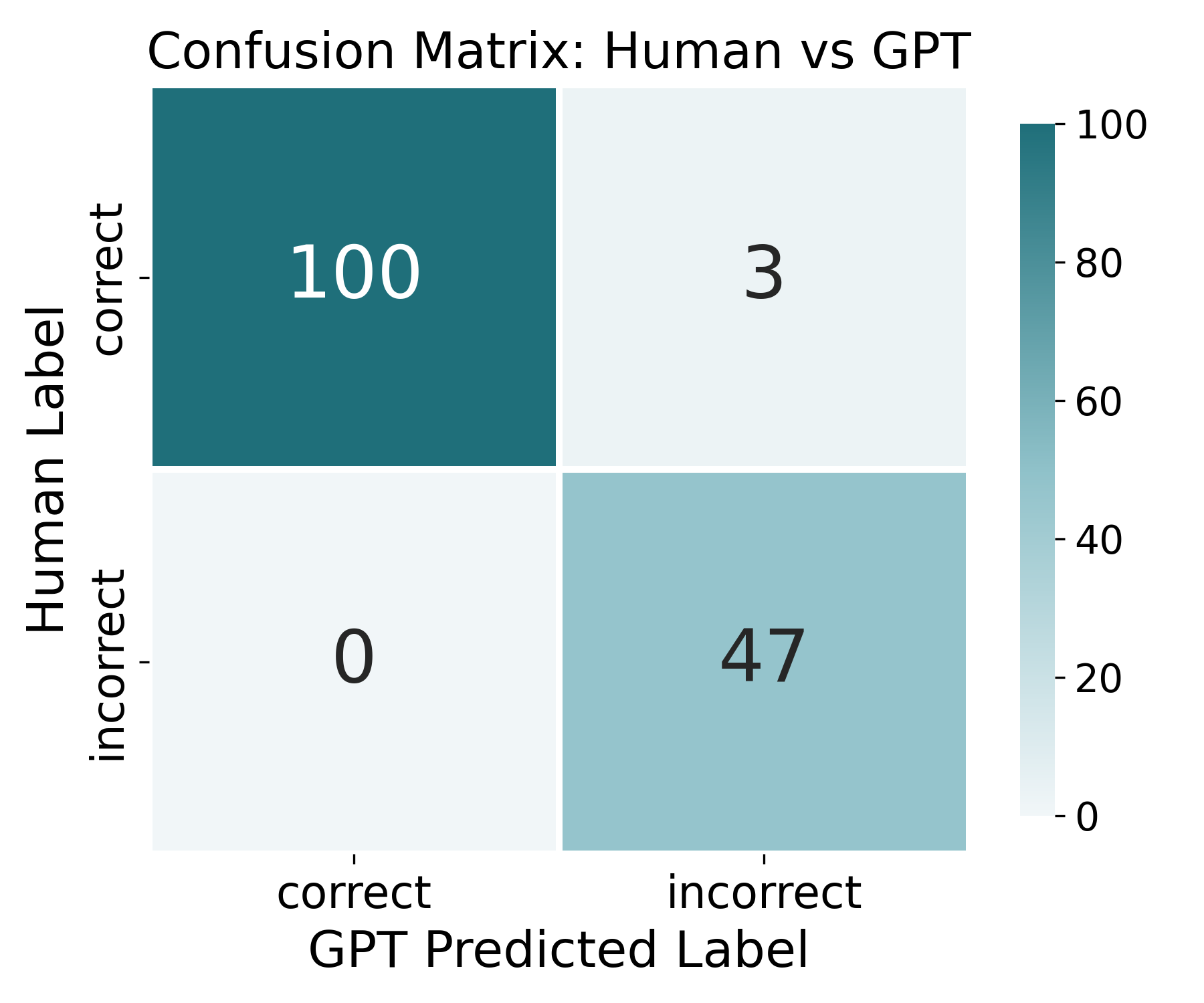}
        \caption{Single-answer}
    \end{subfigure}
    \hfill
    \begin{subfigure}[b]{0.32\textwidth}
        \centering
        \includegraphics[width=\linewidth]{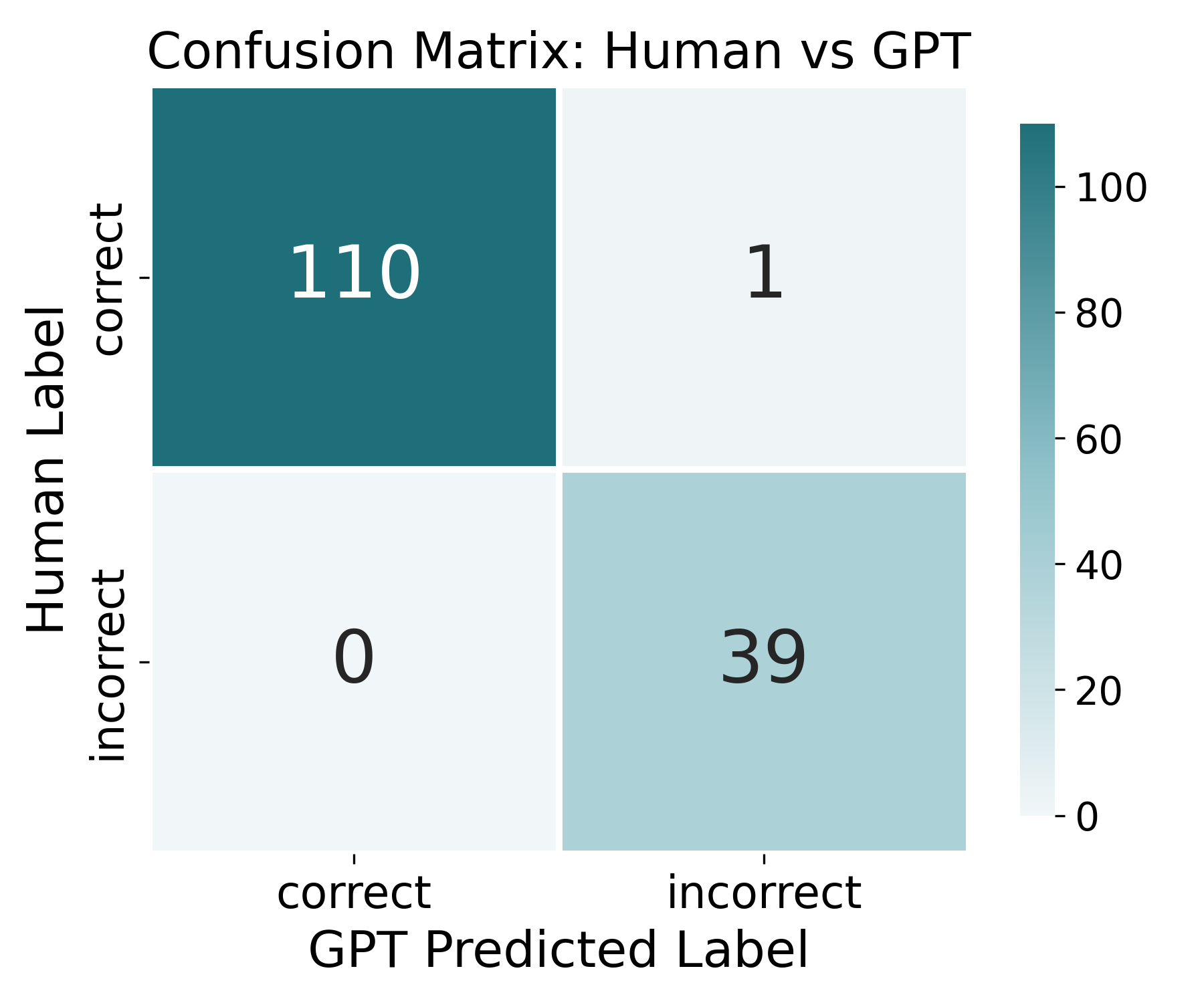}
        \caption{Multi-answer}
    \end{subfigure}
    \hfill
    \begin{subfigure}[b]{0.32\textwidth}
        \centering
        \includegraphics[width=\linewidth]{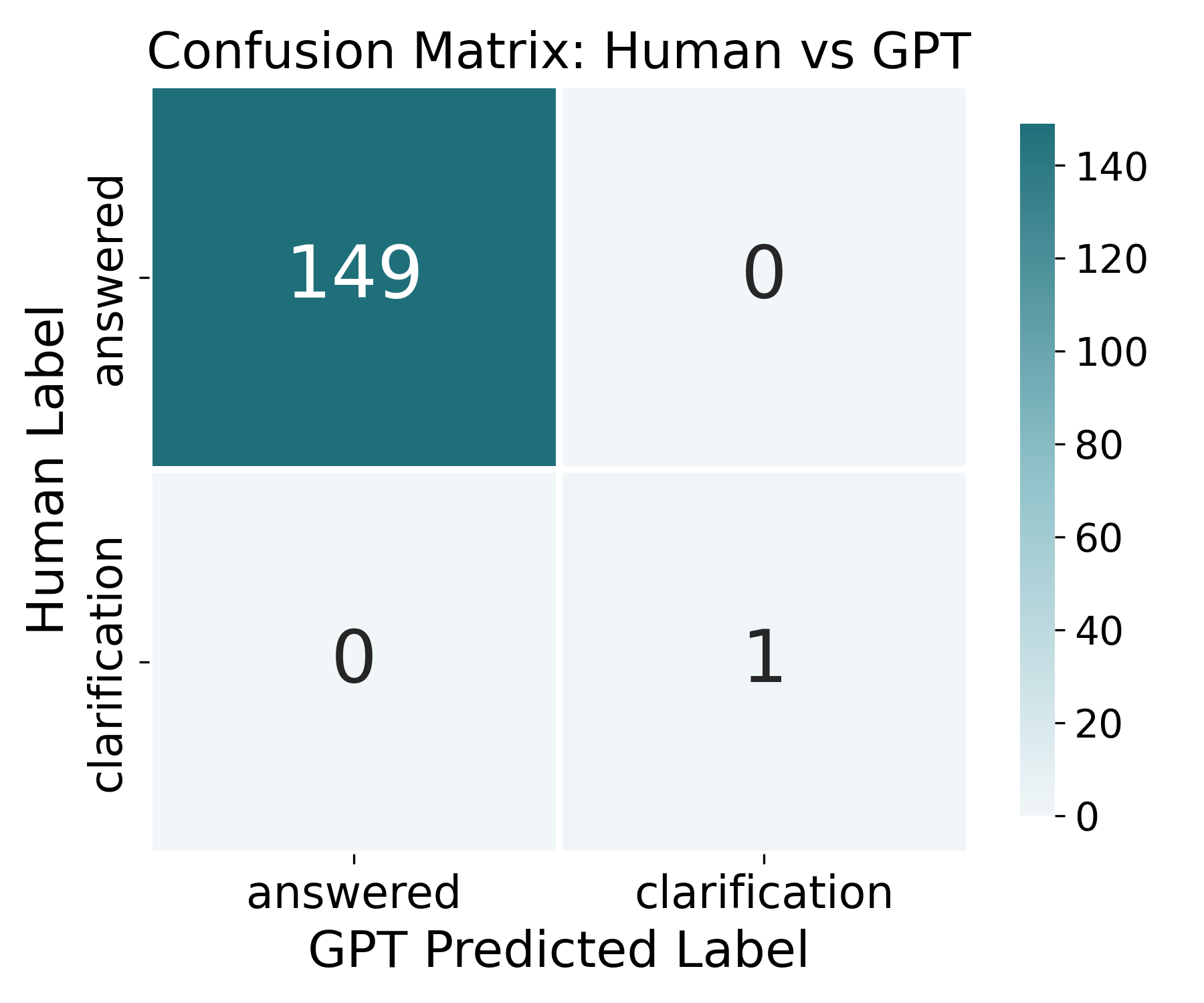}
        \caption{Ambiguous questions}
    \end{subfigure}

  \caption{Agreement between human annotations and LLM-as-a-judge across question types.}
  \label{fig:human_agreement}
\end{figure}

To validate the LLM-as-a-judge, we sampled 50 generations per question type from each of three models (\llama, \gemma, and \qwen), 150 per type in total, and asked a human annotator to label them following the same evaluation guidelines. As shown in Figure~\ref{fig:human_agreement}, the LLM-as-a-judge aligns closely with human judgments.

It achieves 98.0\%, 99.3\%, and 100\% agreement for single-answer, multi-answer, and ambiguous questions, respectively. Cohen's $\kappa$ is 0.95, 0.98, and 1.00 for the same three types (0.98 pooled over all 450 generations), confirming close agreement.

%% file: sections/appendix/app_models.tex

\section{Model Details}
\label{sec:models}
This appendix provides the generation details: the prompt used to obtain model answers and the exact model endpoints (Table~\ref{tab:hf_endpoints}).
\begin{table*}[ht!]
  \centering
  \begin{tabular}{ll}
  \toprule
  \textbf{Model} & \textbf{HuggingFace Endpoint} \\
  \midrule
  \llama    & \texttt{meta-llama/Llama-3.1-8B-Instruct} \\
  \gemma   & \texttt{google/gemma-3-12b-it} \\
  \qwen              & \texttt{Qwen/Qwen2.5-14B-Instruct} \\
  \llamaseventy  & \texttt{meta-llama/Llama-3.3-70B-Instruct} \\
  \bottomrule
  \end{tabular}
  \caption{HuggingFace inference endpoints used in our experiments.}
\label{tab:hf_endpoints}
\end{table*}

\begin{lstlisting}[language={},caption={Prompt used to generate answers to questions.}]
Answer the following question in ONE sentence. Answer as plain text on a single line. No line breaks or markdown. Provide only the answer - no introductions, explanations, or extra text. No words like "here is", "breakdown", "let's", "fascinating". Keep the answer short. Question: {text}
\end{lstlisting}



%% file: sections/appendix/app_methods.tex
\newpage

\section{Detailed Description of UQ Methods}
\label{sec:uq_methods}
In this section, we provide detailed description of Uncertainty Quantification methods evaluated in the study.

\subsection{Token-Probabilities Based Methods}

\noindent\textbf{Sequence Probability} is one of the simplest uncertainty measures, defining uncertainty as the negative log-likelihood of the output sequence conditioned on the input.

 \begin{equation}
    U_{\text{SP}}(\yv \mid \xv) = -\log p(\yv \mid \xv)
  \label{eq:msp}
  \end{equation}

\noindent\textbf{Perplexity}~\citep{fomicheva-etal-2020-unsupervised} normalizes the negative log-likelihood by the sequence length, yielding an average per-token uncertainty.
 
  \begin{equation}
    U_\mathrm{PPL}(\yv \mid \xv) = -\frac{1}{L} \log p(\yv \mid \xv)
  \label{eq:ppl}
  \end{equation}
  
\noindent\textbf{Mean Token Entropy}~\citep{fomicheva-etal-2020-unsupervised} computes the average entropy of the model’s predictive distribution at each decoding step.

  \begin{equation}
    U_{\HC_T}(\yv \mid \xv) = \frac{1}{L} \sum_{l = 1}^L \HC(y_l \mid \yv_{<l}, \xv)
  \label{eq:entropy}
  \end{equation}

\noindent\textbf{Monte Carlo (Normalized) Sequence Entropy}~\citep{malinin2021uncertainty} estimates sequence-level uncertainty by averaging the negative log-likelihood over multiple sampled output sequences. This estimate can be further normalized by the length of each sampled sequence.

\begin{equation}
    U_{\text{MCSE}}(\xv) = -\frac{1}{M} \sum_{i = 1}^M \log P(\yv^{(i)} \mid \xv)
  \label{eq:mcse}
  \end{equation}

\begin{equation}
    U_{\text{MCNSE}}(\xv) = -\frac{1}{M} \sum_{i = 1}^M \frac{1}{L^{(i)}} \log P(\yv^{(i)} \mid \xv)
  \label{eq:mcnse}
  \end{equation} 
  


\subsection{Consistency-Based Methods}

\noindent\textbf{Number of Semantic Sets (NumSemSets)}~\citep{lin2024generating} quantifies uncertainty as the number of distinct meanings in sampled outputs for a given input, where meanings are clustered using a natural language inference (NLI) model.

\noindent\textbf{Sum of Eigen Values of Graph Laplacian (EigValLaplacian)} constructs a similarity matrix over sampled outputs and deriving a continuous uncertainty score from the eigenvalues of its normalized graph Laplacian.

\begin{equation}
    U_{\text{EigV}}(\xv) = \sum_{i = 1}^M \max(0, 1 - \lambda_i(\xv))
  \end{equation}



\noindent\textbf{Degree Matrix}~\citep{lin2024generating} computes uncertainty from the degree matrix $D(\xv)$ which captures the total similarity of each sampled answer to all others. The diagonal matrix $D$ is defined with entries  $D_{ii} = \sum_{j = 1}^M s_{ij}$. Answers that agree with many others have large degrees, so high agreement yields low uncertainty. The final score is then defined as:

 \begin{equation}
    U_{\mathrm{DegMat}}(\xv) = 1 - \mathrm{trace}(D(\xv)) / M^2. 
    \label{eq:degmat}
  \end{equation}

\noindent\textbf{Eccentricity}~\citep{lin2024generating} computes uncertainty by embedding sampled answers into a low-dimensional space using the graph Laplacian, capturing their relational structure based on pairwise similarities. Let $L$ denote the graph Laplacian constructed from the $M \times M$ similarity matrix of the $M$ sampled answers. Let $\mathbf{u}_1, \ldots, \mathbf{u}_K$ be the eigenvectors corresponding to the $K$ smallest eigenvalues of $L$. The embedding of answer $y_j$ is defined as $\mathbf{v}_j = [u_{1,j}, \ldots, u_{K,j}]$, for $j = 1, \ldots, M$. Uncertainty is measured using \emph{eccentricity}, defined as the spread of the embeddings around their centroid $\bar{\mathbf{v}} = \frac{1}{M} \sum_{j=1}^{M} \mathbf{v}_j$:

\begin{equation}
    U_{\mathrm{Ecc}}(\xv) =
    \sqrt{\sum_{j=1}^{M} \left\| \mathbf{v}_j - \bar{\mathbf{v}} \right\|_2^2}
    \label{eq:eccentricity}
\end{equation}

\noindent\textbf{Consistency} is a method introduced by~\citet{lin2024generating} and further utilized by~\citet{vashurin2025cocoa}. It is computed using the similarity matrix $S$, where each entry $s_{ij}$ denotes the similarity between sequences $i$ and $j$. The uncertainty score for a sequence $i$ is defined as the average dissimilarity to all other sequences, excluding self-similarity:

\begin{equation}
    U_{\mathrm{Cons}}(x_i) =
    \frac{1}{M - 1} \sum_{\substack{j=1 \\ j \neq i}}^{M} \left( 1 - s_{ij} \right) 
    \label{eq:consistency}
\end{equation}

\subsection{Hybrid Methods}

\noindent\textbf{Semantic Entropy}~\citep{kuhn2023semantic} mitigates the effect of probability discrepancies among semantically equivalent generations. The method groups generated sequences $\yv^{(i)},\, i=1,\dots,M$ into semantically coherent clusters $\CC_k,\, k=1,\dots,K$ using bi-directional entailment. Sequence probabilities are aggregated within each cluster, and uncertainty is computed as:

\begin{equation}
    U_\mathrm{SE}(\xv) = -\sum_{k=1}^{K} \frac{|\CC_k|}{M} \log \hat{p}_k(\xv)
\end{equation}

where $\hat{p}_k(\xv) = \sum_{\yv \in \CC_k} p(\yv \mid \xv)$.



\noindent\textbf{Semantic Density}~\citep{qiu2024semantic} measures confidence as the probability mass concentrated near the greedy answer in \emph{semantic space}. Given $M$ sampled generations $\{\yv^{(i)}\}_{i=1}^M$, we compute a length-normalized probability
$\tilde{p}_i = \exp\!\bigl(\tfrac{1}{|\yv^{(i)}|}\log p(\yv^{(i)}\mid\xv)\bigr)$.
For each unique sample, a semantic distance to the greedy output is derived from an entailment-based similarity matrix as
$d_i = c_i + \tfrac{1}{2}n_i$ (contradiction $c_i$ and neutral $n_i$). A triangular kernel converts distance to a proximity weight
$k_i = \max(0, 1-d_i)$.
The semantic density is then the normalized weighted average
\begin{align}
\mathrm{SD}(\xv) &=
\frac{\tilde{p}_g + \sum_{i\in\mathcal{U}} \tilde{p}_i\, k_i}
     {\tilde{p}_g + \sum_{i\in\mathcal{U}} \tilde{p}_i}, \\
U_{\mathrm{SemDen}}(\xv) &= -\,\mathrm{SD}(\xv)
\label{eq:semantic_density}
\end{align}
where $\mathcal{U}$ indexes unique sampled texts and $\tilde{p}_g$ is the length-normalized probability of the greedy generation.


\begin{table*}[ht!]
    \centering
    \begin{tabular}{lccc}
        \toprule
        \textbf{Model} &
       \textbf{Single-answer} &
        \textbf{Multi-answer} &
        \textbf{Clarification rate} \\
        \midrule
        \llama &
        68.6\% (370/539) &
        76.3\% (411/539) &
        3.5\% (19/539) \\
        \gemma &
        69.0\% (372/539) &
        76.6\% (413/539) &
        0.0\% (0/539) \\
        \qwen &
        64.7\% (349/539) &
        70.5\% (380/539) &
        0.7\% (4/539) \\
         \llamaseventy &
        88.5\% (477/539) &
        84.6\% (456/539) &
        0.0\% (0/539) \\
        \gptfour &
        94.8\% (511/539) &
        91.8\% (495/539) &
        0.0\% (0/539) \\
        \bottomrule
    \end{tabular}
    \caption{Quality of generations across models. For single- and multi-answer questions we report accuracy; for ambiguous inputs we report the clarification rate (fraction of outputs that explicitly ask for clarification).}
    \label{tab:gen_quality}
  \end{table*}
  
\noindent\textbf{CoCoA}~\citep{vashurin2025cocoa} is a method that combines a token-probability-based uncertainty signal with semantic consistency. Given a target generation $\yv_*$ and $M$ sampled generations, the uncertainty estimate is defined as:

\begin{align}
    \widehat{U}_{\text{CoCoA}}(\yv_* \mid \xv)
    &= u(\yv_* \mid \xv)\cdot
    \frac{1}{M} \sum_{i=1}^{M} (1 - s_{*i}) \notag \\
    &= u(\yv_* \mid \xv)\cdot
    \widehat{U}_{\text{cons}}(\yv_* \mid \xv).
    \label{eq:cocoa_mc}
\end{align}
  

\noindent where $u(\yv_* \mid \xv)$ can be any base uncertainty measure. Following the original formulation, we consider sequence probability, perplexity, and mean token entropy as the underlying uncertainty signals.

\subsection{Verbalized Methods}


\noindent\textbf{Verbalized-2S}~\citep{tian-etal-2023-just} produces a verbalized confidence score for an already generated response, enabling direct comparison across methods on identical generations. The method re-feeds the original prompt and the generated answer to the model and asks it to explicitly estimate the probability that the answer is correct. In our experiments, we use the following prompt:

\begin{quote}
\textbf{The question is:}\\
\{question\}

\vspace{0.5em}
\textbf{YOUR GUESS:}\\
\{candidate\_answer\}

\vspace{0.5em}
\textbf{Now:}\\
Provide the probability that your guess is correct. Give \textbf{ONLY} the probability, no other words or explanation.

\vspace{0.5em}
\textbf{For example:}\\
Probability: \textless the probability between 0.0 and 1.0 that your guess is correct \textgreater
\end{quote}

%% file: sections/appendix/app_generation_details.tex


\section{Generation Details}
\label{sec:gen_details}
  In this section, we provide details about the lengths and quality of the generated outputs.

\begin{figure}[ht]
    \centering

    \begin{subfigure}[b]{0.32\textwidth}
        \centering
        \includegraphics[width=\linewidth]{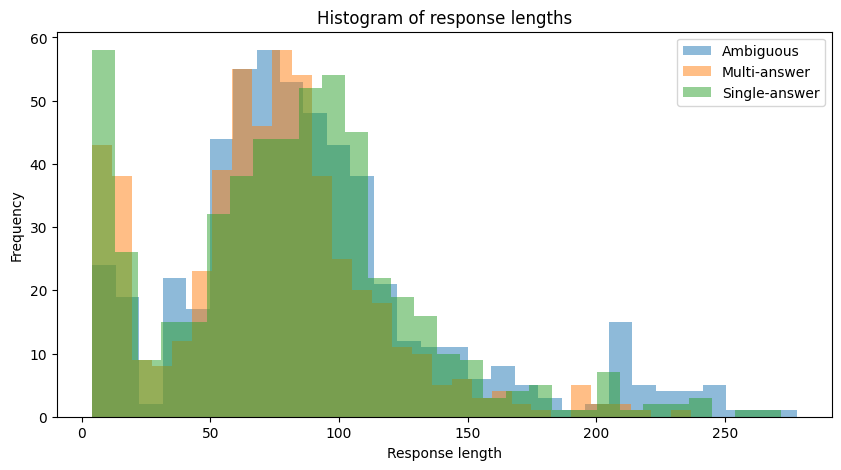}
        \caption{\llama}
        \label{fig:1a}
    \end{subfigure}
    \hfill
    \begin{subfigure}[b]{0.32\textwidth}
        \centering
        \includegraphics[width=\linewidth]{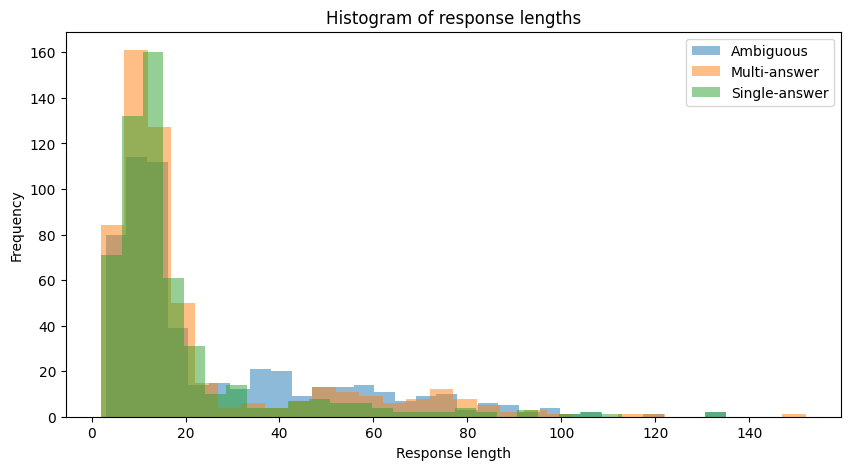}
        \caption{\gemma}
        \label{fig:1b}
    \end{subfigure}
    \hfill
    \begin{subfigure}[b]{0.32\textwidth}
        \centering
        \includegraphics[width=\linewidth]{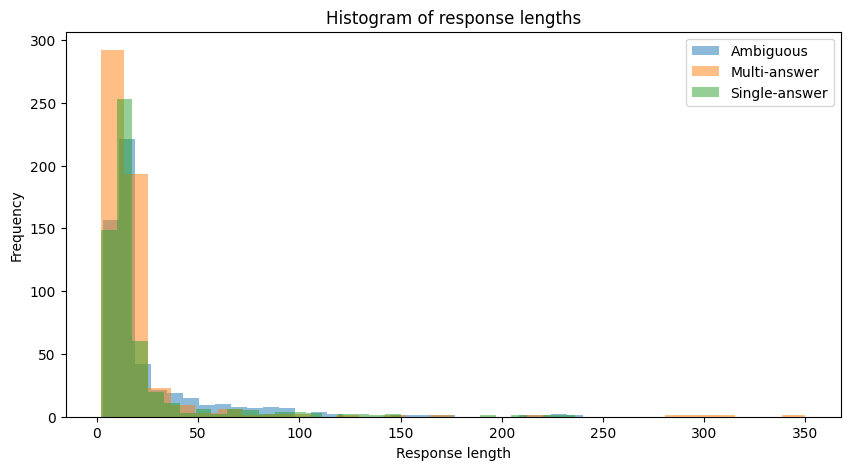}
        \caption{\qwen}
        \label{fig:1c}
    \end{subfigure}

    \caption{Distributions of generation lengths across models.}
    \label{fig:lengths}
  \end{figure}

\paragraph{Lengths.}
  Since generation length can influence uncertainty quantification, it was important to validate that the produced generations are similarly distributed in terms of length. As shown in Figure~\ref{fig:lengths}, all three scenario types exhibit comparable length distributions for each model, meaning that the differences in UQ distributions are not caused merely by differences in lengths. However, the distributions differ across models, with \llama producing the longest generations on average and \qwen yielding the shortest ones.

\paragraph{Quality.}
  As shown in Table~\ref{tab:gen_quality}, when using LLM-as-Judge as a quality assessor, we observe that none of the models reject answering in the single- or multi-answer scenarios when the prompt does not explicitly instruct them to do so. For ambiguous inputs, Gemma always provides an answer regardless of the ambiguity, whereas Qwen and LLaMA exhibit only very low clarification rates. Although such behavior can be improved through prompt engineering, our goal in this study is to evaluate uncertainty metrics under different sources of uncertainty; therefore, we intentionally focus on uninstructed generations.

%% file: sections/appendix/app_detailed_results.tex
\section{Detailed Experimental Results}
This appendix reports per-model PRR and AUROC scores for all 16 UQ methods; the averaged PRR in Table~\ref{tab:prr_avg_methods_rows} is computed from the PRR tables below.

\textbf{Model-Dependent Effect of Output Variability.} In the averaged results, all methods drop under MK+OV, but this is not uniform across models: for \llama and \qwen, 4 and 7 of the 16 methods, respectively, score slightly higher under MK+OV than under MK, whereas \gemma, and larger models, \llamaseventy, and \gptfour degrade for all methods. A potential explanation is that these smaller models may know only one valid answer to a multi-answer question and tend to return it consistently, which keeps token probabilities and consistency measures more stable; larger models may know several valid answers and spread probability across them, which these methods read as uncertainty. Importantly, this does not change the overall picture: once all sources of uncertainty are combined (MK+OV+IA), all methods degrade for every model, including these two.

\label{sec:detailed_experimental_results}


\subsection{PRR}

\input{sections/tables/prr}

\subsection{AUROC}

\input{sections/appendix/app_auroc}

%% file: sections/tables/prr.tex
Table~\ref{tab:prr_llama_8b_it} reports the PRR of all 16 uncertainty-quantification methods for \llama{} under the four settings MK, MK+OV, MK+IA and MK+OV+IA. The arrows give the absolute change from MK, and bold and underline mark the best and second-best method in each column. Adding IA lowers the PRR of every method, while adding OV alone has a mixed effect: 12 methods drop and four (MCSE, NumSemSets, Semantic Entropy and CocoaMTE) improve. Semantic Density is best under MK (0.741) and MK+IA (0.660), whereas MCSE leads under MK+OV (0.643) and EigValLaplacian under MK+OV+IA (0.546).

\begin{table}[H]
\centering
\scalebox{0.67}{
  \begin{tabular}{lcccc}
  \toprule
  \textbf{Metric} & \textbf{MK} & \textbf{MK+OV} & \textbf{MK+IA} & \textbf{MK+OV+IA} \\
  \midrule
  SP & 0.594 & 0.585 {\small {\color{red}$\downarrow$}0.009} & 0.503 {\small {\color{red}$\downarrow$}0.091} & 0.501 {\small {\color{red}$\downarrow$}0.093} \\
  PPL & 0.503 & 0.457 {\small {\color{red}$\downarrow$}0.046} & 0.434 {\small {\color{red}$\downarrow$}0.069} & 0.413 {\small {\color{red}$\downarrow$}0.090} \\
  MTE & 0.568 & 0.541 {\small {\color{red}$\downarrow$}0.027} & 0.492 {\small {\color{red}$\downarrow$}0.076} & 0.485 {\small {\color{red}$\downarrow$}0.083} \\
  MCSE & 0.608 & \textbf{0.643 {\small {\color{blue}$\uparrow$}0.035}} & 0.522 {\small {\color{red}$\downarrow$}0.086} & 0.543 {\small {\color{red}$\downarrow$}0.065} \\
  MCNSE & 0.524 & 0.464 {\small {\color{red}$\downarrow$}0.060} & 0.472 {\small {\color{red}$\downarrow$}0.052} & 0.443 {\small {\color{red}$\downarrow$}0.081} \\
  Consistency & 0.490 & 0.479 {\small {\color{red}$\downarrow$}0.012} & 0.419 {\small {\color{red}$\downarrow$}0.072} & 0.384 {\small {\color{red}$\downarrow$}0.107} \\
  NumSemSets & 0.528 & 0.584 {\small {\color{blue}$\uparrow$}0.056} & 0.467 {\small {\color{red}$\downarrow$}0.061} & 0.475 {\small {\color{red}$\downarrow$}0.054} \\
  EigValLaplacian  & \underline{0.680} & 0.614 {\small {\color{red}$\downarrow$}0.066} & \underline{0.591 {\small {\color{red}$\downarrow$}0.089}} & \textbf{0.546 {\small {\color{red}$\downarrow$}0.134}} \\
  DegMat & 0.639 & 0.595 {\small {\color{red}$\downarrow$}0.044} & 0.550 {\small {\color{red}$\downarrow$}0.088} & 0.520 {\small {\color{red}$\downarrow$}0.119} \\
  Eccentricity & 0.624 & 0.610 {\small {\color{red}$\downarrow$}0.014} & 0.534 {\small {\color{red}$\downarrow$}0.090} & 0.530 {\small {\color{red}$\downarrow$}0.094} \\
  Semantic Entropy & 0.611 & \underline{0.640 {\small {\color{blue}$\uparrow$}0.029}} & 0.530 {\small {\color{red}$\downarrow$}0.082} & \underline{0.544 {\small {\color{red}$\downarrow$}0.067}} \\
  Semantic Density & \textbf{0.741} & 0.528 {\small {\color{red}$\downarrow$}0.213} & \textbf{0.660 {\small {\color{red}$\downarrow$}0.081}} & 0.544 {\small {\color{red}$\downarrow$}0.197} \\
  CocoaMSP & 0.589 & 0.587 {\small {\color{red}$\downarrow$}0.002} & 0.506 {\small {\color{red}$\downarrow$}0.083} & 0.483 {\small {\color{red}$\downarrow$}0.106} \\
  CocoaPPL & 0.519 & 0.511 {\small {\color{red}$\downarrow$}0.008} & 0.452 {\small {\color{red}$\downarrow$}0.067} & 0.424 {\small {\color{red}$\downarrow$}0.095} \\
  CocoaMTE & 0.551 & 0.556 {\small {\color{blue}$\uparrow$}0.005} & 0.491 {\small {\color{red}$\downarrow$}0.060} & 0.465 {\small {\color{red}$\downarrow$}0.086} \\
  Verbalized2S & 0.427 & 0.201 {\small {\color{red}$\downarrow$}0.226} & 0.333 {\small {\color{red}$\downarrow$}0.094} & 0.236 {\small {\color{red}$\downarrow$}0.191} \\
  \bottomrule
  \end{tabular}
}
\caption{PRR scores for \llama. Inline arrows show drop from MK.}
\label{tab:prr_llama_8b_it}
\end{table}

Table~\ref{tab:prr_gemma_12b_it} gives the same breakdown for \gemma{}. Every method loses PRR in all three settings that extend MK, and for all 16 methods the largest drop occurs when OV and IA are combined. CocoaMTE is best under MK (0.623) and MK+OV (0.504), while Semantic Entropy leads under MK+IA (0.434) and MK+OV+IA (0.386).

\begin{table}[H]
\centering
\scalebox{0.67}{
  \begin{tabular}{lcccc}
  \toprule
  \textbf{Metric} & \textbf{MK} & \textbf{MK+OV} & \textbf{MK+IA} & \textbf{MK+OV+IA} \\
  \midrule
  SP & 0.564 & 0.461 {\small {\color{red}$\downarrow$}0.103} & 0.394 {\small {\color{red}$\downarrow$}0.170} & 0.358 {\small {\color{red}$\downarrow$}0.206} \\
  PPL & 0.544 & 0.456 {\small {\color{red}$\downarrow$}0.088} & 0.357 {\small {\color{red}$\downarrow$}0.187} & 0.325 {\small {\color{red}$\downarrow$}0.219} \\
  MTE & 0.587 & \underline{0.499 {\small {\color{red}$\downarrow$}0.089}} & 0.378 {\small {\color{red}$\downarrow$}0.209} & 0.346 {\small {\color{red}$\downarrow$}0.242} \\
  MCSE & 0.587 & 0.471 {\small {\color{red}$\downarrow$}0.117} & 0.413 {\small {\color{red}$\downarrow$}0.174} & 0.374 {\small {\color{red}$\downarrow$}0.213} \\
  MCNSE & 0.551 & 0.467 {\small {\color{red}$\downarrow$}0.085} & 0.355 {\small {\color{red}$\downarrow$}0.197} & 0.330 {\small {\color{red}$\downarrow$}0.221} \\
  Consistency & 0.519 & 0.424 {\small {\color{red}$\downarrow$}0.095} & 0.394 {\small {\color{red}$\downarrow$}0.125} & 0.328 {\small {\color{red}$\downarrow$}0.191} \\
  NumSemSets & 0.388 & 0.383 {\small {\color{red}$\downarrow$}0.005} & 0.374 {\small {\color{red}$\downarrow$}0.014} & 0.324 {\small {\color{red}$\downarrow$}0.063} \\
  EigValLaplacian & 0.531 & 0.473 {\small {\color{red}$\downarrow$}0.058} & 0.402 {\small {\color{red}$\downarrow$}0.129} & 0.361 {\small {\color{red}$\downarrow$}0.170} \\
  DegMat & 0.486 & 0.420 {\small {\color{red}$\downarrow$}0.066} & 0.343 {\small {\color{red}$\downarrow$}0.143} & 0.289 {\small {\color{red}$\downarrow$}0.197} \\
  Eccentricity  & 0.510 & 0.453 {\small {\color{red}$\downarrow$}0.056} & 0.391 {\small {\color{red}$\downarrow$}0.119} & 0.349 {\small {\color{red}$\downarrow$}0.160} \\
  Semantic Entropy & \underline{0.605} & 0.484 {\small {\color{red}$\downarrow$}0.120} & \textbf{0.434 {\small {\color{red}$\downarrow$}0.171}} & \textbf{0.386 {\small {\color{red}$\downarrow$}0.219}} \\
  Semantic Density & 0.589 & 0.401 {\small {\color{red}$\downarrow$}0.188} & 0.356 {\small {\color{red}$\downarrow$}0.234} & 0.320 {\small {\color{red}$\downarrow$}0.270} \\
  CocoaMSP & 0.603 & 0.483 {\small {\color{red}$\downarrow$}0.120} & 0.425 {\small {\color{red}$\downarrow$}0.178} & \underline{0.376 {\small {\color{red}$\downarrow$}0.227}} \\
  CocoaPPL & 0.591 & 0.480 {\small {\color{red}$\downarrow$}0.110} & 0.404 {\small {\color{red}$\downarrow$}0.187} & 0.356 {\small {\color{red}$\downarrow$}0.235} \\
  CocoaMTE & \textbf{0.623} & \textbf{0.504 {\small {\color{red}$\downarrow$}0.119}} & \underline{0.428 {\small {\color{red}$\downarrow$}0.195}} & 0.373 {\small {\color{red}$\downarrow$}0.250} \\
  Verbalized2S & 0.437 & 0.338 {\small {\color{red}$\downarrow$}0.099} & 0.415 {\small {\color{red}$\downarrow$}0.022} & 0.312 {\small {\color{red}$\downarrow$}0.125} \\
  \bottomrule
  \end{tabular}
}
\caption{PRR scores for \gemma. Inline arrows show drop from MK.}
\label{tab:prr_gemma_12b_it}
\end{table}

Table~\ref{tab:prr_qwen_14b_it} shows the results for \qwen{}. As for \llama{}, IA lowers the PRR of every method, whereas OV alone is mixed: nine methods drop and seven improve, by at most 0.064. CocoaMTE is best under MK (0.584) and MK+OV+IA (0.416), MTE leads under MK+OV (0.568), and Verbalized2S under MK+IA (0.528).

\begin{table}[H]
\centering
\scalebox{0.67}{
  \begin{tabular}{lcccc}
  \toprule
  \textbf{Metric} & \textbf{MK} & \textbf{MK+OV} & \textbf{MK+IA} & \textbf{MK+OV+IA} \\
  \midrule
  SP & 0.539 & 0.550 {\small {\color{blue}$\uparrow$}0.010} & 0.395 {\small {\color{red}$\downarrow$}0.145} & 0.394 {\small {\color{red}$\downarrow$}0.145} \\
  PPL & 0.544 & 0.524 {\small {\color{red}$\downarrow$}0.020} & 0.392 {\small {\color{red}$\downarrow$}0.152} & 0.371 {\small {\color{red}$\downarrow$}0.173} \\
  MTE & 0.570 & \textbf{0.568 {\small {\color{red}$\downarrow$}0.001}} & 0.423 {\small {\color{red}$\downarrow$}0.146} & \underline{0.408 {\small {\color{red}$\downarrow$}0.162}} \\
  MCSE & 0.520 & 0.531 {\small {\color{blue}$\uparrow$}0.010} & 0.381 {\small {\color{red}$\downarrow$}0.140} & 0.381 {\small {\color{red}$\downarrow$}0.140} \\
  MCNSE & 0.524 & 0.519 {\small {\color{red}$\downarrow$}0.005} & 0.380 {\small {\color{red}$\downarrow$}0.144} & 0.358 {\small {\color{red}$\downarrow$}0.166} \\
  Consistency & 0.467 & 0.498 {\small {\color{blue}$\uparrow$}0.031} & 0.327 {\small {\color{red}$\downarrow$}0.140} & 0.329 {\small {\color{red}$\downarrow$}0.138} \\
  NumSemSets & 0.448 & 0.354 {\small {\color{red}$\downarrow$}0.094} & 0.338 {\small {\color{red}$\downarrow$}0.111} & 0.303 {\small {\color{red}$\downarrow$}0.146} \\
  EigValLaplacian & 0.433 & 0.489 {\small {\color{blue}$\uparrow$}0.056} & 0.344 {\small {\color{red}$\downarrow$}0.090} & 0.342 {\small {\color{red}$\downarrow$}0.092} \\
  DegMat & 0.424 & 0.485 {\small {\color{blue}$\uparrow$}0.061} & 0.318 {\small {\color{red}$\downarrow$}0.106} & 0.312 {\small {\color{red}$\downarrow$}0.112} \\
  Eccentricity  & 0.414 & 0.478 {\small {\color{blue}$\uparrow$}0.064} & 0.301 {\small {\color{red}$\downarrow$}0.113} & 0.307 {\small {\color{red}$\downarrow$}0.107} \\
  Semantic Entropy & 0.533 & 0.530 {\small {\color{red}$\downarrow$}0.003} & 0.392 {\small {\color{red}$\downarrow$}0.141} & 0.386 {\small {\color{red}$\downarrow$}0.147} \\
  Semantic Density & 0.470 & 0.490 {\small {\color{blue}$\uparrow$}0.020} & 0.316 {\small {\color{red}$\downarrow$}0.155} & 0.337 {\small {\color{red}$\downarrow$}0.134} \\
  CocoaMSP & 0.561 & 0.548 {\small {\color{red}$\downarrow$}0.014} & 0.419 {\small {\color{red}$\downarrow$}0.143} & 0.403 {\small {\color{red}$\downarrow$}0.158} \\
  CocoaPPL & 0.560 & 0.536 {\small {\color{red}$\downarrow$}0.024} & 0.417 {\small {\color{red}$\downarrow$}0.144} & 0.390 {\small {\color{red}$\downarrow$}0.171} \\
  CocoaMTE & \textbf{0.584} & \underline{0.566 {\small {\color{red}$\downarrow$}0.018}} & \underline{0.442 {\small {\color{red}$\downarrow$}0.142}} & \textbf{0.416 {\small {\color{red}$\downarrow$}0.168}} \\
  Verbalized2S & \underline{0.572} & 0.400 {\small {\color{red}$\downarrow$}0.171} & \textbf{0.528 {\small {\color{red}$\downarrow$}0.044}} & 0.382 {\small {\color{red}$\downarrow$}0.190} \\
  \bottomrule
  \end{tabular}
}
\caption{PRR scores for \qwen. Inline arrows show drop from MK.}
\label{tab:prr_qwen_14b_it}
\end{table}

\noindent Table~\ref{tab:prr_llama_70b} covers \llamaseventy{}, where Semantic Density reaches 0.814 under MK but the degradation is also the sharpest: adding OV alone is the largest drop for all 16 methods (0.392 to 0.688), larger than adding IA or both. Semantic Density remains best under MK+OV, but at only 0.320; Verbalized2S leads under MK+IA (0.530) and MCSE under MK+OV+IA (0.447).

\begin{table}[H]
\centering
\scalebox{0.65}{
  \begin{tabular}{lcccc}
  \toprule
  \textbf{Metric} & \textbf{MK} & \textbf{MK+OV} & \textbf{MK+IA} & \textbf{MK+OV+IA} \\
  \midrule
  SP & 0.707 & 0.206 {\small {\color{red}$\downarrow$}0.501} & 0.476 {\small {\color{red}$\downarrow$}0.231} & 0.428 {\small {\color{red}$\downarrow$}0.279} \\
  PPL & 0.680 & 0.213 {\small {\color{red}$\downarrow$}0.467} & 0.483 {\small {\color{red}$\downarrow$}0.197} & 0.409 {\small {\color{red}$\downarrow$}0.271} \\
  MTE & 0.702 & 0.213 {\small {\color{red}$\downarrow$}0.489} & \underline{0.514 {\small {\color{red}$\downarrow$}0.188}} & 0.423 {\small {\color{red}$\downarrow$}0.279} \\
  MCSE & 0.751 & 0.240 {\small {\color{red}$\downarrow$}0.511} & 0.490 {\small {\color{red}$\downarrow$}0.261} & \textbf{0.447 {\small {\color{red}$\downarrow$}0.304}} \\
  MCNSE & 0.704 & 0.222 {\small {\color{red}$\downarrow$}0.482} & 0.446 {\small {\color{red}$\downarrow$}0.258} & 0.385 {\small {\color{red}$\downarrow$}0.318} \\
  Consistency & 0.682 & 0.290 {\small {\color{red}$\downarrow$}0.392} & 0.376 {\small {\color{red}$\downarrow$}0.305} & 0.311 {\small {\color{red}$\downarrow$}0.371} \\
  NumSemSets & 0.585 & 0.147 {\small {\color{red}$\downarrow$}0.437} & 0.279 {\small {\color{red}$\downarrow$}0.306} & 0.237 {\small {\color{red}$\downarrow$}0.348} \\
  EigValLaplacian & \underline{0.786} & 0.305 {\small {\color{red}$\downarrow$}0.481} & 0.443 {\small {\color{red}$\downarrow$}0.343} & 0.395 {\small {\color{red}$\downarrow$}0.391} \\
  DegMat & 0.757 & \underline{0.306 {\small {\color{red}$\downarrow$}0.451}} & 0.356 {\small {\color{red}$\downarrow$}0.401} & 0.314 {\small {\color{red}$\downarrow$}0.443} \\
  Eccentricity & 0.762 & 0.260 {\small {\color{red}$\downarrow$}0.502} & 0.414 {\small {\color{red}$\downarrow$}0.348} & 0.371 {\small {\color{red}$\downarrow$}0.391} \\
  Semantic Entropy & 0.743 & 0.237 {\small {\color{red}$\downarrow$}0.506} & 0.478 {\small {\color{red}$\downarrow$}0.265} & \underline{0.436 {\small {\color{red}$\downarrow$}0.308}} \\
  Semantic Density & \textbf{0.814} & \textbf{0.320 {\small {\color{red}$\downarrow$}0.494}} & 0.442 {\small {\color{red}$\downarrow$}0.371} & 0.378 {\small {\color{red}$\downarrow$}0.436} \\
  CocoaMSP & 0.707 & 0.197 {\small {\color{red}$\downarrow$}0.511} & 0.463 {\small {\color{red}$\downarrow$}0.244} & 0.406 {\small {\color{red}$\downarrow$}0.302} \\
  CocoaPPL & 0.701 & 0.208 {\small {\color{red}$\downarrow$}0.493} & 0.468 {\small {\color{red}$\downarrow$}0.232} & 0.395 {\small {\color{red}$\downarrow$}0.306} \\
  CocoaMTE & 0.718 & 0.210 {\small {\color{red}$\downarrow$}0.508} & 0.484 {\small {\color{red}$\downarrow$}0.234} & 0.400 {\small {\color{red}$\downarrow$}0.317} \\
  Verbalized2S & 0.764 & 0.076 {\small {\color{red}$\downarrow$}0.688} & \textbf{0.530 {\small {\color{red}$\downarrow$}0.233}} & 0.347 {\small {\color{red}$\downarrow$}0.417} \\
  \bottomrule
  \end{tabular}
}
\caption{PRR scores for \llamaseventy. Inline arrows show drop from MK.}
\label{tab:prr_llama_70b}
\end{table}

Table~\ref{tab:prr_gpt_41} reports \gptfour{}. As for \llamaseventy{}, adding OV alone gives the largest drop for all 16 methods. Verbalized2S is by far the best method under MK (0.855; next best is Semantic Density at 0.605) but falls to 0.295--0.320 in the other settings, where CocoaMSP leads under MK+OV (0.296) and MK+IA (0.425) and SP under MK+OV+IA (0.392).

\begin{table}[H]
\centering
\scalebox{0.65}{
  \begin{tabular}{lcccc}
  \toprule
  \textbf{Metric} & \textbf{MK} & \textbf{MK+OV} & \textbf{MK+IA} & \textbf{MK+OV+IA} \\
  \midrule
  SP & 0.495 & 0.284 {\small {\color{red}$\downarrow$}0.210} & \underline{0.424 {\small {\color{red}$\downarrow$}0.071}} & \textbf{0.392 {\small {\color{red}$\downarrow$}0.102}} \\
  PPL & 0.510 & 0.209 {\small {\color{red}$\downarrow$}0.302} & 0.409 {\small {\color{red}$\downarrow$}0.101} & 0.342 {\small {\color{red}$\downarrow$}0.168} \\
  MTE & 0.537 & 0.202 {\small {\color{red}$\downarrow$}0.335} & 0.419 {\small {\color{red}$\downarrow$}0.118} & 0.341 {\small {\color{red}$\downarrow$}0.196} \\
  MCSE & 0.470 & 0.242 {\small {\color{red}$\downarrow$}0.228} & 0.388 {\small {\color{red}$\downarrow$}0.082} & \underline{0.386 {\small {\color{red}$\downarrow$}0.084}} \\
  MCNSE & 0.503 & 0.185 {\small {\color{red}$\downarrow$}0.318} & 0.347 {\small {\color{red}$\downarrow$}0.156} & 0.325 {\small {\color{red}$\downarrow$}0.178} \\
  Consistency & 0.395 & 0.237 {\small {\color{red}$\downarrow$}0.157} & 0.273 {\small {\color{red}$\downarrow$}0.122} & 0.246 {\small {\color{red}$\downarrow$}0.149} \\
  NumSemSets & 0.291 & 0.041 {\small {\color{red}$\downarrow$}0.250} & 0.222 {\small {\color{red}$\downarrow$}0.069} & 0.203 {\small {\color{red}$\downarrow$}0.088} \\
  EigValLaplacian  & 0.468 & 0.220 {\small {\color{red}$\downarrow$}0.248} & 0.355 {\small {\color{red}$\downarrow$}0.113} & 0.317 {\small {\color{red}$\downarrow$}0.151} \\
  DegMat  & 0.468 & 0.167 {\small {\color{red}$\downarrow$}0.302} & 0.291 {\small {\color{red}$\downarrow$}0.177} & 0.246 {\small {\color{red}$\downarrow$}0.222} \\
  Eccentricity  & 0.457 & 0.272 {\small {\color{red}$\downarrow$}0.186} & 0.344 {\small {\color{red}$\downarrow$}0.114} & 0.345 {\small {\color{red}$\downarrow$}0.113} \\
  Semantic Entropy & 0.454 & 0.252 {\small {\color{red}$\downarrow$}0.202} & 0.389 {\small {\color{red}$\downarrow$}0.065} & 0.375 {\small {\color{red}$\downarrow$}0.079} \\
  Semantic Density & \underline{0.605} & 0.220 {\small {\color{red}$\downarrow$}0.385} & 0.362 {\small {\color{red}$\downarrow$}0.243} & 0.315 {\small {\color{red}$\downarrow$}0.290} \\
  CocoaMSP & 0.519 & \textbf{0.296 {\small {\color{red}$\downarrow$}0.223}} & \textbf{0.425 {\small {\color{red}$\downarrow$}0.094}} & 0.384 {\small {\color{red}$\downarrow$}0.135} \\
  CocoaPPL & 0.530 & 0.230 {\small {\color{red}$\downarrow$}0.299} & 0.415 {\small {\color{red}$\downarrow$}0.115} & 0.349 {\small {\color{red}$\downarrow$}0.180} \\
  CocoaMTE & 0.551 & 0.231 {\small {\color{red}$\downarrow$}0.319} & 0.421 {\small {\color{red}$\downarrow$}0.130} & 0.348 {\small {\color{red}$\downarrow$}0.202} \\
  Verbalized2S & \textbf{0.855} & \underline{0.295 {\small {\color{red}$\downarrow$}0.560}} & 0.305 {\small {\color{red}$\downarrow$}0.550} & 0.320 {\small {\color{red}$\downarrow$}0.535} \\
  \bottomrule
  \end{tabular}
}
\caption{PRR scores for \gptfour. Inline arrows show drop from MK.}
\label{tab:prr_gpt_41}
\end{table}

%% file: sections/appendix/app_auroc.tex
Table~\ref{tab:auroc_llama8bit} reports the AUROC of all 16 uncertainty-quantification methods for \llama{} under the four settings MK, MK+OV, MK+IA and MK+OV+IA (higher is better). The arrows give the absolute change from MK, and bold and underline mark the best and second-best method in each column. Adding IA lowers the AUROC of every method, while adding OV alone has a mixed effect: 13 methods drop and three (MCSE, NumSemSets and Semantic Entropy) improve. Semantic Density is best under MK (0.872) and MK+IA (0.834), whereas MCSE leads under MK+OV (0.814) and EigValLaplacian under MK+OV+IA (0.788).

\begin{table}[H]
\centering
\scalebox{0.65}{
  \begin{tabular}{lcccc}
  \toprule
  \textbf{Metric} & \textbf{MK} & \textbf{MK+OV} & \textbf{MK+IA} & \textbf{MK+OV+IA} \\
  \midrule
  SP & 0.796 & 0.789 {\small {\color{red}$\downarrow$}0.007} & 0.770 {\small {\color{red}$\downarrow$}0.026} & 0.764 {\small {\color{red}$\downarrow$}0.031} \\
  PPL & 0.752 & 0.720 {\small {\color{red}$\downarrow$}0.031} & 0.741 {\small {\color{red}$\downarrow$}0.011} & 0.722 {\small {\color{red}$\downarrow$}0.030} \\
  MTE & 0.781 & 0.761 {\small {\color{red}$\downarrow$}0.019} & 0.771 {\small {\color{red}$\downarrow$}0.010} & 0.757 {\small {\color{red}$\downarrow$}0.023} \\
  MCSE & 0.806 & \textbf{0.814 {\small {\color{blue}$\uparrow$}0.007}} & 0.780 {\small {\color{red}$\downarrow$}0.026} & \underline{0.785 {\small {\color{red}$\downarrow$}0.022}} \\
  MCNSE & 0.763 & 0.732 {\small {\color{red}$\downarrow$}0.031} & 0.752 {\small {\color{red}$\downarrow$}0.011} & 0.734 {\small {\color{red}$\downarrow$}0.029} \\
  Consistency & 0.747 & 0.727 {\small {\color{red}$\downarrow$}0.020} & 0.731 {\small {\color{red}$\downarrow$}0.016} & 0.705 {\small {\color{red}$\downarrow$}0.043} \\
  NumSemSets & 0.780 & 0.802 {\small {\color{blue}$\uparrow$}0.023} & 0.747 {\small {\color{red}$\downarrow$}0.032} & 0.747 {\small {\color{red}$\downarrow$}0.032} \\
  EigValLaplacian & \underline{0.842} & 0.807 {\small {\color{red}$\downarrow$}0.034} & \underline{0.810 {\small {\color{red}$\downarrow$}0.032}} & \textbf{0.788 {\small {\color{red}$\downarrow$}0.054}} \\
  DegMat & 0.823 & 0.794 {\small {\color{red}$\downarrow$}0.029} & 0.792 {\small {\color{red}$\downarrow$}0.031} & 0.776 {\small {\color{red}$\downarrow$}0.048} \\
  Eccentricity & 0.816 & 0.801 {\small {\color{red}$\downarrow$}0.015} & 0.784 {\small {\color{red}$\downarrow$}0.032} & 0.778 {\small {\color{red}$\downarrow$}0.038} \\
  Semantic Entropy & 0.807 & \underline{0.813 {\small {\color{blue}$\uparrow$}0.006}} & 0.782 {\small {\color{red}$\downarrow$}0.025} & 0.785 {\small {\color{red}$\downarrow$}0.022} \\
  Semantic Density & \textbf{0.872} & 0.763 {\small {\color{red}$\downarrow$}0.108} & \textbf{0.834 {\small {\color{red}$\downarrow$}0.037}} & 0.781 {\small {\color{red}$\downarrow$}0.091} \\
  CocoaMSP & 0.797 & 0.793 {\small {\color{red}$\downarrow$}0.005} & 0.770 {\small {\color{red}$\downarrow$}0.027} & 0.757 {\small {\color{red}$\downarrow$}0.040} \\
  CocoaPPL & 0.763 & 0.746 {\small {\color{red}$\downarrow$}0.017} & 0.748 {\small {\color{red}$\downarrow$}0.015} & 0.727 {\small {\color{red}$\downarrow$}0.036} \\
  CocoaMTE & 0.781 & 0.772 {\small {\color{red}$\downarrow$}0.009} & 0.766 {\small {\color{red}$\downarrow$}0.014} & 0.749 {\small {\color{red}$\downarrow$}0.032} \\
  Verbalized2S & 0.691 & 0.607 {\small {\color{red}$\downarrow$}0.084} & 0.651 {\small {\color{red}$\downarrow$}0.040} & 0.610 {\small {\color{red}$\downarrow$}0.081} \\
  \bottomrule
  \end{tabular}
}
\caption{AUROC scores for \llama.}
\label{tab:auroc_llama8bit}
\end{table}

Table~\ref{tab:auroc_gemma12bit} gives the same breakdown for \gemma{}. Every method loses AUROC in all three settings, most when OV and IA are combined. CocoaMTE is best under MK (0.815) and MK+OV (0.753); Semantic Entropy leads under MK+IA (0.734) and MK+OV+IA (0.707).

\begin{table}[H]
\centering
\scalebox{0.65}{
  \begin{tabular}{lcccc}
  \toprule
  \textbf{Metric} & \textbf{MK} & \textbf{MK+OV} & \textbf{MK+IA} & \textbf{MK+OV+IA} \\
  \midrule
  SP & 0.778 & 0.721 {\small {\color{red}$\downarrow$}0.057} & 0.710 {\small {\color{red}$\downarrow$}0.068} & 0.689 {\small {\color{red}$\downarrow$}0.089} \\
  PPL & 0.767 & 0.719 {\small {\color{red}$\downarrow$}0.048} & 0.691 {\small {\color{red}$\downarrow$}0.076} & 0.672 {\small {\color{red}$\downarrow$}0.096} \\
  MTE & 0.795 & \underline{0.749 {\small {\color{red}$\downarrow$}0.046}} & 0.707 {\small {\color{red}$\downarrow$}0.088} & 0.688 {\small {\color{red}$\downarrow$}0.107} \\
  MCSE & 0.795 & 0.732 {\small {\color{red}$\downarrow$}0.063} & 0.723 {\small {\color{red}$\downarrow$}0.071} & 0.701 {\small {\color{red}$\downarrow$}0.093} \\
  MCNSE & 0.776 & 0.730 {\small {\color{red}$\downarrow$}0.046} & 0.692 {\small {\color{red}$\downarrow$}0.084} & 0.678 {\small {\color{red}$\downarrow$}0.099} \\
  Consistency & 0.763 & 0.715 {\small {\color{red}$\downarrow$}0.048} & 0.698 {\small {\color{red}$\downarrow$}0.065} & 0.671 {\small {\color{red}$\downarrow$}0.092} \\
  NumSemSets & 0.743 & 0.714 {\small {\color{red}$\downarrow$}0.029} & 0.691 {\small {\color{red}$\downarrow$}0.052} & 0.676 {\small {\color{red}$\downarrow$}0.068} \\
  EigValLaplacian & 0.772 & 0.740 {\small {\color{red}$\downarrow$}0.033} & 0.712 {\small {\color{red}$\downarrow$}0.060} & 0.693 {\small {\color{red}$\downarrow$}0.079} \\
  DegMat & 0.753 & 0.715 {\small {\color{red}$\downarrow$}0.039} & 0.676 {\small {\color{red}$\downarrow$}0.077} & 0.651 {\small {\color{red}$\downarrow$}0.102} \\
  Eccentricity & 0.763 & 0.724 {\small {\color{red}$\downarrow$}0.039} & 0.706 {\small {\color{red}$\downarrow$}0.057} & 0.685 {\small {\color{red}$\downarrow$}0.078} \\
  Semantic Entropy & \underline{0.804} & 0.738 {\small {\color{red}$\downarrow$}0.066} & \textbf{0.734 {\small {\color{red}$\downarrow$}0.070}} & \textbf{0.707 {\small {\color{red}$\downarrow$}0.097}} \\
  Semantic Density & 0.799 & 0.700 {\small {\color{red}$\downarrow$}0.099} & 0.684 {\small {\color{red}$\downarrow$}0.115} & 0.668 {\small {\color{red}$\downarrow$}0.131} \\
  CocoaMSP & 0.802 & 0.738 {\small {\color{red}$\downarrow$}0.064} & 0.729 {\small {\color{red}$\downarrow$}0.073} & 0.702 {\small {\color{red}$\downarrow$}0.101} \\
  CocoaPPL & 0.797 & 0.738 {\small {\color{red}$\downarrow$}0.058} & 0.718 {\small {\color{red}$\downarrow$}0.078} & 0.692 {\small {\color{red}$\downarrow$}0.105} \\
  CocoaMTE & \textbf{0.815} & \textbf{0.753 {\small {\color{red}$\downarrow$}0.062}} & \underline{0.732 {\small {\color{red}$\downarrow$}0.083}} & \underline{0.703 {\small {\color{red}$\downarrow$}0.112}} \\
  Verbalized2S & 0.723 & 0.676 {\small {\color{red}$\downarrow$}0.046} & 0.711 {\small {\color{red}$\downarrow$}0.011} & 0.658 {\small {\color{red}$\downarrow$}0.065} \\
  \bottomrule
  \end{tabular}
}
\caption{AUROC scores for \gemma.}
\label{tab:auroc_gemma12bit}
\end{table}

Table~\ref{tab:auroc_qwen14bit} shows the results for \qwen{}. As for \llama{}, IA lowers the AUROC of every method, whereas OV alone is mixed: 12 methods drop and four (Consistency, EigValLaplacian, DegMat and Eccentricity) improve, by at most 0.029. CocoaMTE is best under MK (0.798), MK+OV (0.777) and MK+OV+IA (0.724), while Verbalized2S leads under MK+IA.

\begin{table}[H]
\centering
\scalebox{0.65}{
  \begin{tabular}{lcccc}
  \toprule
  \textbf{Metric} & \textbf{MK} & \textbf{MK+OV} & \textbf{MK+IA} & \textbf{MK+OV+IA} \\
  \midrule
  SP & 0.772 & 0.770 {\small {\color{red}$\downarrow$}0.002} & 0.717 {\small {\color{red}$\downarrow$}0.056} & 0.712 {\small {\color{red}$\downarrow$}0.060} \\
  PPL & 0.777 & 0.756 {\small {\color{red}$\downarrow$}0.021} & 0.716 {\small {\color{red}$\downarrow$}0.061} & 0.700 {\small {\color{red}$\downarrow$}0.076} \\
  MTE & 0.789 & 0.773 {\small {\color{red}$\downarrow$}0.015} & 0.731 {\small {\color{red}$\downarrow$}0.058} & 0.717 {\small {\color{red}$\downarrow$}0.071} \\
  MCSE & 0.761 & 0.760 {\small {\color{red}$\downarrow$}0.000} & 0.709 {\small {\color{red}$\downarrow$}0.052} & 0.705 {\small {\color{red}$\downarrow$}0.056} \\
  MCNSE & 0.765 & 0.757 {\small {\color{red}$\downarrow$}0.009} & 0.710 {\small {\color{red}$\downarrow$}0.055} & 0.694 {\small {\color{red}$\downarrow$}0.071} \\
  Consistency & 0.741 & 0.744 {\small {\color{blue}$\uparrow$}0.002} & 0.673 {\small {\color{red}$\downarrow$}0.068} & 0.672 {\small {\color{red}$\downarrow$}0.070} \\
  NumSemSets & 0.725 & 0.717 {\small {\color{red}$\downarrow$}0.008} & 0.669 {\small {\color{red}$\downarrow$}0.056} & 0.663 {\small {\color{red}$\downarrow$}0.062} \\
  EigValLaplacian & 0.726 & 0.744 {\small {\color{blue}$\uparrow$}0.018} & 0.678 {\small {\color{red}$\downarrow$}0.048} & 0.679 {\small {\color{red}$\downarrow$}0.047} \\
  DegMat & 0.720 & 0.738 {\small {\color{blue}$\uparrow$}0.018} & 0.666 {\small {\color{red}$\downarrow$}0.054} & 0.663 {\small {\color{red}$\downarrow$}0.057} \\
  Eccentricity & 0.709 & 0.738 {\small {\color{blue}$\uparrow$}0.029} & 0.655 {\small {\color{red}$\downarrow$}0.054} & 0.663 {\small {\color{red}$\downarrow$}0.046} \\
  Semantic Entropy & 0.767 & 0.761 {\small {\color{red}$\downarrow$}0.006} & 0.715 {\small {\color{red}$\downarrow$}0.053} & 0.708 {\small {\color{red}$\downarrow$}0.059} \\
  Semantic Density & 0.750 & 0.745 {\small {\color{red}$\downarrow$}0.005} & 0.669 {\small {\color{red}$\downarrow$}0.080} & 0.683 {\small {\color{red}$\downarrow$}0.066} \\
  CocoaMSP & 0.788 & \underline{0.773 {\small {\color{red}$\downarrow$}0.014}} & 0.732 {\small {\color{red}$\downarrow$}0.056} & \underline{0.720 {\small {\color{red}$\downarrow$}0.068}} \\
  CocoaPPL & 0.789 & 0.767 {\small {\color{red}$\downarrow$}0.022} & 0.732 {\small {\color{red}$\downarrow$}0.057} & 0.714 {\small {\color{red}$\downarrow$}0.075} \\
  CocoaMTE & \textbf{0.798} & \textbf{0.777 {\small {\color{red}$\downarrow$}0.021}} & \underline{0.741 {\small {\color{red}$\downarrow$}0.057}} & \textbf{0.724 {\small {\color{red}$\downarrow$}0.074}} \\
  Verbalized2S & \underline{0.793} & 0.701 {\small {\color{red}$\downarrow$}0.091} & \textbf{0.773 {\small {\color{red}$\downarrow$}0.020}} & 0.698 {\small {\color{red}$\downarrow$}0.095} \\
  \bottomrule
  \end{tabular}
}
\caption{AUROC scores for \qwen.}
\label{tab:auroc_qwen14bit}
\end{table}
 
Table~\ref{tab:auroc_llama_3.3_70b} covers \llamaseventy{}, where Semantic Density reaches 0.882 under MK, but the degradation is also the sharpest. Every method loses AUROC in all three settings that extend MK, and for all 16 methods adding IA costs the least (0.058 to 0.157), adding both OV and IA costs more (0.102 to 0.195), and adding OV alone costs the most (0.172 to 0.300). Verbalized2S is hit hardest, falling from 0.819 to 0.519 under MK+OV, and Semantic Density falls from 0.882 to 0.608. MCSE leads under MK+OV (0.649) and MK+OV+IA (0.716), though only by 0.001 and 0.002 over Semantic Entropy, whereas Semantic Density leads under MK+IA (0.769).

\begin{table}[H]
\centering
\scalebox{0.65}{
  \begin{tabular}{lcccc}
  \toprule
  \textbf{Metric} & \textbf{MK} & \textbf{MK+OV} & \textbf{MK+IA} & \textbf{MK+OV+IA} \\
  \midrule
  SP & 0.812 & 0.632 {\small {\color{red}$\downarrow$}0.180} & 0.732 {\small {\color{red}$\downarrow$}0.080} & 0.701 {\small {\color{red}$\downarrow$}0.111} \\
  PPL & 0.782 & 0.610 {\small {\color{red}$\downarrow$}0.172} & 0.722 {\small {\color{red}$\downarrow$}0.060} & 0.681 {\small {\color{red}$\downarrow$}0.102} \\
  MTE & 0.795 & 0.605 {\small {\color{red}$\downarrow$}0.190} & 0.737 {\small {\color{red}$\downarrow$}0.058} & 0.685 {\small {\color{red}$\downarrow$}0.110} \\
  MCSE & 0.846 & \textbf{0.649 {\small {\color{red}$\downarrow$}0.197}} & \underline{0.747 {\small {\color{red}$\downarrow$}0.099}} & \textbf{0.716 {\small {\color{red}$\downarrow$}0.130}} \\
  MCNSE & 0.806 & 0.613 {\small {\color{red}$\downarrow$}0.192} & 0.713 {\small {\color{red}$\downarrow$}0.092} & 0.676 {\small {\color{red}$\downarrow$}0.129} \\
  Consistency & 0.779 & 0.594 {\small {\color{red}$\downarrow$}0.185} & 0.688 {\small {\color{red}$\downarrow$}0.091} & 0.647 {\small {\color{red}$\downarrow$}0.132} \\
  NumSemSets & 0.822 & 0.591 {\small {\color{red}$\downarrow$}0.230} & 0.665 {\small {\color{red}$\downarrow$}0.157} & 0.641 {\small {\color{red}$\downarrow$}0.181} \\
  EigValLaplacian & \underline{0.869} & 0.606 {\small {\color{red}$\downarrow$}0.263} & 0.745 {\small {\color{red}$\downarrow$}0.124} & 0.690 {\small {\color{red}$\downarrow$}0.179} \\
  DegMat & 0.855 & 0.612 {\small {\color{red}$\downarrow$}0.244} & 0.727 {\small {\color{red}$\downarrow$}0.128} & 0.680 {\small {\color{red}$\downarrow$}0.175} \\
  Eccentricity & 0.844 & 0.630 {\small {\color{red}$\downarrow$}0.214} & 0.734 {\small {\color{red}$\downarrow$}0.109} & 0.697 {\small {\color{red}$\downarrow$}0.146} \\
  Semantic Entropy & 0.847 & \underline{0.648 {\small {\color{red}$\downarrow$}0.199}} & 0.745 {\small {\color{red}$\downarrow$}0.103} & \underline{0.714 {\small {\color{red}$\downarrow$}0.134}} \\
  Semantic Density & \textbf{0.882} & 0.608 {\small {\color{red}$\downarrow$}0.273} & \textbf{0.769 {\small {\color{red}$\downarrow$}0.113}} & 0.701 {\small {\color{red}$\downarrow$}0.181} \\
  CocoaMSP & 0.817 & 0.623 {\small {\color{red}$\downarrow$}0.194} & 0.728 {\small {\color{red}$\downarrow$}0.088} & 0.692 {\small {\color{red}$\downarrow$}0.125} \\
  CocoaPPL & 0.800 & 0.607 {\small {\color{red}$\downarrow$}0.194} & 0.724 {\small {\color{red}$\downarrow$}0.076} & 0.680 {\small {\color{red}$\downarrow$}0.120} \\
  CocoaMTE & 0.810 & 0.605 {\small {\color{red}$\downarrow$}0.205} & 0.732 {\small {\color{red}$\downarrow$}0.078} & 0.682 {\small {\color{red}$\downarrow$}0.128} \\
  Verbalized2S & 0.819 & 0.519 {\small {\color{red}$\downarrow$}0.300} & 0.740 {\small {\color{red}$\downarrow$}0.078} & 0.624 {\small {\color{red}$\downarrow$}0.195} \\
  \bottomrule
  \end{tabular}
}
\caption{AUROC scores for \llamaseventy.}
\label{tab:auroc_llama_3.3_70b}
\end{table}

Table~\ref{tab:auroc_gpt_4.1} reports \gptfour{}. As for \llamaseventy{}, adding OV alone gives the largest drop for all 16 methods. Verbalized2S is by far the best method under MK (0.921; next best is Semantic Density at 0.794) but falls to 0.625--0.662 in the other settings, where CocoaMSP leads under MK+OV (0.637) and MK+IA (0.730) and MCSE under MK+OV+IA (0.705). Under MK+IA, SP is the only method that improves over MK.

\begin{table}[H]
\centering
\scalebox{0.65}{
  \begin{tabular}{lcccc}
  \toprule
  \textbf{Metric} & \textbf{MK} & \textbf{MK+OV} & \textbf{MK+IA} & \textbf{MK+OV+IA} \\
  \midrule
  SP & 0.722 & 0.629 {\small {\color{red}$\downarrow$}0.093} & \underline{0.729 {\small {\color{blue}$\uparrow$}0.007}} & \underline{0.701 {\small {\color{red}$\downarrow$}0.021}} \\
  PPL & 0.727 & 0.598 {\small {\color{red}$\downarrow$}0.129} & 0.718 {\small {\color{red}$\downarrow$}0.008} & 0.672 {\small {\color{red}$\downarrow$}0.055} \\
  MTE & 0.742 & 0.596 {\small {\color{red}$\downarrow$}0.146} & 0.723 {\small {\color{red}$\downarrow$}0.019} & 0.671 {\small {\color{red}$\downarrow$}0.071} \\
  MCSE & 0.737 & 0.619 {\small {\color{red}$\downarrow$}0.118} & 0.716 {\small {\color{red}$\downarrow$}0.021} & \textbf{0.705 {\small {\color{red}$\downarrow$}0.032}} \\
  MCNSE & 0.748 & 0.593 {\small {\color{red}$\downarrow$}0.155} & 0.692 {\small {\color{red}$\downarrow$}0.057} & 0.667 {\small {\color{red}$\downarrow$}0.081} \\
  Consistency & 0.671 & 0.599 {\small {\color{red}$\downarrow$}0.072} & 0.634 {\small {\color{red}$\downarrow$}0.037} & 0.614 {\small {\color{red}$\downarrow$}0.057} \\
  NumSemSets & 0.620 & 0.552 {\small {\color{red}$\downarrow$}0.068} & 0.617 {\small {\color{red}$\downarrow$}0.003} & 0.606 {\small {\color{red}$\downarrow$}0.014} \\
  EigValLaplacian & 0.737 & 0.607 {\small {\color{red}$\downarrow$}0.130} & 0.692 {\small {\color{red}$\downarrow$}0.044} & 0.662 {\small {\color{red}$\downarrow$}0.074} \\
  DegMat & 0.730 & 0.574 {\small {\color{red}$\downarrow$}0.156} & 0.648 {\small {\color{red}$\downarrow$}0.082} & 0.618 {\small {\color{red}$\downarrow$}0.111} \\
  Eccentricity & 0.731 & \underline{0.630 {\small {\color{red}$\downarrow$}0.101}} & 0.686 {\small {\color{red}$\downarrow$}0.045} & 0.677 {\small {\color{red}$\downarrow$}0.054} \\
  Semantic Entropy & 0.731 & 0.620 {\small {\color{red}$\downarrow$}0.111} & 0.715 {\small {\color{red}$\downarrow$}0.015} & 0.699 {\small {\color{red}$\downarrow$}0.032} \\
  Semantic Density & \underline{0.794} & 0.601 {\small {\color{red}$\downarrow$}0.192} & 0.692 {\small {\color{red}$\downarrow$}0.102} & 0.661 {\small {\color{red}$\downarrow$}0.133} \\
  CocoaMSP & 0.736 & \textbf{0.637 {\small {\color{red}$\downarrow$}0.099}} & \textbf{0.730 {\small {\color{red}$\downarrow$}0.006}} & 0.697 {\small {\color{red}$\downarrow$}0.039} \\
  CocoaPPL & 0.737 & 0.610 {\small {\color{red}$\downarrow$}0.127} & 0.724 {\small {\color{red}$\downarrow$}0.013} & 0.678 {\small {\color{red}$\downarrow$}0.059} \\
  CocoaMTE & 0.749 & 0.610 {\small {\color{red}$\downarrow$}0.139} & 0.728 {\small {\color{red}$\downarrow$}0.021} & 0.678 {\small {\color{red}$\downarrow$}0.071} \\
  Verbalized2S & \textbf{0.921} & 0.625 {\small {\color{red}$\downarrow$}0.296} & 0.662 {\small {\color{red}$\downarrow$}0.260} & 0.656 {\small {\color{red}$\downarrow$}0.266} \\
  \bottomrule
  \end{tabular}
}
\caption{AUROC scores for \gptfour.}
\label{tab:auroc_gpt_4.1}
\end{table}

%% file: sections/appendix/app_density_plots.tex
\section{Uncertainty Score Distributions}
\label{sec:density_plots}

\begin{figure}[H]
    \centering
    \begin{subfigure}[b]{0.48\columnwidth}
        \includegraphics[width=\textwidth]{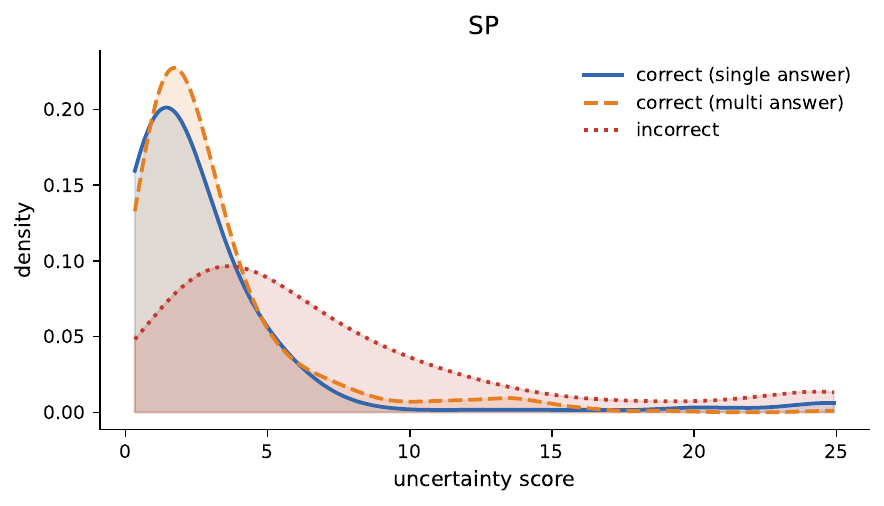}
        \caption{Sequence Probability}
    \end{subfigure}
    \hfill
    \begin{subfigure}[b]{0.48\columnwidth}
        \includegraphics[width=\textwidth]{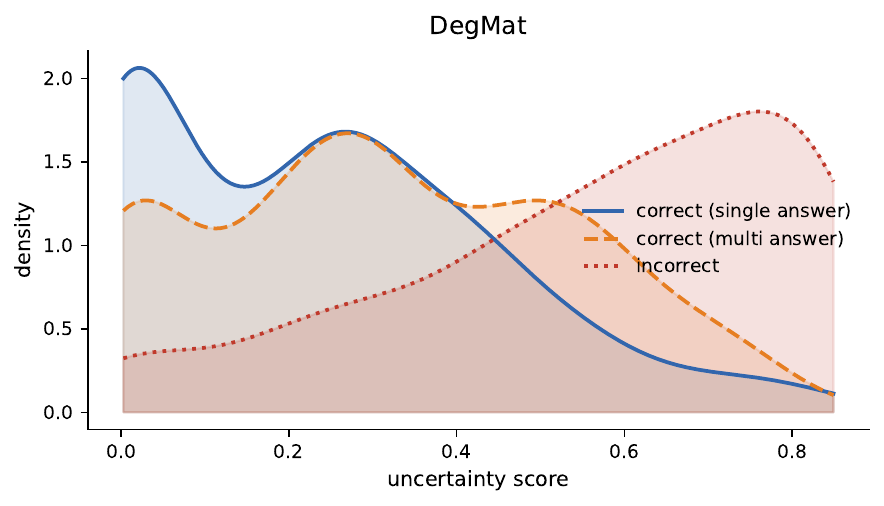}
        \caption{Degree Matrix}
    \end{subfigure}

    \vspace{0.1em}

    \begin{subfigure}[b]{0.48\columnwidth}
        \includegraphics[width=\textwidth]{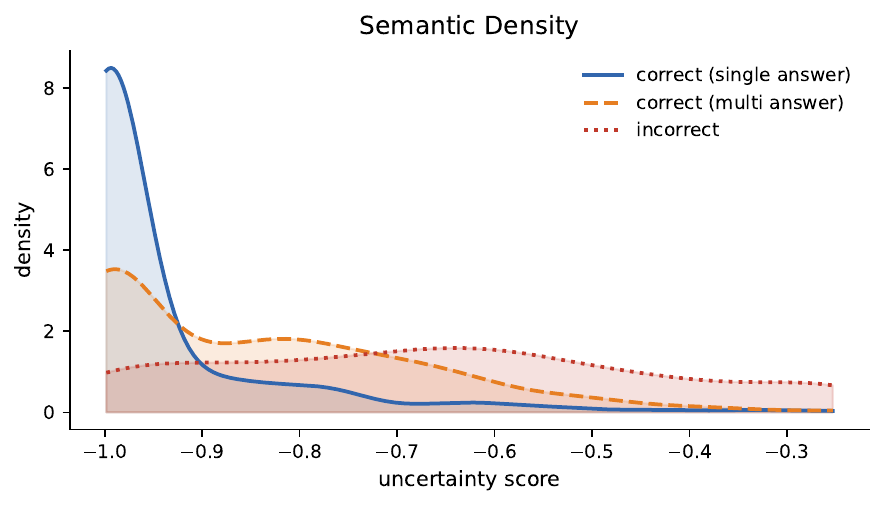}
        \caption{Semantic Density}
    \end{subfigure}
    \hfill
    \begin{subfigure}[b]{0.48\columnwidth}
        \includegraphics[width=\textwidth]{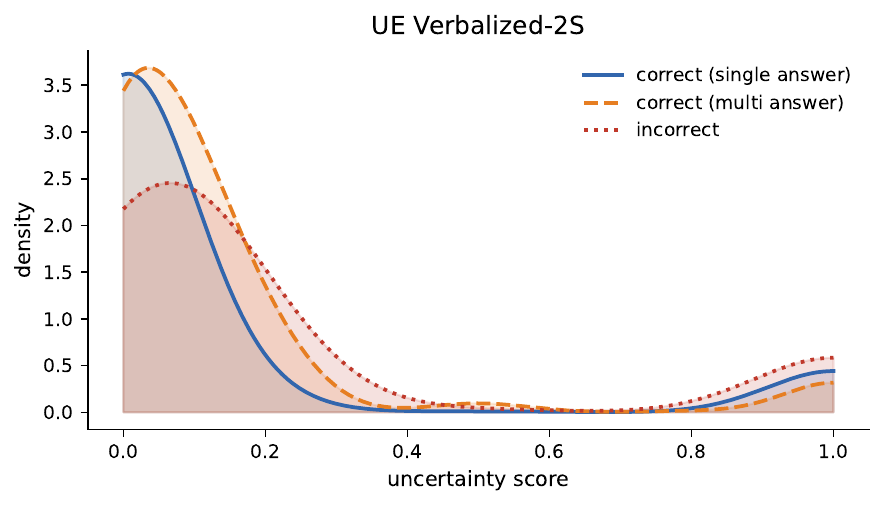}
        \caption{Verbalized-2S}
    \end{subfigure}
    \caption{Density of uncertainty scores for correct (single answer), correct (multi answer),
    and incorrect (pooled) responses. Model - \llama.}
    \label{fig:ue_distributions_llama}
\end{figure}

\begin{figure}[H]
    \centering
    \begin{subfigure}[b]{0.48\columnwidth}
        \includegraphics[width=\textwidth]{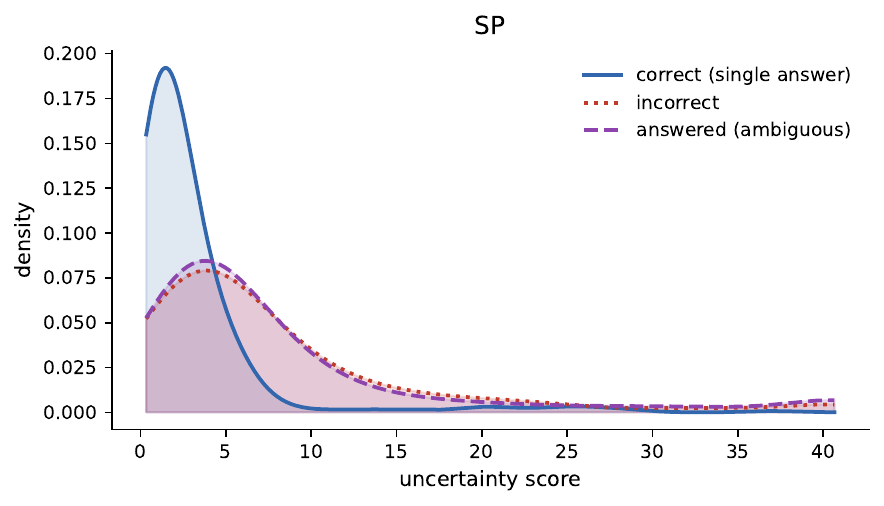}
        \caption{Sequence Probability}
    \end{subfigure}
    \hfill
    \begin{subfigure}[b]{0.48\columnwidth}
        \includegraphics[width=\textwidth]{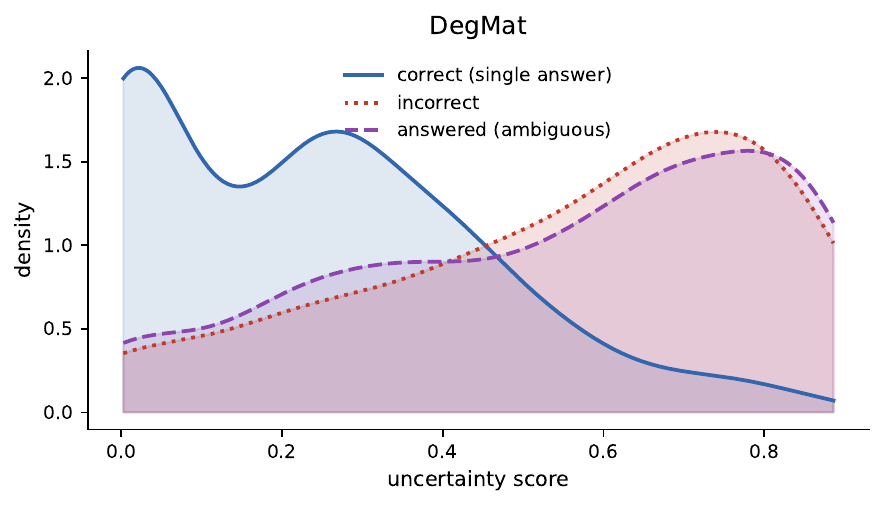}
        \caption{Degree Matrix}
    \end{subfigure}

    \vspace{0.1em}

    \begin{subfigure}[b]{0.48\columnwidth}
        \includegraphics[width=\textwidth]{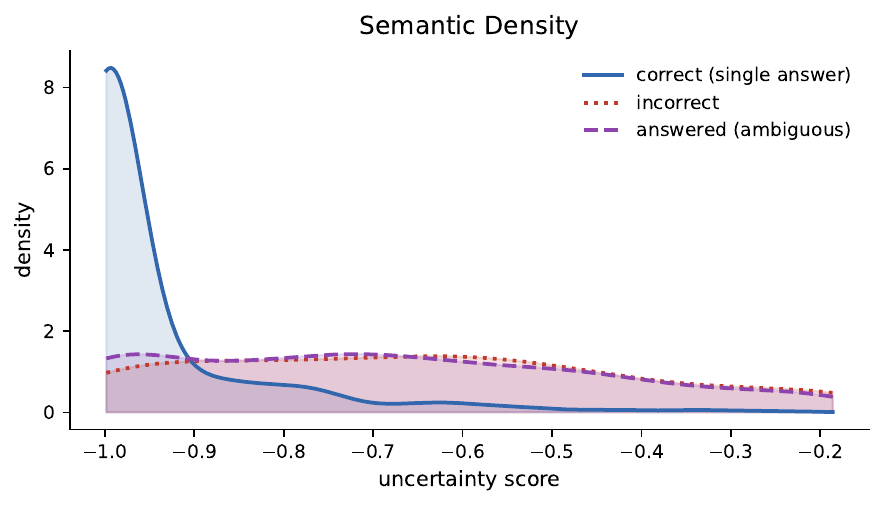}
        \caption{Semantic Density}
    \end{subfigure}
    \hfill
    \begin{subfigure}[b]{0.48\columnwidth}
        \includegraphics[width=\textwidth]{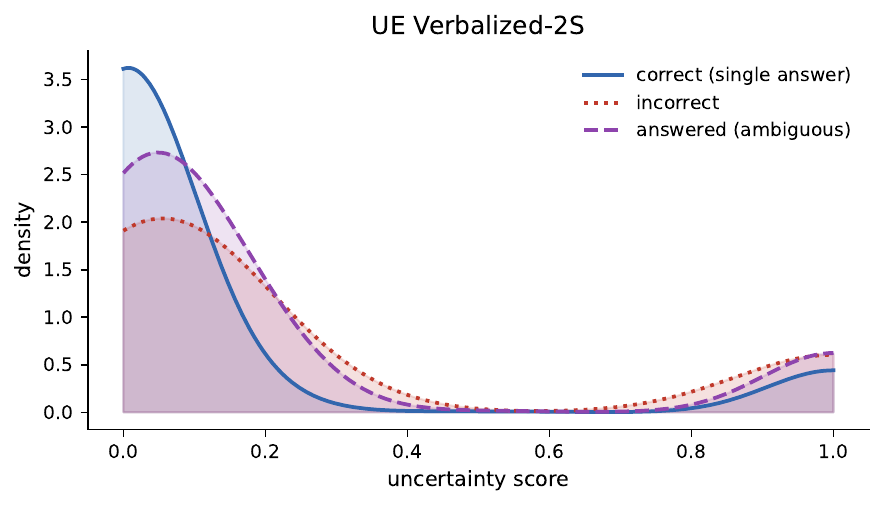}
        \caption{Verbalized-2S}
    \end{subfigure}
    \caption{Density of uncertainty scores for correct (single answer), incorrect,
    and answered (ambiguous) responses. Model - \llama.}
    \label{fig:ue_distributions_ambiguity_llama}
\end{figure}

\begin{figure}[H]
    \centering
    \begin{subfigure}[b]{0.48\columnwidth}
        \includegraphics[width=\textwidth]{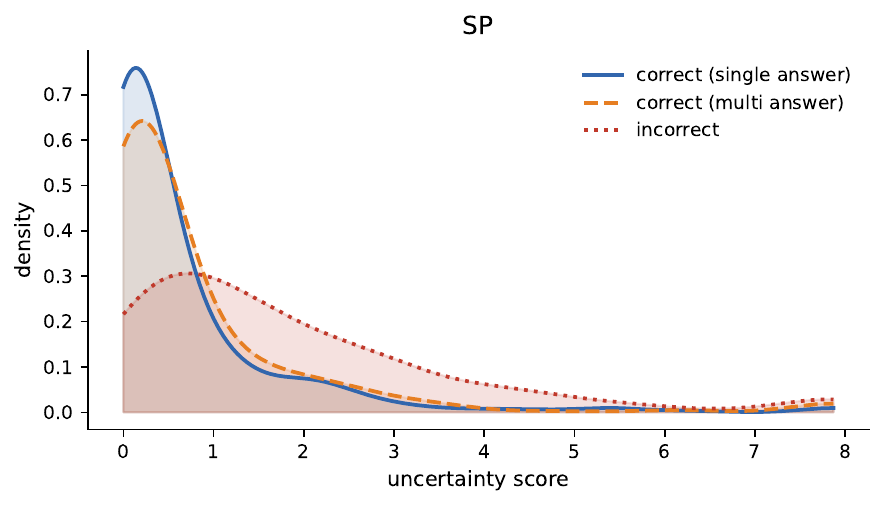}
        \caption{Sequence Probability}
    \end{subfigure}
    \hfill
    \begin{subfigure}[b]{0.48\columnwidth}
        \includegraphics[width=\textwidth]{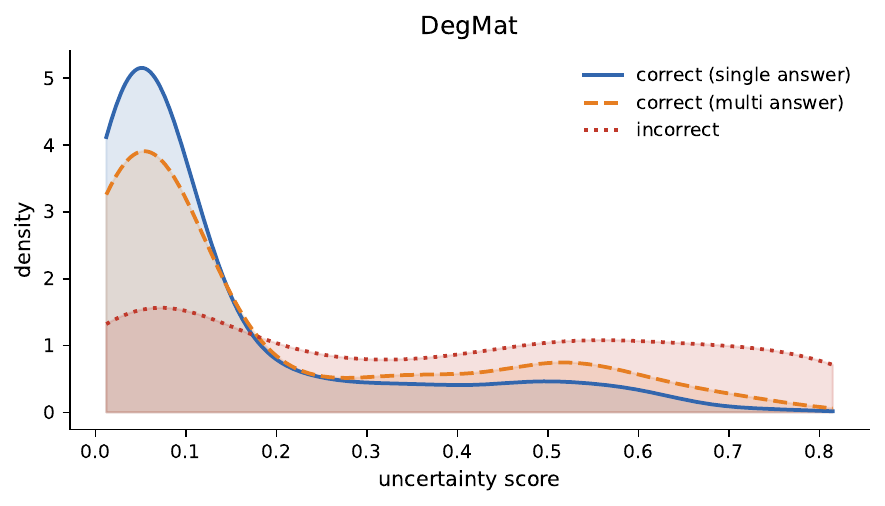}
        \caption{Degree Matrix}
    \end{subfigure}

    \vspace{0.1em}

    \begin{subfigure}[b]{0.48\columnwidth}
        \includegraphics[width=\textwidth]{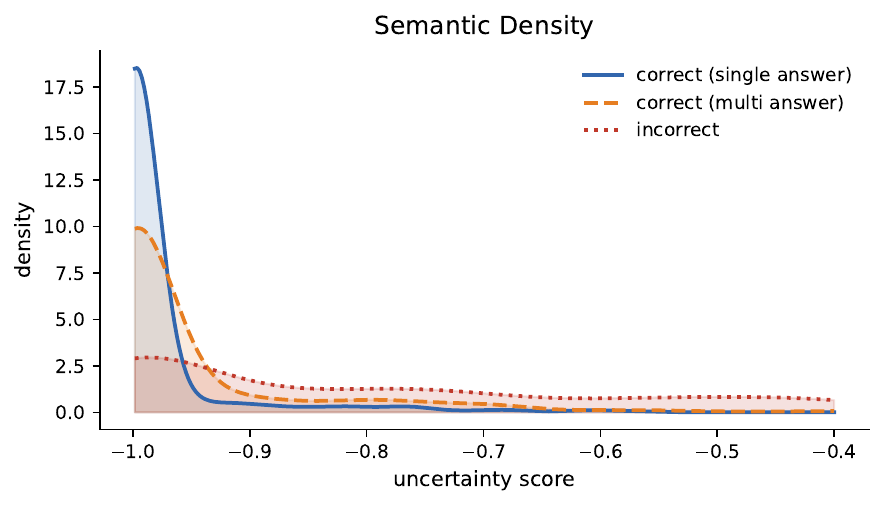}
        \caption{Semantic Density}
    \end{subfigure}
    \hfill
    \begin{subfigure}[b]{0.48\columnwidth}
        \includegraphics[width=\textwidth]{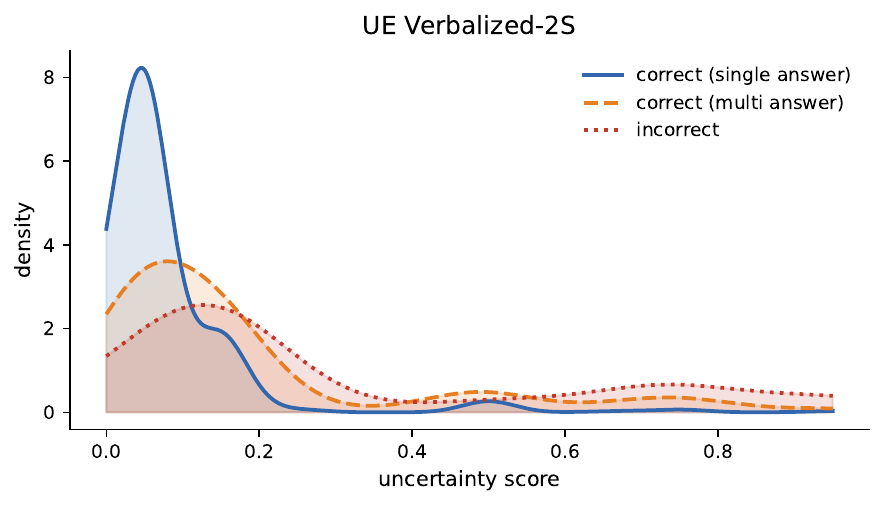}
        \caption{Verbalized-2S}
    \end{subfigure}
    \caption{Density of uncertainty scores for correct (single answer), correct (multi answer),
    and incorrect (pooled) responses. Model - \qwen.}
    \label{fig:ue_distributions_qwen}
\end{figure}

\begin{figure}[H]
    \centering
    \begin{subfigure}[b]{0.48\columnwidth}
        \includegraphics[width=\textwidth]{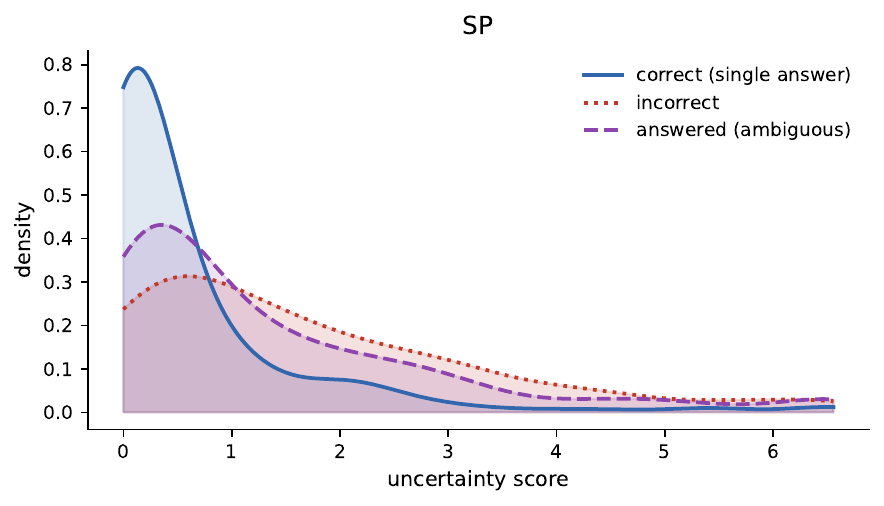}
        \caption{Sequence Probability}
    \end{subfigure}
    \hfill
    \begin{subfigure}[b]{0.48\columnwidth}
        \includegraphics[width=\textwidth]{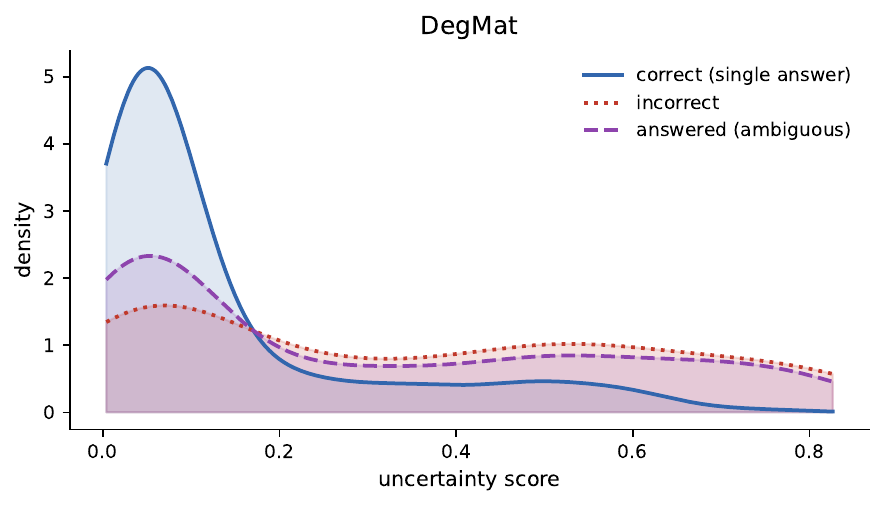}
        \caption{Degree Matrix}
    \end{subfigure}

    \vspace{0.1em}

    \begin{subfigure}[b]{0.48\columnwidth}
        \includegraphics[width=\textwidth]{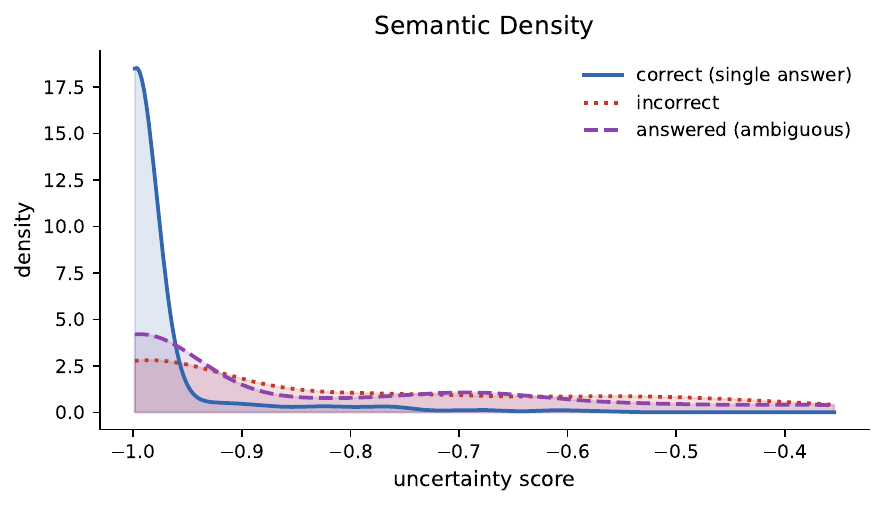}
        \caption{Semantic Density}
    \end{subfigure}
    \hfill
    \begin{subfigure}[b]{0.48\columnwidth}
        \includegraphics[width=\textwidth]{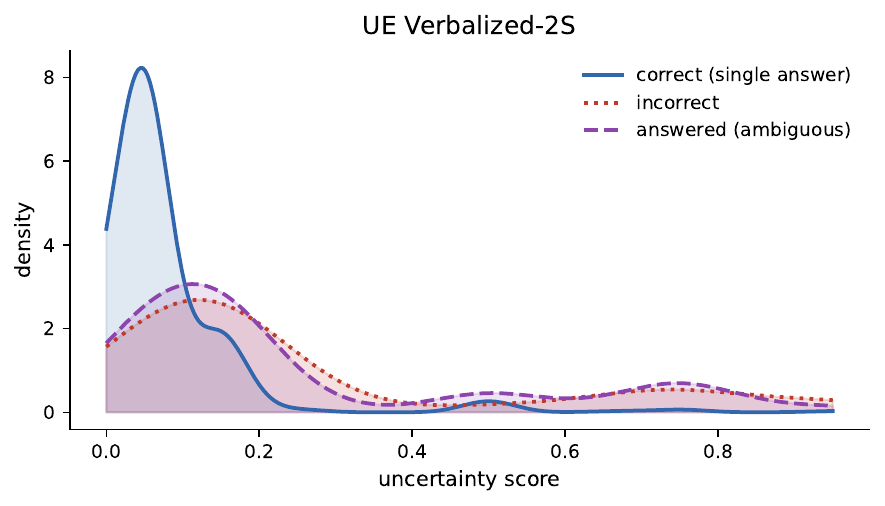}
        \caption{Verbalized-2S}
    \end{subfigure}
    \caption{Density of uncertainty scores for correct (single answer), incorrect,
    and answered (ambiguous) responses. Model - \qwen.}
    \label{fig:ue_distributions_ambiguity_qwen}
\end{figure}

\begin{figure}[H]
    \centering
    \begin{subfigure}[b]{0.48\columnwidth}
        \includegraphics[width=\textwidth]{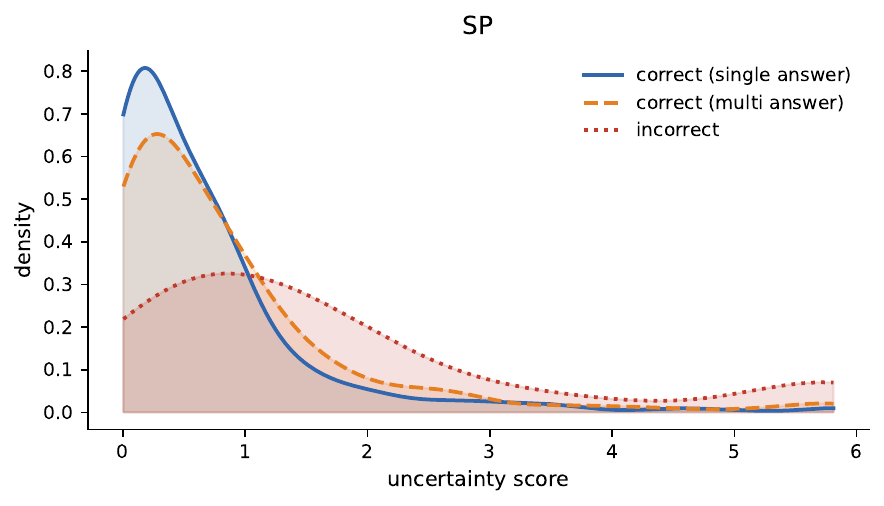}
        \caption{Sequence Probability}
    \end{subfigure}
    \hfill
    \begin{subfigure}[b]{0.48\columnwidth}
        \includegraphics[width=\textwidth]{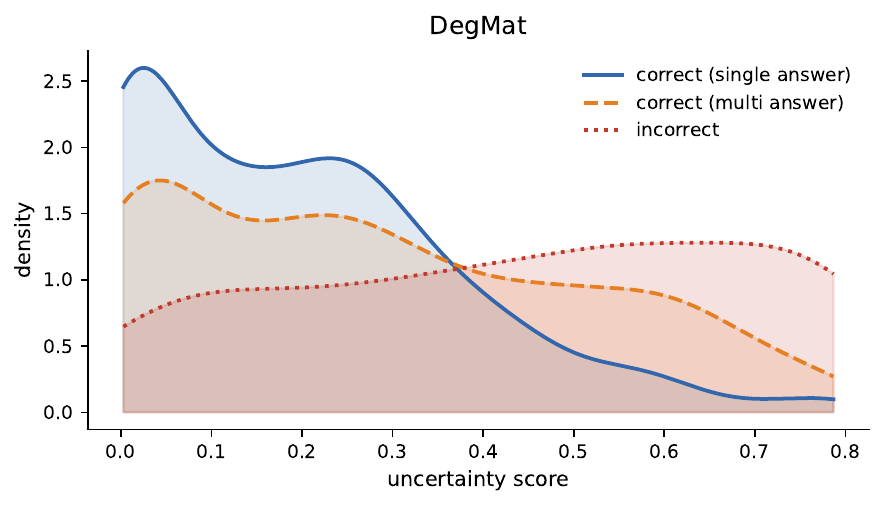}
        \caption{Degree Matrix}
    \end{subfigure}

    \vspace{0.1em}

    \begin{subfigure}[b]{0.48\columnwidth}
        \includegraphics[width=\textwidth]{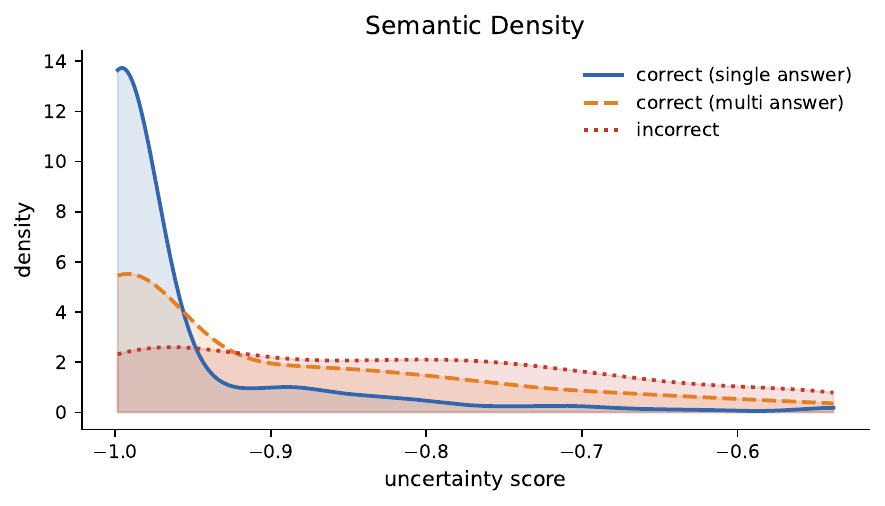}
        \caption{Semantic Density}
    \end{subfigure}
    \hfill
    \begin{subfigure}[b]{0.48\columnwidth}
        \includegraphics[width=\textwidth]{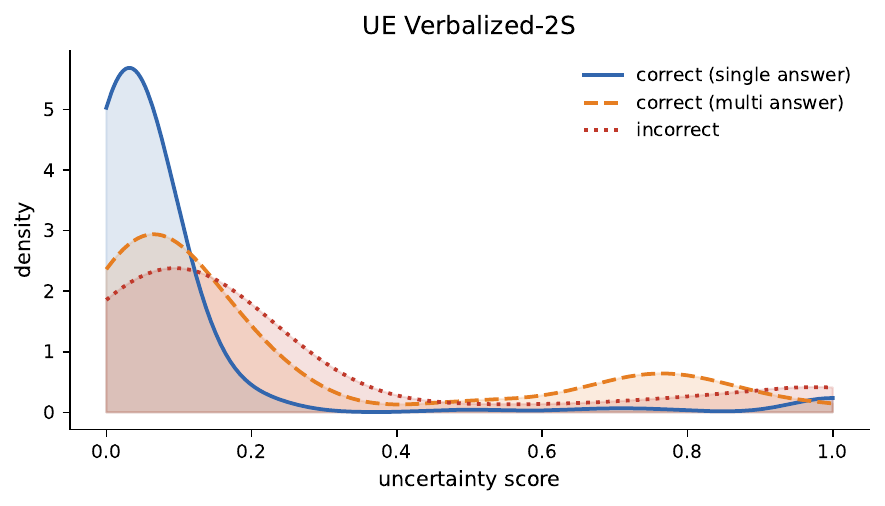}
        \caption{Verbalized-2S}
    \end{subfigure}
    \caption{Density of uncertainty scores for correct (single answer), correct (multi answer),
    and incorrect (pooled) responses. Model - \llamaseventy.}
    \label{fig:ue_distributions_llama70}
\end{figure}

\begin{figure}[H]
    \centering
    \begin{subfigure}[b]{0.48\columnwidth}
        \includegraphics[width=\textwidth]{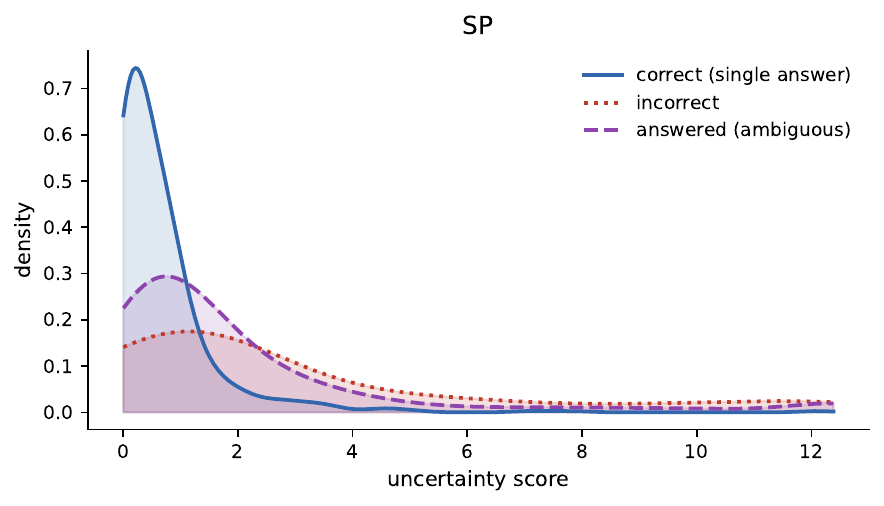}
        \caption{Sequence Probability}
    \end{subfigure}
    \hfill
    \begin{subfigure}[b]{0.48\columnwidth}
        \includegraphics[width=\textwidth]{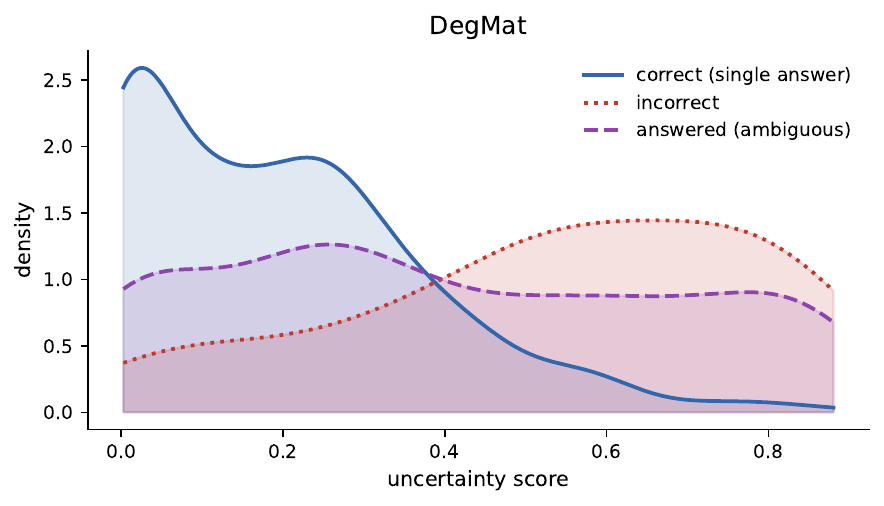}
        \caption{Degree Matrix}
    \end{subfigure}

    \vspace{0.1em}

    \begin{subfigure}[b]{0.48\columnwidth}
        \includegraphics[width=\textwidth]{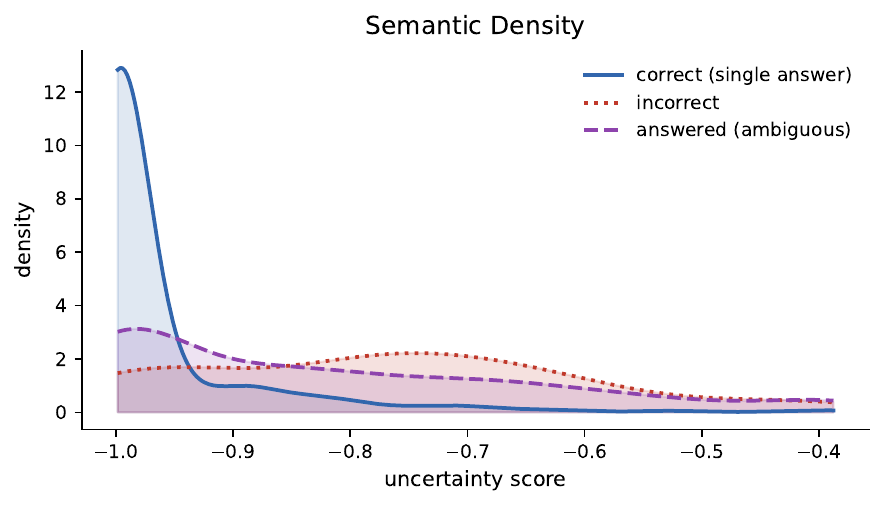}
        \caption{Semantic Density}
    \end{subfigure}
    \hfill
    \begin{subfigure}[b]{0.48\columnwidth}
        \includegraphics[width=\textwidth]{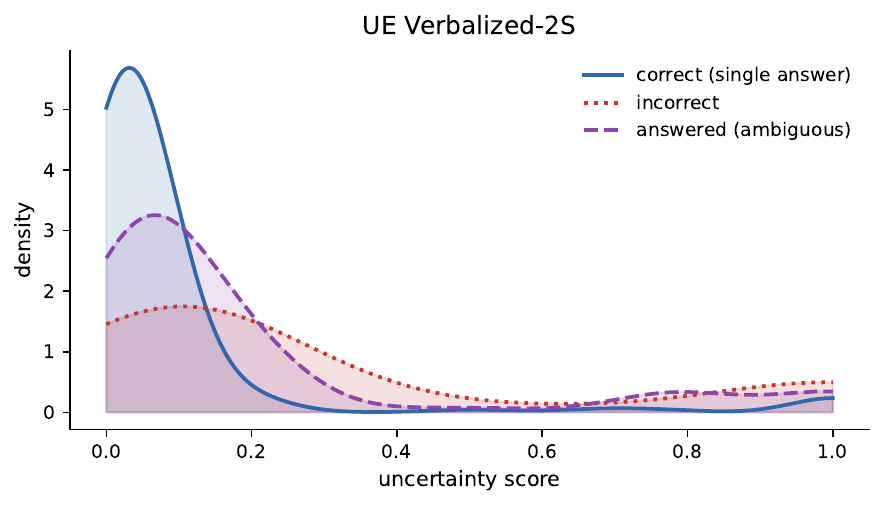}
        \caption{Verbalized-2S}
    \end{subfigure}
    \caption{Density of uncertainty scores for correct (single answer), incorrect,
    and answered (ambiguous) responses. Model - \llamaseventy.}
    \label{fig:ue_distributions_ambiguity_llama70}
\end{figure}

\begin{figure}[H]
    \centering
    \begin{subfigure}[b]{0.48\columnwidth}
        \includegraphics[width=\textwidth]{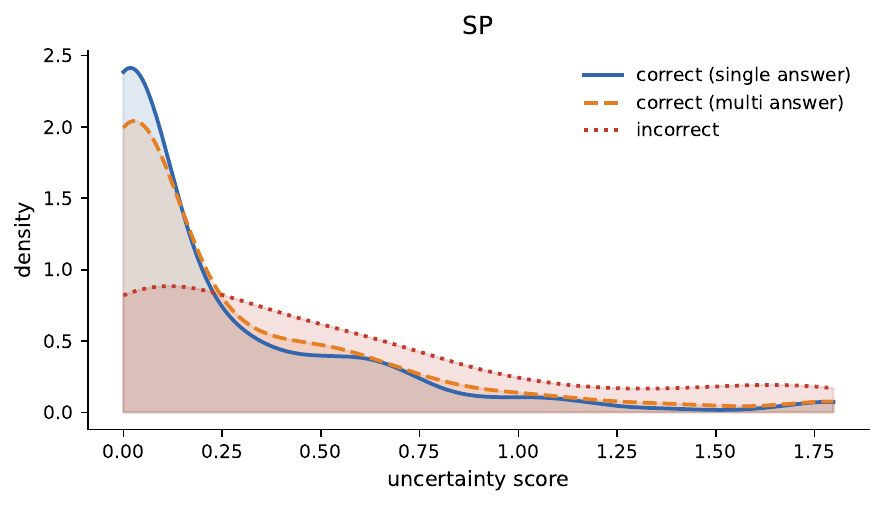}
        \caption{Sequence Probability}
    \end{subfigure}
    \hfill
    \begin{subfigure}[b]{0.48\columnwidth}
        \includegraphics[width=\textwidth]{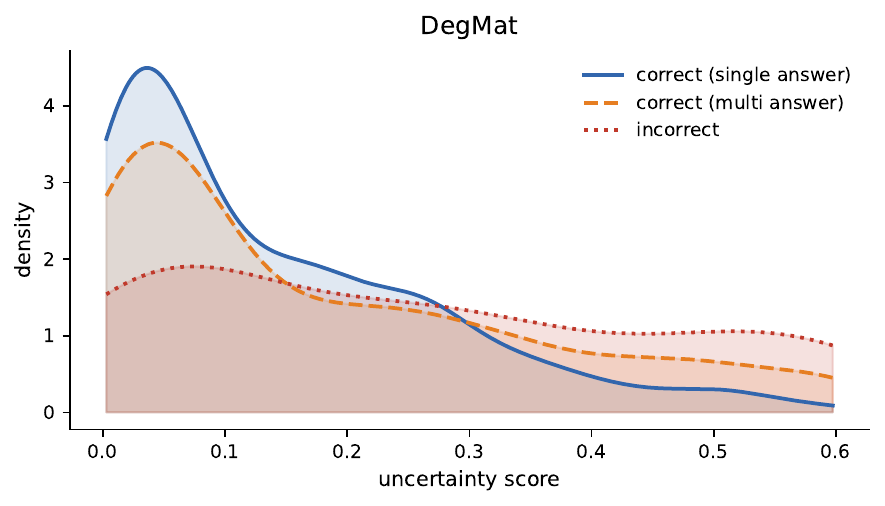}
        \caption{Degree Matrix}
    \end{subfigure}

    \vspace{0.1em}

    \begin{subfigure}[b]{0.48\columnwidth}
        \includegraphics[width=\textwidth]{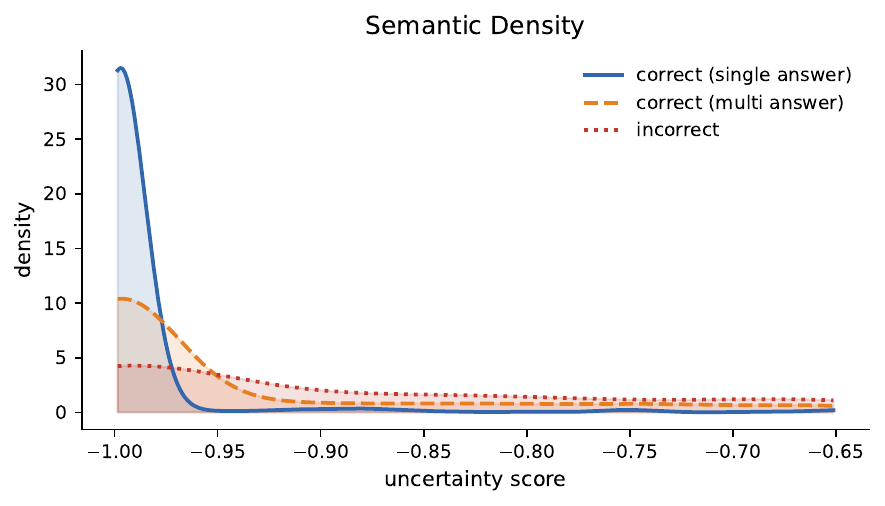}
        \caption{Semantic Density}
    \end{subfigure}
    \hfill
    \begin{subfigure}[b]{0.48\columnwidth}
        \includegraphics[width=\textwidth]{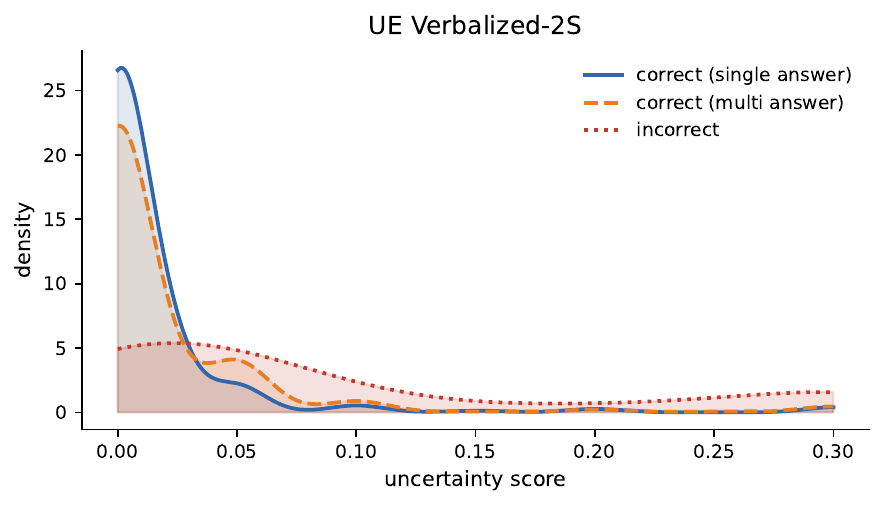}
        \caption{Verbalized-2S}
    \end{subfigure}
    \caption{Density of uncertainty scores for correct (single answer), correct (multi answer),
    and incorrect (pooled) responses. Model - \gptfour.}
    \label{fig:ue_distributions_gpt}
\end{figure}

\begin{figure}[H]
    \centering
    \begin{subfigure}[b]{0.48\columnwidth}
        \includegraphics[width=\textwidth]{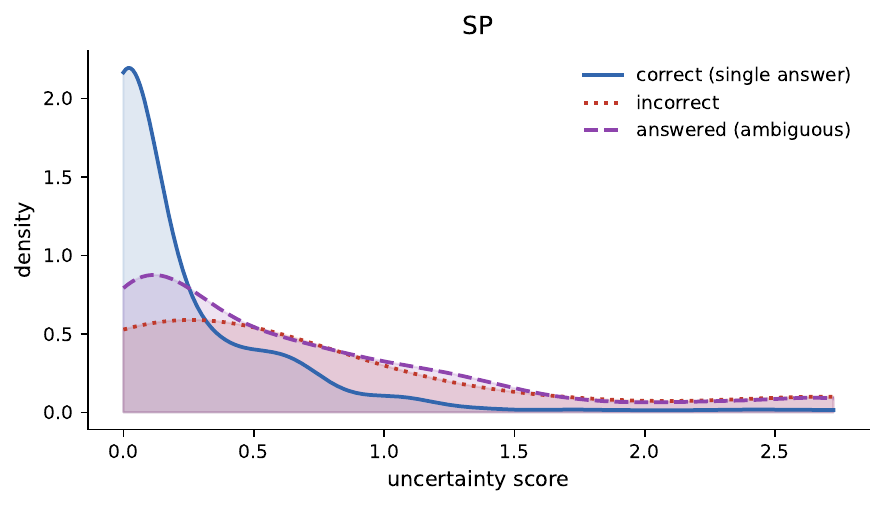}
        \caption{Sequence Probability}
    \end{subfigure}
    \hfill
    \begin{subfigure}[b]{0.48\columnwidth}
        \includegraphics[width=\textwidth]{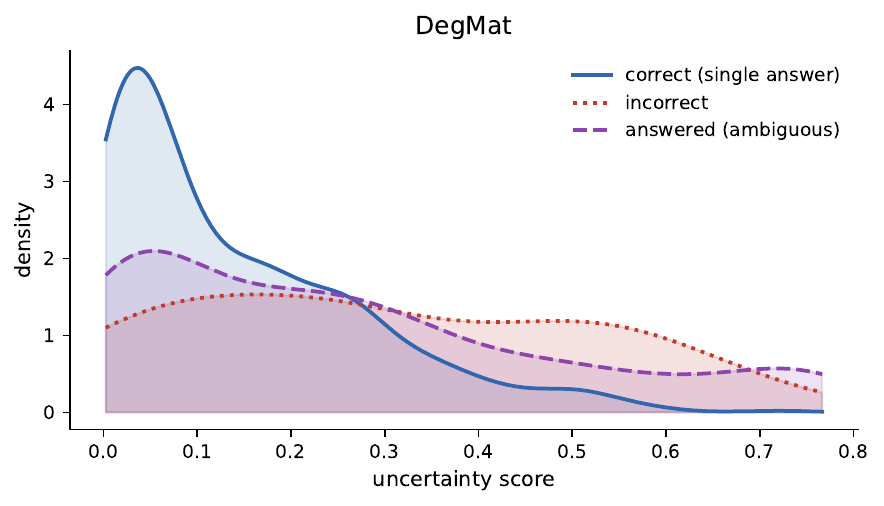}
        \caption{Degree Matrix}
    \end{subfigure}

    \vspace{0.1em}

    \begin{subfigure}[b]{0.48\columnwidth}
        \includegraphics[width=\textwidth]{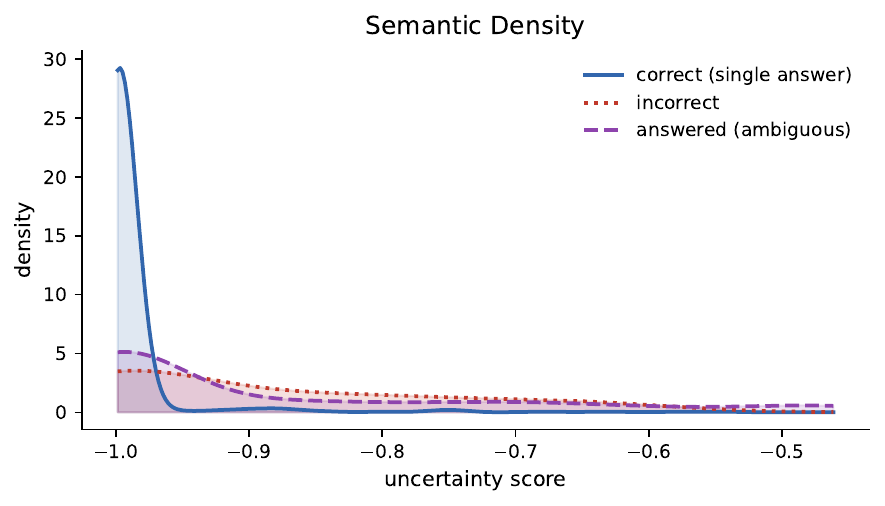}
        \caption{Semantic Density}
    \end{subfigure}
    \hfill
    \begin{subfigure}[b]{0.48\columnwidth}
        \includegraphics[width=\textwidth]{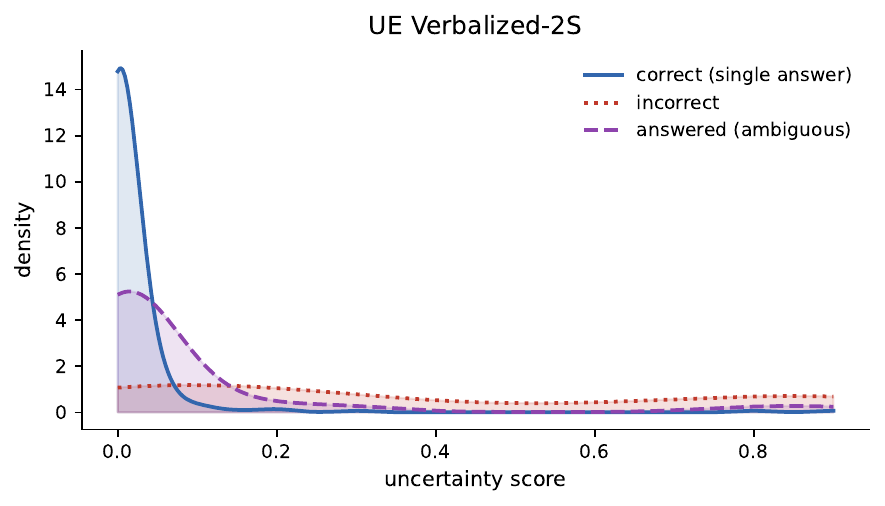}
        \caption{Verbalized-2S}
    \end{subfigure}
    \caption{Density of uncertainty scores for correct (single answer), incorrect,
    and answered (ambiguous) responses.  Model - \gptfour.}
    \label{fig:ue_distributions_ambiguity_gpt}
\end{figure}

This appendix provides density plots of uncertainty scores for each evaluated model, using one representative method per family: Sequence Probability, Degree Matrix, Semantic Density, and Verbalized-2S. 
Figures~\ref{fig:ue_distributions_llama}--\ref{fig:ue_distributions_ambiguity_gpt} show these plots, with two figures per model and one panel per method. The first figure for each model (Figures~\ref{fig:ue_distributions_llama}, \ref{fig:ue_distributions_qwen}, \ref{fig:ue_distributions_llama70} and \ref{fig:ue_distributions_gpt}) compares correct (single answer), correct (multi-answer) and incorrect responses; the second (Figures~\ref{fig:ue_distributions_ambiguity_llama}, \ref{fig:ue_distributions_ambiguity_qwen}, \ref{fig:ue_distributions_ambiguity_llama70} and \ref{fig:ue_distributions_ambiguity_gpt}) compares correct (single answer), incorrect and answered (ambiguous) responses.

%% file: sections/appendix/app_large_models.tex
\section{Experiments with Reasoning Models}
\label{app:reasoning_models}
To examine whether our findings extend to more recent reasoning models, we additionally evaluate GPT-5.4 and GPT-5.6. These models do not expose token-level probabilities or other internal signals, so we restrict this analysis to the uncertainty estimation methods that operate in a fully black-box setting, namely NumSemSets, EigValLaplacian, DegMat, Eccentricity, and Verbalized2s. Methods requiring white-box access are therefore excluded. Tables~\ref{tab:prr_gpt_5.4} and~\ref{tab:prr_gpt_5.6} report PRR scores for each setting.  The overall pattern observed in the main experiments also holds for these newer models: uncertainty quantification methods perform best when model knowledge limitation is the only source of uncertainty, and their performance degrades consistently once additional sources of uncertainty are introduced.


\begin{table}[H]
\centering
\scalebox{0.65}{
  \begin{tabular}{lcccc}
  \toprule
  \textbf{Metric} & \textbf{MK} & \textbf{MK+OV} & \textbf{MK+IA} & \textbf{MK+OV+IA} \\
  \midrule
  NumSemSets & 0.441 & 0.182 {\small {\color{red}$\downarrow$}0.259} & 0.355 {\small {\color{red}$\downarrow$}0.086} & 0.280 {\small {\color{red}$\downarrow$}0.161} \\
  EigValLaplacian & \underline{0.729} & \underline{0.405 {\small {\color{red}$\downarrow$}0.324}} & \underline{0.445 {\small {\color{red}$\downarrow$}0.284}} & 0.369 {\small {\color{red}$\downarrow$}0.360} \\
  DegMat & 0.690 & 0.269 {\small {\color{red}$\downarrow$}0.421} & 0.397 {\small {\color{red}$\downarrow$}0.293} & 0.317 {\small {\color{red}$\downarrow$}0.373} \\
  Eccentricity & \textbf{0.745} & 0.392 {\small {\color{red}$\downarrow$}0.353} & 0.435 {\small {\color{red}$\downarrow$}0.310} & \underline{0.385 {\small {\color{red}$\downarrow$}0.360}} \\
  Verbalized2S & 0.625 & \textbf{0.475 {\small {\color{red}$\downarrow$}0.150}} & \textbf{0.577 {\small {\color{red}$\downarrow$}0.048}} & \textbf{0.542 {\small {\color{red}$\downarrow$}0.083}} \\
  \bottomrule
  \end{tabular}
}
\caption{PRR scores for GPT-5.4. Inline arrows show drop from MK.}
\label{tab:prr_gpt_5.4}
\end{table}

\begin{table}[H]
\centering
\scalebox{0.65}{
  \begin{tabular}{lcccc}
  \toprule
  \textbf{Metric} & \textbf{MK} & \textbf{MK+OV} & \textbf{MK+IA} & \textbf{MK+OV+IA} \\
  \midrule
  NumSemSets & 0.340 & -0.029 {\small {\color{red}$\downarrow$}0.369} & 0.265 {\small {\color{red}$\downarrow$}0.075} & 0.216 {\small {\color{red}$\downarrow$}0.124} \\
  EigValLaplacian & 0.506 & \textbf{0.314 {\small {\color{red}$\downarrow$}0.192}} & \underline{0.304 {\small {\color{red}$\downarrow$}0.202}} & \underline{0.277 {\small {\color{red}$\downarrow$}0.229}} \\
  DegMat & \textbf{0.623} & 0.180 {\small {\color{red}$\downarrow$}0.443} & 0.202 {\small {\color{red}$\downarrow$}0.421} & 0.182 {\small {\color{red}$\downarrow$}0.441} \\
  Eccentricity & \underline{0.583} & \underline{0.278 {\small {\color{red}$\downarrow$}0.305}} & 0.250 {\small {\color{red}$\downarrow$}0.333} & 0.243 {\small {\color{red}$\downarrow$}0.340} \\
  Verbalized2S & 0.478 & 0.209 {\small {\color{red}$\downarrow$}0.269} & \textbf{0.387 {\small {\color{red}$\downarrow$}0.091}} & \textbf{0.422 {\small {\color{red}$\downarrow$}0.056}} \\
  \bottomrule
  \end{tabular}
}
\caption{PRR scores for GPT-5.6. Inline arrows show drop from MK.}
\label{tab:prr_gpt_5.6}
\end{table}


%% file: custom.bib
@misc{Baan2023UncertaintyIN,
  title         = {Uncertainty in Natural Language Generation: From Theory to Applications},
  author        = {Baan, Joris and Daheim, Nico and Ilia, Evgenia and Ulmer, Dennis and Li, Haau-Sing and Fern{\'a}ndez, Raquel and Plank, Barbara and Sennrich, Rico and Zerva, Chrysoula and Aziz, Wilker},
  year          = {2023},
  eprint        = {2307.15703},
  archivePrefix = {arXiv},
  primaryClass  = {cs.CL},
  url           = {https://arxiv.org/abs/2307.15703},
}

@inproceedings{duan-etal-2024-shifting,
  title     = {Shifting Attention to Relevance: Towards the Predictive Uncertainty Quantification of Free-Form Large Language Models},
  author    = {Duan, Jinhao and Cheng, Hao and Wang, Shiqi and Zavalny, Alex and Wang, Chenan and Xu, Renjing and Kailkhura, Bhavya and Xu, Kaidi},
  booktitle = {Proceedings of the 62nd Annual Meeting of the Association for Computational Linguistics (Volume 1: Long Papers)},
  pages     = {5050--5063},
  year      = {2024},
  month     = aug,
  address   = {Bangkok, Thailand},
  publisher = {Association for Computational Linguistics},
  doi       = {10.18653/v1/2024.acl-long.276},
  url       = {https://aclanthology.org/2024.acl-long.276/},
}

@inproceedings{fadeeva-etal-2023-lm,
  title     = {{LM-Polygraph}: Uncertainty Estimation for Language Models},
  author    = {Fadeeva, Ekaterina and Vashurin, Roman and Tsvigun, Akim and Vazhentsev, Artem and Petrakov, Sergey and Fedyanin, Kirill and Vasilev, Daniil and Goncharova, Elizaveta and Panchenko, Alexander and Panov, Maxim and Baldwin, Timothy and Shelmanov, Artem},
  booktitle = {Proceedings of the 2023 Conference on Empirical Methods in Natural Language Processing: System Demonstrations},
  pages     = {446--461},
  year      = {2023},
  month     = dec,
  address   = {Singapore},
  publisher = {Association for Computational Linguistics},
  doi       = {10.18653/v1/2023.emnlp-demo.41},
  url       = {https://aclanthology.org/2023.emnlp-demo.41/},
}

@inproceedings{fadeeva2024factchecking,
  title     = {Fact-Checking the Output of Large Language Models via Token-Level Uncertainty Quantification},
  author    = {Fadeeva, Ekaterina and Rubashevskii, Aleksandr and Shelmanov, Artem and Petrakov, Sergey and Li, Haonan and Mubarak, Hamdy and Tsymbalov, Evgenii and Kuzmin, Gleb and Panchenko, Alexander and Baldwin, Timothy and Nakov, Preslav and Panov, Maxim},
  booktitle = {Findings of the Association for Computational Linguistics: ACL 2024},
  pages     = {9367--9385},
  year      = {2024},
  month     = aug,
  address   = {Bangkok, Thailand},
  publisher = {Association for Computational Linguistics},
  doi       = {10.18653/v1/2024.findings-acl.558},
  url       = {https://aclanthology.org/2024.findings-acl.558/},
}

@article{fomicheva-etal-2020-unsupervised,
  title     = {Unsupervised Quality Estimation for Neural Machine Translation},
  author    = {Fomicheva, Marina and Sun, Shuo and Yankovskaya, Lisa and Blain, Fr{\'e}d{\'e}ric and Guzm{\'a}n, Francisco and Fishel, Mark and Aletras, Nikolaos and Chaudhary, Vishrav and Specia, Lucia},
  journal   = {Transactions of the Association for Computational Linguistics},
  volume    = {8},
  pages     = {539--555},
  year      = {2020},
  publisher = {MIT Press},
  doi       = {10.1162/tacl_a_00330},
  url       = {https://aclanthology.org/2020.tacl-1.35/},
}

@misc{gemma,
  title         = {{Gemma} 3 Technical Report},
  author        = {{Gemma Team}},
  year          = {2025},
  eprint        = {2503.19786},
  archivePrefix = {arXiv},
  primaryClass  = {cs.CL},
  url           = {https://arxiv.org/abs/2503.19786},
}

@misc{gpt41,
  title        = {Introducing {GPT-4.1} in the {API}},
  author       = {{OpenAI}},
  year         = {2025},
  month        = apr,
  howpublished = {\url{https://openai.com/index/gpt-4-1/}},
  url          = {https://openai.com/index/gpt-4-1/},
}

@misc{hu2023uncertaintynaturallanguageprocessing,
  title         = {Uncertainty in Natural Language Processing: Sources, Quantification, and Applications},
  author        = {Hu, Mengting and Zhang, Zhen and Zhao, Shiwan and Huang, Minlie and Wu, Bingzhe},
  year          = {2023},
  eprint        = {2306.04459},
  archivePrefix = {arXiv},
  primaryClass  = {cs.CL},
  url           = {https://arxiv.org/abs/2306.04459},
}

@article{Hllermeier2021,
  title     = {Aleatoric and Epistemic Uncertainty in Machine Learning: An Introduction to Concepts and Methods},
  author    = {H{\"u}llermeier, Eyke and Waegeman, Willem},
  journal   = {Machine Learning},
  volume    = {110},
  number    = {3},
  pages     = {457--506},
  year      = {2021},
  publisher = {Springer},
  doi       = {10.1007/s10994-021-05946-3},
}

@misc{kadavath2022languagemodelsmostlyknow,
  title         = {Language Models (Mostly) Know What They Know},
  author        = {Kadavath, Saurav and Conerly, Tom and Askell, Amanda and Henighan, Tom and Drain, Dawn and Perez, Ethan and Schiefer, Nicholas and Hatfield-Dodds, Zac and DasSarma, Nova and Tran-Johnson, Eli and Johnston, Scott and El-Showk, Sheer and Jones, Andy and Elhage, Nelson and Hume, Tristan and Chen, Anna and Bai, Yuntao and Bowman, Sam and Fort, Stanislav and Ganguli, Deep and Hernandez, Danny and Jacobson, Josh and Kernion, Jackson and Kravec, Shauna and Lovitt, Liane and Ndousse, Kamal and Olsson, Catherine and Ringer, Sam and Amodei, Dario and Brown, Tom and Clark, Jack and Joseph, Nicholas and Mann, Ben and McCandlish, Sam and Olah, Chris and Kaplan, Jared},
  year          = {2022},
  eprint        = {2207.05221},
  archivePrefix = {arXiv},
  primaryClass  = {cs.CL},
  url           = {https://arxiv.org/abs/2207.05221},
}

@inproceedings{kirchhof2025position,
  title     = {Position: Uncertainty Quantification Needs Reassessment for Large Language Model Agents},
  author    = {Kirchhof, Michael and Kasneci, Gjergji and Kasneci, Enkelejda},
  booktitle = {Proceedings of the 42nd International Conference on Machine Learning},
  series    = {Proceedings of Machine Learning Research},
  volume    = {267},
  pages     = {81665--81677},
  year      = {2025},
  month     = jul,
  address   = {Vancouver, Canada},
  publisher = {PMLR},
  url       = {https://proceedings.mlr.press/v267/kirchhof25b.html},
}

@inproceedings{kuhn2023semantic,
  title     = {Semantic Uncertainty: Linguistic Invariances for Uncertainty Estimation in Natural Language Generation},
  author    = {Kuhn, Lorenz and Gal, Yarin and Farquhar, Sebastian},
  booktitle = {Proceedings of the Eleventh International Conference on Learning Representations},
  year      = {2023},
  month     = may,
  address   = {Kigali, Rwanda},
  publisher = {OpenReview.net},
  url       = {https://openreview.net/forum?id=VD-AYtP0dve},
}

@article{lin2024generating,
  title   = {Generating with Confidence: Uncertainty Quantification for Black-box Large Language Models},
  author  = {Lin, Zhen and Trivedi, Shubhendu and Sun, Jimeng},
  journal = {Transactions on Machine Learning Research},
  year    = {2024},
  url     = {https://openreview.net/forum?id=DWkJCSxKU5},
}

@inproceedings{10.1145/3711896.3736569,
  title     = {Uncertainty Quantification and Confidence Calibration in Large Language Models: A Survey},
  author    = {Liu, Xiaoou and Chen, Tiejin and Da, Longchao and Chen, Chacha and Lin, Zhen and Wei, Hua},
  booktitle = {Proceedings of the 31st ACM SIGKDD Conference on Knowledge Discovery and Data Mining V.2},
  pages     = {6107--6117},
  year      = {2025},
  month     = aug,
  address   = {Toronto, Canada},
  publisher = {Association for Computing Machinery},
  doi       = {10.1145/3711896.3736569},
}

@misc{llama,
  title         = {The {Llama} 3 Herd of Models},
  author        = {{Llama Team}},
  year          = {2024},
  eprint        = {2407.21783},
  archivePrefix = {arXiv},
  primaryClass  = {cs.AI},
  url           = {https://arxiv.org/abs/2407.21783},
}

@inproceedings{malinin2021uncertainty,
  title     = {Uncertainty Estimation in Autoregressive Structured Prediction},
  author    = {Malinin, Andrey and Gales, Mark J. F.},
  booktitle = {Proceedings of the Ninth International Conference on Learning Representations},
  year      = {2021},
  month     = may,
  address   = {Online},
  publisher = {OpenReview.net},
  url       = {https://openreview.net/forum?id=jN5y-zb5Q7m},
}

@inproceedings{min-etal-2020-ambigqa,
  title     = {{AmbigQA}: Answering Ambiguous Open-domain Questions},
  author    = {Min, Sewon and Michael, Julian and Hajishirzi, Hannaneh and Zettlemoyer, Luke},
  booktitle = {Proceedings of the 2020 Conference on Empirical Methods in Natural Language Processing (EMNLP)},
  pages     = {5783--5797},
  year      = {2020},
  month     = nov,
  address   = {Online},
  publisher = {Association for Computational Linguistics},
  doi       = {10.18653/v1/2020.emnlp-main.466},
  url       = {https://aclanthology.org/2020.emnlp-main.466/},
}

@inproceedings{pletenev-etal-2025-will,
    title = "Will It Still Be True Tomorrow? Multilingual Evergreen Question Classification to Improve Trustworthy {QA}",
    author = "Pletenev, Sergey  and
      Marina, Maria  and
      Ivanov, Nikolay  and
      Galimzianova, Daria  and
      Krayko, Nikita  and
      Salnikov, Mikhail  and
      Konovalov, Vasily  and
      Panchenko, Alexander  and
      Moskvoretskii, Viktor",
    editor = "Christodoulopoulos, Christos  and
      Chakraborty, Tanmoy  and
      Rose, Carolyn  and
      Peng, Violet",
    booktitle = "Proceedings of the 2025 Conference on Empirical Methods in Natural Language Processing",
    month = nov,
    year = "2025",
    address = "Suzhou, China",
    publisher = "Association for Computational Linguistics",
    url = "https://aclanthology.org/2025.emnlp-main.434/",
    doi = "10.18653/v1/2025.emnlp-main.434",
    pages = "8603--8620",
    ISBN = "979-8-89176-332-6"
}

@misc{Prabhudesai2024-tw,
  title         = {A Taxonomy for Understanding and Identifying Uncertainty in {AI}-Generated Responses},
  author        = {Prabhudesai, Snehal and Goldstein, Daniel G. and Hofman, Jake M. and Rothschild, David M.},
  year          = {2024},
  month         = may,
  eprint        = {4836380},
  archivePrefix = {SSRN},
  doi           = {10.2139/ssrn.4836380},
  url           = {https://papers.ssrn.com/sol3/papers.cfm?abstract_id=4836380},
}

@inproceedings{qiu2024semantic,
  title     = {Semantic Density: Uncertainty Quantification for Large Language Models through Confidence Measurement in Semantic Space},
  author    = {Qiu, Xin and Miikkulainen, Risto},
  booktitle = {Advances in Neural Information Processing Systems},
  volume    = {37},
  pages     = {134507--134533},
  year      = {2024},
  month     = dec,
  address   = {Vancouver, Canada},
  publisher = {Curran Associates, Inc.},
  doi       = {10.52202/079017-4274},
  url       = {https://openreview.net/forum?id=LOH6qzI7T6},
}

@misc{qwen,
  title        = {{Qwen2.5}: A Party of Foundation Models!},
  author       = {{Qwen Team}},
  year         = {2024},
  month        = sep,
  howpublished = {\url{https://qwenlm.github.io/blog/qwen2.5/}},
  url          = {https://qwenlm.github.io/blog/qwen2.5/},
}

@inproceedings{rajpurkar-etal-2018-know,
  title     = {Know What You Don{'}t Know: Unanswerable Questions for {SQuAD}},
  author    = {Rajpurkar, Pranav and Jia, Robin and Liang, Percy},
  booktitle = {Proceedings of the 56th Annual Meeting of the Association for Computational Linguistics (Volume 2: Short Papers)},
  pages     = {784--789},
  year      = {2018},
  month     = jul,
  address   = {Melbourne, Australia},
  publisher = {Association for Computational Linguistics},
  doi       = {10.18653/v1/P18-2124},
  url       = {https://aclanthology.org/P18-2124/},
}

@inproceedings{tao2025revisiting,
  title     = {Revisiting Uncertainty Estimation and Calibration of Large Language Models},
  author    = {Tao, Linwei and Yeh, Yi-Fan and Dong, Minjing and Huang, Tao and Yu, Jialin and Torr, Philip and Xu, Chang},
  booktitle = {NeurIPS 2025 Workshop on Scaling Environments for Agents (SEA)},
  year      = {2025},
  month     = dec,
  address   = {San Diego, California, USA},
  url       = {https://openreview.net/forum?id=Q9CreVjHH7},
}

@inproceedings{tian-etal-2023-just,
  title     = {Just Ask for Calibration: Strategies for Eliciting Calibrated Confidence Scores from Language Models Fine-Tuned with Human Feedback},
  author    = {Tian, Katherine and Mitchell, Eric and Zhou, Allan and Sharma, Archit and Rafailov, Rafael and Yao, Huaxiu and Finn, Chelsea and Manning, Christopher},
  booktitle = {Proceedings of the 2023 Conference on Empirical Methods in Natural Language Processing},
  pages     = {5433--5442},
  year      = {2023},
  month     = dec,
  address   = {Singapore},
  publisher = {Association for Computational Linguistics},
  doi       = {10.18653/v1/2023.emnlp-main.330},
  url       = {https://aclanthology.org/2023.emnlp-main.330/},
}

@misc{tomov2026illusioncertaintyuncertaintyquantification,
      title={The Illusion of Certainty: Uncertainty Quantification for LLMs Fails under Ambiguity}, 
      author={Tim Tomov and Dominik Fuchsgruber and Tom Wollschläger and Stephan Günnemann},
      year={2026},
      eprint={2511.04418},
      archivePrefix={arXiv},
      primaryClass={cs.LG},
      url={https://arxiv.org/abs/2511.04418}, 
}

@article{vashurin-etal-2025-benchmarking,
  title     = {Benchmarking Uncertainty Quantification Methods for Large Language Models with {LM-Polygraph}},
  author    = {Vashurin, Roman and Fadeeva, Ekaterina and Vazhentsev, Artem and Rvanova, Lyudmila and Vasilev, Daniil and Tsvigun, Akim and Petrakov, Sergey and Xing, Rui and Sadallah, Abdelrahman and Grishchenkov, Kirill and Panchenko, Alexander and Baldwin, Timothy and Nakov, Preslav and Panov, Maxim and Shelmanov, Artem},
  journal   = {Transactions of the Association for Computational Linguistics},
  volume    = {13},
  pages     = {220--248},
  year      = {2025},
  publisher = {MIT Press},
  doi       = {10.1162/tacl_a_00737},
  url       = {https://aclanthology.org/2025.tacl-1.11/},
}

@inproceedings{vashurin2025cocoa,
  title     = {{CoCoA}: A Minimum {Bayes} Risk Framework Bridging Confidence and Consistency for Uncertainty Quantification in {LLMs}},
  author    = {Vashurin, Roman and Goloburda, Maiya and Ilina, Albina and Rubashevskii, Aleksandr and Nakov, Preslav and Shelmanov, Artem and Panov, Maxim},
  booktitle = {Advances in Neural Information Processing Systems},
  volume    = {38},
  pages     = {117655--117700},
  year      = {2025},
  month     = dec,
  address   = {San Diego, California, USA},
  publisher = {Curran Associates, Inc.},
  doi       = {10.52202/085713-3546},
  url       = {https://openreview.net/forum?id=H1NGlLNaVC},
}

@inproceedings{xia-etal-2025-survey,
  title     = {A Survey of Uncertainty Estimation Methods on Large Language Models},
  author    = {Xia, Zhiqiu and Xu, Jinxuan and Zhang, Yuqian and Liu, Hang},
  booktitle = {Findings of the Association for Computational Linguistics: ACL 2025},
  pages     = {21381--21396},
  year      = {2025},
  month     = jul,
  address   = {Vienna, Austria},
  publisher = {Association for Computational Linguistics},
  doi       = {10.18653/v1/2025.findings-acl.1101},
  url       = {https://aclanthology.org/2025.findings-acl.1101/},
}

@inproceedings{yang-etal-2025-maqa,
    title = "{MAQA}: Evaluating Uncertainty Quantification in {LLM}s Regarding Data Uncertainty",
    author = "Yang, Yongjin  and
      Yoo, Haneul  and
      Lee, Hwaran",
    editor = "Chiruzzo, Luis  and
      Ritter, Alan  and
      Wang, Lu",
    booktitle = "Findings of the {A}ssociation for {C}omputational {L}inguistics: {NAACL} 2025",
    month = apr,
    year = "2025",
    address = "Albuquerque, New Mexico",
    publisher = "Association for Computational Linguistics",
    url = "https://aclanthology.org/2025.findings-naacl.325/",
    doi = "10.18653/v1/2025.findings-naacl.325",
    pages = "5861--5878",
    ISBN = "979-8-89176-195-7"
}

@inproceedings{zhang-etal-2024-clamber,
    title = "{CLAMBER}: A Benchmark of Identifying and Clarifying Ambiguous Information Needs in Large Language Models",
    author = "Zhang, Tong  and
      Qin, Peixin  and
      Deng, Yang  and
      Huang, Chen  and
      Lei, Wenqiang  and
      Liu, Junhong  and
      Jin, Dingnan  and
      Liang, Hongru  and
      Chua, Tat-Seng",
    editor = "Ku, Lun-Wei  and
      Martins, Andre  and
      Srikumar, Vivek",
    booktitle = "Proceedings of the 62nd Annual Meeting of the {A}ssociation for {C}omputational {L}inguistics (Volume 1: Long Papers)",
    month = aug,
    year = "2024",
    address = "Bangkok, Thailand",
    publisher = "Association for Computational Linguistics",
    url = "https://aclanthology.org/2024.acl-long.578/",
    doi = "10.18653/v1/2024.acl-long.578",
    pages = "10746--10766"
}

@inproceedings{kirichenko2026abstentionbench,
  title     = {AbstentionBench: Reasoning {LLM}s Fail on Unanswerable Questions},
author={Polina Kirichenko and Mark Ibrahim and Kamalika Chaudhuri and Samuel Bell},
booktitle={The Thirty-ninth Annual Conference on Neural Information Processing Systems Datasets and Benchmarks Track},
url={https://openreview.net/forum?id=OkHC30LLpO},
  volume    = {38},
  year      = {2025},
  month     = dec,
  address   = {San Diego, California, USA},
  publisher = {Curran Associates, Inc.},
}

@inproceedings{manakul-etal-2023-selfcheckgpt,
    title = "{S}elf{C}heck{GPT}: Zero-Resource Black-Box Hallucination Detection for Generative Large Language Models",
    author = "Manakul, Potsawee  and
      Liusie, Adian  and
      Gales, Mark",
    editor = "Bouamor, Houda  and
      Pino, Juan  and
      Bali, Kalika",
    booktitle = "Proceedings of the 2023 Conference on Empirical Methods in Natural Language Processing",
    month = dec,
    year = "2023",
    address = "Singapore",
    publisher = "Association for Computational Linguistics",
    url = "https://aclanthology.org/2023.emnlp-main.557/",
    doi = "10.18653/v1/2023.emnlp-main.557",
    pages = "9004--9017"
}

@article{farquhar2024detecting,
  title   = {Detecting Hallucinations in Large Language Models Using Semantic Entropy},
  author  = {Farquhar, Sebastian and Kossen, Jannik and Kuhn, Lorenz and Gal, Yarin},
  journal = {Nature},
  volume  = {630},
  number  = {8017},
  pages   = {625--630},
  year    = {2024},
  doi     = {10.1038/s41586-024-07421-0},
}

@inproceedings{stelmakh-etal-2022-asqa,
  title     = {{ASQA}: Factoid Questions Meet Long-Form Answers},
  author    = {Stelmakh, Ivan and Luan, Yi and Dhingra, Bhuwan and Chang, Ming-Wei},
  booktitle = {Proceedings of the 2022 Conference on Empirical Methods in Natural Language Processing},
  pages     = {8273--8288},
  year      = {2022},
  month     = dec,
  address   = {Abu Dhabi, United Arab Emirates},
  publisher = {Association for Computational Linguistics},
  doi       = {10.18653/v1/2022.emnlp-main.566},
  url       = {https://aclanthology.org/2022.emnlp-main.566/},
}

@inproceedings{hou2024decomposing,
  title     = {Decomposing Uncertainty for Large Language Models through Input Clarification Ensembling},
  author    = {Hou, Bairu and Liu, Yujian and Qian, Kaizhi and Andreas, Jacob and Chang, Shiyu and Zhang, Yang},
  booktitle = {Proceedings of the 41st International Conference on Machine Learning},
  series    = {Proceedings of Machine Learning Research},
  volume    = {235},
  pages     = {19023--19042},
  year      = {2024},
  publisher = {PMLR},
  url       = {https://proceedings.mlr.press/v235/hou24b.html},
}

@inproceedings{kendall2017uncertainties,
  title     = {What Uncertainties Do We Need in {Bayesian} Deep Learning for Computer Vision?},
  author    = {Kendall, Alex and Gal, Yarin},
  booktitle = {Advances in Neural Information Processing Systems},
  volume    = {30},
  pages     = {5574--5584},
  year      = {2017},
  address   = {Long Beach, California, USA},
  publisher = {Curran Associates, Inc.},
  url       = {https://papers.nips.cc/paper/2017/hash/2650d6089a6d640c5e85b2b88265dc2b-Abstract.html},
}
